\documentclass[11pt]{article}

\let\counterwithin\relax

\usepackage{amsmath, amsfonts, amssymb, amsthm, bm, graphicx, mathtools, enumerate,multirow}
\usepackage[sort&compress]{natbib}
\usepackage[CJKbookmarks=true,
bookmarksnumbered=true,
bookmarksopen=true,
colorlinks=true,
citecolor=blue,
linkcolor=blue,
anchorcolor=blue,
urlcolor=blue]{hyperref}
\usepackage[usenames]{color}
\usepackage[letterpaper, left=1.2truein, right=1.2truein, top = 1.2truein,
bottom = 1.2truein]{geometry}
\usepackage[ruled, lined, commentsnumbered]{algorithm2e}
\usepackage{prettyref,soul}

\usepackage{apptools} 
\usepackage{chngcntr} 
\AtAppendix{\counterwithin{lemma}{section}} 
\usepackage{tikz}
\usepackage[font=small,labelfont=bf]{caption}
\usepackage{enumitem}

\usepackage[caption=false,font=normalsize,labelfont=sf,textfont=sf]{subfig}

\usepackage{booktabs} 
\usepackage{multirow} 
\usepackage{makecell}
\usepackage{siunitx}
\usepackage{adjustbox}
\usepackage{longtable}

\theoremstyle{definition}

\newrefformat{eq}{(\ref{#1})}
\newrefformat{chap}{Chapter~\ref{#1}}
\newrefformat{sec}{Section~\ref{#1}}
\newrefformat{algo}{Algorithm~\ref{#1}}
\newrefformat{fig}{Fig.~\ref{#1}}
\newrefformat{tab}{Table~\ref{#1}}
\newrefformat{rmk}{Remark~\ref{#1}}
\newrefformat{clm}{Claim~\ref{#1}}
\newrefformat{def}{Definition~\ref{#1}}
\newrefformat{cor}{Corollary~\ref{#1}}
\newrefformat{lmm}{Lemma~\ref{#1}}
\newrefformat{lemma}{Lemma~\ref{#1}}
\newrefformat{prop}{Proposition~\ref{#1}}
\newrefformat{app}{Appendix~\ref{#1}}
\newrefformat{ex}{Example~\ref{#1}}
\newrefformat{cond}{Condition~\ref{#1}}

\usepackage{threeparttablex}
\usepackage{float}
\usepackage{listings}
\usepackage{xcolor}

\usepackage{threeparttable}
\usepackage{threeparttablex}

\lstdefinestyle{mystyle}{
	backgroundcolor=\color{backcolour},   
	commentstyle=\color{codegreen},
	keywordstyle=\color{magenta},
	numberstyle=\tiny\color{codegray},
	stringstyle=\color{codepurple},
	basicstyle=\ttfamily\footnotesize,
	breakatwhitespace=false,         
	breaklines=true,                 
	captionpos=b,                    
	keepspaces=true,                 
	numbers=left,                    
	numbersep=5pt,                  
	showspaces=false,                
	showstringspaces=false,
	showtabs=false,                  
	tabsize=2
}

\usepackage{makecell}

\usepackage[T1]{fontenc}
\usepackage[utf8]{inputenc}
\usepackage{authblk}
\title{
Detecting low left ventricular ejection fraction from ECG using an interpretable and scalable predictor-driven framework
}

\author[1,*]{Ya Zhou}
\author[2]{Tianxiang Hao}
\author[3]{Ziyi Cai}
\author[4]{Haojie Zhu}
\author[3]{Kejun He}
\author[1]{Jia Liu}
\author[4,5, *]{Xiaohan Fan}
\author[1, *]{Jing Yuan}

\affil[1]{Department of Information Center, Fuwai Hospital, Chinese Academy of Medical Sciences and Peking Union Medical College, Beijing, China}
\affil[2]{The Tsinghua Shenzhen International Graduate School, Tsinghua University, Shenzhen, China}
\affil[3]{Institute of Statistics and Big Data, Renmin University of China, Beijing, China}
\affil[4]{Cardiac Arrhythmia Center, Fuwai Hospital, National Center for Cardiovascular Diseases, Chinese Academy of Medical Sciences and Peking Union Medical College, Beijing,  China}
\affil[5]{Function Test Center, Fuwai Hospital, National Center for Cardiovascular Diseases, Chinese Academy of Medical Sciences and Peking Union Medical
	College, Beijing, China}

\date{}

\begin{document}
\maketitle
\def\thefootnote{*}\footnotetext{Email: Jing Yuan (yuanjing@fuwai.com), Xiaohan Fan (fanxiaohan@fuwaihospital.org), Ya Zhou (zhouya@fuwai.com) }\def\thefootnote{\arabic{footnote}}

\begin{abstract}
Low left ventricular ejection fraction (LEF) frequently remains undetected until progression to symptomatic heart failure, underscoring the need for scalable screening strategies. Although artificial intelligence-enabled electrocardiography (AI-ECG) has shown promise, existing approaches rely solely on end-to-end black-box models with limited interpretability or on tabular systems dependent on commercial ECG measurement algorithms with suboptimal performance. We introduced ECG-based Predictor-Driven LEF (ECGPD-LEF), a structured framework that integrates foundation model-derived diagnostic probabilities with interpretable modeling  for detecting  LEF from ECG. Trained on the benchmark EchoNext dataset comprising 72,475 ECG-echocardiogram pairs and evaluated in predefined independent internal (n=5,442) and external (n=16,017) cohorts, our framework achieved robust discrimination for moderate LEF (internal AUROC 88.4\%, F1 64.5\%; external AUROC 86.8\%, F1 53.6\%), consistently outperforming the official end-to-end baseline provided with the benchmark across demographic and clinical subgroups. Interpretability analyses identified high-impact predictors, including normal ECG, incomplete left bundle branch block, and subendocardial injury in anterolateral leads, driving LEF risk estimation. Notably, these predictors independently enabled zero-shot-like inference without task-specific retraining (internal AUROC 75.3-81.0\%; external AUROC 71.6-78.6\%), indicating that ventricular dysfunction is intrinsically encoded within structured diagnostic probability representations. This framework reconciles predictive performance with mechanistic transparency, supporting scalable enhancement through additional predictors and seamless integration with existing AI-ECG systems.


\end{abstract}

\section{Introduction}
\label{sec:intro}
Heart failure (HF) is a major global health burden, affecting over 64 million individuals worldwide, with prevalence continuing to rise \citep{savarese2022global}. In many patients, left ventricular systolic dysfunction (LVSD) is initially asymptomatic, yet it frequently progresses to symptomatic HF and is associated with poor prognosis if left undetected. Low left ventricular ejection fraction (LEF) is a critical indicator that reflects the progression of LVSD toward clinically manifest HF \citep{heidenreich20222022}. Early detection of LVSD is critical, as timely initiation of evidence-based therapies can mitigate symptoms, slow disease progression, and improve survival. Although cardiovascular imaging, particularly echocardiography (ECHO), remains the clinical gold standard for assessing ventricular function, it is time- and resource-intensive, requires specialized operators and equipment, and is not readily scalable for population-level screening \citep{ciampi2007role}. These limitations underscore the need for cost-effective, widely accessible approaches capable of reliably identifying patients with LEF \citep{attia2019screening, yao2021artificial, tran2025electrocardiogram, lakshmanan2023comparison}.

Electrocardiography (ECG), which records cardiac electrical activity, is a promising tool for large-scale screening \citep{hughes2024simple}. Since its introduction by Willem Einthoven, which earned him the Nobel Prize in 1924, ECG science has steadily advanced: acquisition has become easier, signal quality has improved, and pathognomonic features for a wide range of rhythm and conduction disorders have been defined \citep{khera2024ai}. More recently, artificial intelligence (AI) has transformed ECG analysis, enabling models to achieve cardiologist-level performance in traditional interpretation tasks \citep{hannun2019cardiologist, ribeiro2020automatic, jiang2024self, li2025electrocardiogram} and to detect subtle patterns associated with conditions such as aortic stenosis \citep{kwon2020deep}, tricuspid regurgitation \citep{diao2025speed}, and LEF \citep{attia2019screening}, which are often imperceptible to humans. Despite these advances, clinical adoption varies: AI-ECG models are widely accepted for well-established conditions such as atrial fibrillation and sinus bradycardia, which can be visually confirmed according to guidelines \citep{joglar20242023, kusumoto20192018}, but remain limited for emerging tasks such as LEF detection due to concerns about interpretability and the difficulty of linking predictions to recognizable ECG features \citep{van2022improving, hughes2024simple}. To enhance interpretability and facilitate broader clinical deployment, a recent study proposed a random forest model based on 555 discrete ECG measurements based on commercial algorithms, which demonstrated superior performance compared with N-terminal prohormone brain natriuretic peptide (NT-proBNP) for LEF detection \citep{hughes2024simple}. Although this approach showed promising results, its performance is slightly below deep learning. Moreover, as highlighted in the original study, many features cannot be obtained when applied across different ECG machines, and measurement values differ markedly between devices \citep{strodthoff2023ptb}, which presents a significant barrier to broader clinical adoption.

In this study, we aim to evaluate the efficacy of interpretable methods for detecting LEF using standard 12-lead ECGs. We propose ECG-based Predictor-Driven LEF (ECGPD-LEF), a framework that integrates traditional AI-ECG interpretation models with either a single-predictor approach or a multi-predictor approach to enhance recognition of subtle ECG features and improve LEF detection. The single-predictor approach performs zero-shot-like inference, requiring no additional training on the target dataset. The multi-predictor approach was trained on the publicly available ECG-ECHO paired dataset EchoNext, a comprehensive dataset encompassing diverse populations across races and clinical contexts. We also constructed an external validation dataset, MIMIC-LEF, based on MIMIC-IV \citep{PhysioNet-mimiciv-3.1}, MIMIC-IV-ECG \citep{PhysioNet-mimic-iv-ecg-1.0}, and MIMIC-IV-Note\citep{PhysioNet-mimic-iv-note-2.2}, which similarly covers a heterogeneous population. The model was evaluated on multi-center datasets, including internal validation (the EchoNext test set) and external validation (MIMIC-LEF). The model performance was assessed across multiple dimensions, including the presence or absence of other structural heart disease (SHD) and valvular heart disease (VHD), as well as across different care settings and racial or ethnic subgroups. Global and local model explanations were performed to identify important predictors contributing to LEF detection, and to characterize these subtle ECG changes associated with LEF.

\section{Methods}
\label{sec:method}
\subsection{Data sources}

\begin{figure*}[!htbp]
	\centering
	\includegraphics[width=1.0\textwidth]{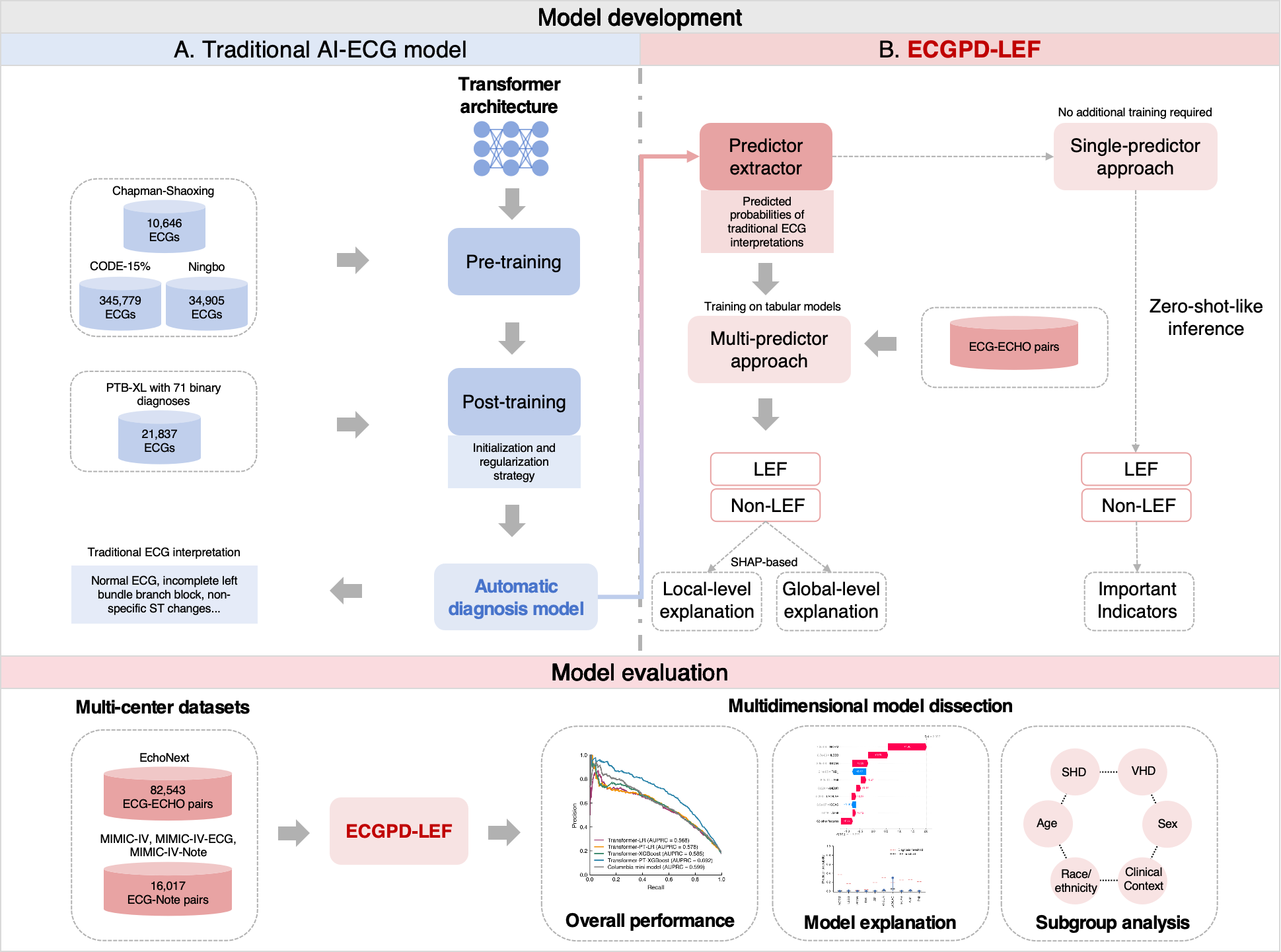}
\caption{
	\textbf{Flowchart of model development and evaluation.} Model development is based on a traditional AI-ECG interpretation model, where a large unlabeled ECG dataset is used for pre-training and a smaller labeled ECG dataset is used for enhanced post-training. The automatic ECG diagnosis model outputs diagnostic probabilities for each ECG finding and serves as the predictor extractor. Based on the predictor extractor, we developed both a single-predictor approach and a multi-predictor approach. The single-predictor approach enables zero-shot-like inference without further training to detect LEF and can serve as an important indicator. The multi-predictor approach further trains a tabular model using ECG-ECHO pairs and provides both local- and global-level explanations based on SHAP values. Model evaluation is conducted on the EchoNext test set and an independently collected ECHO-Note pairs dataset. Multidimensional model dissection is performed, including overall performance evaluation, model interpretability analysis, and subgroup analyses across diverse populations. LEF, low left ventricular ejection fraction; No-LEF, absence of low left ventricular ejection fraction; SHD, structural heart disease; VHD, valvular heart disease, SHAP, Shapley Additive exPlanations.
}
	\label{fig:flowchart}
\end{figure*}

An overview of the study design is provided in Figure~\ref{fig:flowchart}, illustrating the data sources, construction of the predictor extractor, model development for low left ventricular ejection fraction (LEF) detection, and downstream evaluation. Four ECG-only datasets were implicitly used to develop the predictor extractor, and two independent paired datasets (ECG-ECHO and ECG-Note) were explicitly used for development and testing.

The ECG-only datasets included Ningbo~\citep{zheng2020optimal}, Chapman~\citep{zheng202012}, CODE-15\%~\citep{ribeiro_2021_4916206}, and PTB-XL~\citep{wagner2020ptb}. Information from Ningbo, Chapman, and CODE-15\% was implicitly incorporated through the pre-trained weights of a Transformer-based model (ST-MEM)~\citep{na2024guiding}. PTB-XL was subsequently used for post-training to obtain the final predictor extractor.

For the detection of LEF, we used the ECG-ECHO paired EchoNext dataset collected at Columbia University Irving Medical Center~\citep{elias2025echonext}. External validation was performed using an ECG-Note paired dataset constructed from MIMIC-IV \citep{PhysioNet-mimiciv-3.1}, MIMIC-IV-ECG \citep{PhysioNet-mimic-iv-ecg-1.0}, and MIMIC-IV-Note \citep{PhysioNet-mimic-iv-note-2.2}, comprising records from Beth Israel Deaconess Medical Center.

\subsection{Implicit ECG-only datasets}
Chapman \citep{zheng202012} and Ningbo, created under the auspices of Chapman University, Shaoxing People’s Hospital, and Ningbo First Hospital, contains 10,646 and 34,905 12-lead ECG, respectively, stored in MUSE ECG system with a sampling rate of 500Hz and a duration of 10 seconds. Code-15\% \citep{ribeiro_2021_4916206} is obtained through stratified sampling from the CODE dataset \citep{ribeiro2020automatic}, containing 345,779 exams from 233,770 patients, collected by the Telehealth Network of Minas Gerais in the period between 2010 and 2016. The S12L-ECG exam was performed mostly in primary care facilities using a tele-electrocardiograph manufactured by Tecnologia Eletrônica Brasileira (São Paulo, Brazil)—model TEB ECGPC—or Micromed Biotecnologia (Brasilia, Brazil)—model ErgoPC 13 and the duration of the ECG recordings is between 7 and 10 s sampled at frequencies ranging from 300 to 600 Hz. The processed subset of these datasets are implicitly in the pre-trained weights of the traditional automatic ECG diagnosis model  (the traditional AI-ECG model).

PTB-XL \citep{wagner2020ptb} dataset was recorded by devices from Schiller AG between October 1989 and June 1996, consisting 21,837 clinical 12-lead ECG records of 10 seconds length from 18885 patients. The ECG records were annotated by up to two cardiologists with potentially multiple ECG statements out of a set of 71 different statements conforming to the SCP-ECG standards \citep{scp_ecp}. This dataset is a commonly used benchmark for traditional ECG interpretation algorithms \citep{strodthoff2020deep}. The dataset id divided into training, validation, and test sets with a ratio of 8:1:1. In this paper, the dataset is implicitly in the post-trained weights of traditional automatic ECG diagnosis models (the traditional AI-ECG model).

\subsection{ECG-ECHO and ECG-Note datasets}
EchoNext is a recent published ECG-ECHO paired dataset for benchmarking ECG screening models \citep{elias2025echonext}.  It involves 82,543 de-identified paired ECG-ECHO records from 36,286 unique patients and those aged 18 years or older were identified, who underwent a digitally stored 12-lead ECG and a transthoracic ECHO within a 1-year interval between 2008 and 2022. The dataset is divided into training, validation, and test splits. There might includes multiple ECG-ECHO pairs in the training set, while only the latest ECG are adopted in the validation and test sets. 
The ECG signals were extracted from the GE MUSE ECG management system at a sampling frequency of 250 Hz across all 12 leads with 10 seconds duration.

The ECG-Note dataset is constructed based on MIMIC-IV \citep{PhysioNet-mimiciv-3.1}, MIMIC-IV-ECG \citep{PhysioNet-mimic-iv-ecg-1.0}, and MIMIC-IV-Note \citep{PhysioNet-mimic-iv-note-2.2}, as illustrated in Figure \ref{fig:lvef_llm_workflow} of the Appendix. We initially identified adult patients who possessed at least one standard 10-second 12-lead ECG with sampling frequency of 500Hz and a corresponding clinical note recorded within one year following the ECG. Exclusion criteria were applied to remove: 1) pairs where clinical notes lacked keywords related to ejection fraction(EF) and ECHO/TTE; 2) ECGs containing missing data (NaN). Following these exclusions, the samples were cross-referenced with MIMIC-ECG-LVEF \citep{li2025electrocardiogram} to identify consistent ECG-EF class pairs. For each patient, only the ECG-Note pair with the shortest time interval was retained, resulting in a final testing set of 16,017 ECGs.

\subsection{Outcomes}

For the ECG-ECHO paired dataset, values of left ventricular ejection fraction were extracted from Syngo Dynamics (Siemens) and Xcelera (Philips). The labeling strategy was defined in accordance with \cite{poterucha2025detecting}, using an upper threshold of 45\% to define  low left ventricular ejection fraction (LEF). An ECG was labeled positive if it was performed within 1 year prior to an echocardiogram demonstrating an ejection fraction $\le$ 45\%. For patients confirmed to have ejection fraction $>$ 45\%, all ECGs prior to the most recent ECHO were labeled negative.

For the ECG-Note paired dataset, a large language model\citep{qwen2025qwen25technicalreport} using a modified strategy adapted from \cite{gao2025machine} was employed to extract ejection fraction values from the discharge table of MIMIC-IV-Note (Figure \ref{fig:lvef_llm_workflow}b in the Appendix). The positive class was defined as ejection fraction $\le$ 45\%, consistent with the definition applied to the ECG-ECHO dataset.

\subsection{Model development}
We developed an ECG-based Predictor-Driven LEF framework (ECGPD-LEF) for the detection of LEF. As illustrated in Figure \ref{fig:flowchart}b, the framework comprises two components: (1) a predictor extractor that generates structured probabilistic representations from raw ECG waveforms and (2) predictor-based inference models, including single-predictor and multi-predictor approaches. The single-predictor approach performs LEF inference without additional task-specific training, whereas the multi-predictor approach learns a lightweight tabular classifier based on the extracted predictors.

\subsubsection{Predictor extractor}
To instantiate the predictor extractor, we adopted a Transformer-based automatic ECG diagnosis model. Specifically, we used the ST-MEM architecture \citep{na2024guiding}, pre-trained on the Chapman, Ningbo, and CODE-15\% datasets. Using the publicly available pre-trained weights, we further fine-tuned the model on the PTB-XL dataset to predict 71 conventional ECG diagnoses. Two training strategies were implemented: the original approach (denoted as Transformer) and a modified post-training strategy (denoted as Transformer-PT) described previously \citep{zhou2025bridging}. The final post-trained model, equipped with sigmoid activation, outputs probability estimates for each of the 71 diagnoses. These probabilistic outputs were used as predictors for the downstream LEF modeling task.

\subsubsection{Single-predictor and multi-predictor approaches}
For each ECG recording, the predictor extractor generates 71 predictor values ranging from 0 to 1, each corresponding to a clinically defined ECG interpretation. In the single-predictor approach, each predictor value (PV) was evaluated independently for LEF detection without additional training on the ECG-ECHO dataset. For predictors corresponding to normal ECG or sinus rhythm, we used $(1 - \mathrm{PV})$ to reflect their inverse clinical association with reduced LEF; for all other predictors, the original PV was used directly. Threshold-independent metrics, including AUROC and AUPRC, were computed directly on the test set. For the F1 score, the classification threshold was selected by maximizing validation-set performance.

In the multi-predictor approach, the 71 predictors were jointly modeled using lightweight tabular classifiers. As a linear model, we implemented logistic regression with an $l_2$ penalty, with the regularization parameter selected via grid search over \{0.001, 0.01, 0.1, 1.0, 10.0\} on the validation set. As a nonlinear alternative, we implemented XGBoost \citep{chen2016xgboost}, a gradient-boosted decision tree method well suited for structured data. The learning rate and maximum tree depth were tuned via grid search over \{0.05, 0.1, 0.2\} and \{3, 5, 7\}, respectively. The number of estimators was set to 1000 with early stopping (30 rounds) based on validation performance. For both models, the final classification threshold was determined by maximizing the F1 score on the validation set.

\subsection{Performance evaluation}
We first evaluated the predictor extractor on the PTB-XL test set, as it constitutes a key component of the proposed framework. LEF detection performance was subsequently assessed on two independent datasets: the internal test set, consisting of held-out test set from the ECG-ECHO dataset, and the external test set, ECG-Note, which was constructed in this study (Figure \ref{fig:flowchart}). Both single- and multi-predictor approaches were evaluated on these datasets.

Model performance was quantified using the area under the receiver operating characteristic curve (AUROC), area under the precision-recall curve (AUPRC), and F1 score, consistent with prior work \citep{poterucha2025detecting}. Confidence intervals were estimated via 1,000 bootstrap resamples. For benchmarking, we compared our method against the Columbia mini deep learning model \citep{poterucha2025detecting}, the official end-to-end baseline for LEF detection, which is publicly available with source code and pre-trained weights, enabling reproducible comparison. The Columbia mini model was also evaluated on the external test set (see Section \ref{Apedix:external_comparison} of the Appendix).

To assess the relative contributions of different components and design choices, we further evaluated multiple configurations of the multi-predictor approach, including alternative predictor extractors, tabular models, and varying numbers of predictors.

\subsection{Interpretability}
The predictor-driven framework enables transparent interpretation for both single- and multi-predictor approaches. In the single-predictor approach, each predictor value directly represents the probability of the corresponding ECG diagnosis, providing inherent clinical interpretability. We identified the most important predictors and performed combined analyses with the multi-predictor approach.

For the multi-predictor approach, we employed SHAP (SHapley Additive exPlanations) values \citep{shapley1953value, lundberg2017unified} to quantify feature contributions at global and local levels. SHAP is a game-theoretic, additive feature attribution method that provides consistent and locally accurate explanations \citep{lundberg2017unified}. For computational efficiency in tree-based models, we used the Tree SHAP algorithm \citep{lundberg2018consistent}. Global explanation plots included cumulative contributions, beeswarm summaries, and SHAP versus predictor value plots to identify key predictors and characterize their behavior in the model. Local plots highlighted the ten predictors SHAP values and displayed predicted probabilities relative to both the ECG diagnosis thresholds and the LEF decision thresholds, facilitating interpretation of individual model predictions.

\subsection{Subgroup analysis}
We evaluated model performance across clinically relevant subgroups in both the internal and external test sets. In the internal test set, predefined subgroups included the presence or absence of other structural heart disease (SHD), valvular heart disease (VHD), age groups, sex, race/ethnicity, and clinical context (definitions of SHD and VHD are provided in Section \ref{apedix:subgroup_multi_predictor} of the Appendix). In the external test set, subgroup evaluation was performed for the available variables, including age groups, sex, race/ethnicity, and clinical context.

\section{Results}
\subsection{Population characteristics}
The ECG-ECHO (EchoNext) cohort is a publicly available benchmark comprising 82,543 ECG examinations, partitioned into training (n=72,475), validation (n=4,626), and internal test (n=5,442) sets according to the official protocol \citep{poterucha2025detecting}. We further constructed an external cohort (ECG-Note; n=16,017) as an independent test set (Figure \ref{fig:lvef_llm_workflow}). Baseline demographic and clinical characteristics are summarized in Table \ref{tab:num_echonext}.

Compared with the internal test cohort, the external cohort exhibited a higher proportion of male patients (54.9\%) and White individuals (73.0\%), whereas racial/ethnic representation in the ECG-ECHO cohort was more evenly distributed. Clinical context distributions also differed substantially, with the external cohort enriched for emergency encounters (57.3\%) compared with the more evenly distributed emergency (36.2\%), inpatient (40.5\%), and outpatient (19.5\%) settings in the internal test cohort. These differences reflect substantial demographic and clinical heterogeneity across cohorts. The distribution of the 71 extracted predictors is detailed in Appendix Table~\ref{tab:num_echonext_2}, which also shows differences in binarized counts for selected predictors, including NORM and ILBBB.

\begin{table}[!htbp]
	\centering
	\begin{ThreePartTable}
		\footnotesize
	\setlength{\tabcolsep}{4pt}{
\caption{
	Baseline demographic and clinical characteristics in the ECG-ECHO and external ECG-Note cohorts.
}
	\label{tab:num_echonext}
	\begin{tabular}{l|c|c|c|c}
		\toprule
		& \multicolumn{3}{c|}{\textbf{ECG--ECHO}} & {\textbf{ECG--Note}} \\
		\cmidrule(lr){2-4}\cmidrule(lr){5-5}
		& \textbf{Training set} & \textbf{Validation set} & \textbf{Internal test set} & \textbf{External test set} \\
		\midrule
		{Patients (n)} & 26,218 & 4,626 & 5,442 & 16,017 \\
		{ECGs (n)} & 72,475 & 4,626 & 5,442 & 16,017 \\
		\midrule
		\multicolumn{5}{l}{ {Age groups } } 	 \\
		\midrule
		\hspace{0.5em}  18--59 & 29,783 (41.1\%) & 1,787 (38.6\%) & 2,124 (39.0\%) & 4,270 (26.7\%) \\
		\hspace{0.5em}  60--69 & 18,745 (25.9\%) & 1,093 (23.6\%) & 1,318 (24.2\%) & 3,637 (22.7\%) \\
		\hspace{0.5em}  70--79 & 14,898 (20.6\%) & 975 (21.1\%) & 1,154 (21.2\%) & 3,761 (23.5\%) \\
		\hspace{0.5em}  80+ & 9,049 (12.5\%) & 771 (16.7\%) & 846 (15.5\%) & 4,349 (27.2\%) \\
		\midrule
		\multicolumn{5}{l}{ {Sex}  }  \\
		\midrule
		\hspace{0.5em}  Female & 33,524 (46.3\%) & 2,356 (50.9\%) & 2,731 (50.2\%) & 7,222 (45.1\%) \\
		\hspace{0.5em}  Male & 38,951 (53.7\%) & 2,270 (49.1\%) & 2,711 (49.8\%) & 8,795 (54.9\%) \\
		\midrule
		\multicolumn{5}{l}{{Race/ethnicity} } \\
		\midrule
		\hspace{0.5em}  Hispanic & 22,806 (31.5\%) & 1,351 (29.2\%) & 1,649 (30.3\%) & 638 (4.0\%) \\
		\hspace{0.5em}  White & 21,289 (29.4\%) & 1,385 (29.9\%) & 1,569 (28.8\%) & 11,688 (73.0\%) \\
		\hspace{0.5em}  Black & 11,559 (15.9\%) & 728 (15.7\%) & 846 (15.5\%) & 1,845 (11.5\%) \\
		\hspace{0.5em}  Asian & 2,602 (3.6\%) & 134 (2.9\%) & 153 (2.8\%) & 414 (2.6\%) \\
		\hspace{0.5em}  Other & 5,272 (7.3\%) & 380 (8.2\%) & 457 (8.4\%) & 524 (3.3\%) \\
		\hspace{0.5em}  Unknown & 8,947 (12.3\%) & 648 (14.0\%) & 768 (14.1\%) & 908 (5.7\%) \\
		\midrule
		\multicolumn{5}{l}{{Clinical context} }	  \\
		\midrule
		\hspace{0.5em}  Emergency & 22,811 (31.5\%) & 1,688 (36.5\%) & 1,971 (36.2\%) & 9,170 (57.3\%) \\
		\hspace{0.5em}  Inpatient & 34,906 (48.2\%) & 1,903 (41.1\%) & 2,203 (40.5\%) & - \\
		\hspace{0.5em}  Outpatient & 12,423 (17.1\%) & 858 (18.5\%) & 1,059 (19.5\%) & - \\
		\hspace{0.5em}  Procedural & 2,335 (3.2\%) & 177 (3.8\%) & 209 (3.8\%) & - \\
            \hspace{0.5em}  Urgent & - & - & - & 3,028 (18.9\%) \\
            \hspace{0.5em}  Observation & - & - & - & 2,173 (13.6\%) \\
            \hspace{0.5em}  Surgical Same Day & - & - & - & 1,022 (6.4\%) \\
            \hspace{0.5em}  Elective & - & - & - & 624 (3.9\%) \\
		\midrule
		\multicolumn{5}{l}{Outcome} \\
		\midrule
		\hspace{0.5em} Ejection Fraction \(\leq\) 45\% & 16,962 (23.4\%) & 866 (18.7\%) & 962 (17.7\%) & 2,517 (15.7\%) \\
		\bottomrule
	\end{tabular}
}
\begin{tablenotes}
	\footnotesize
	\item The training, validation, and internal test splits correspond to the official splits of the ECG-ECHO dataset, EchoNext \citep{elias2025echonext}. The external test cohort, ECG-Note, was derived from the MIMIC-IV database and its associated ECG and clinical note modules \citep{PhysioNet-mimiciv-3.1,PhysioNet-mimic-iv-ecg-1.0,PhysioNet-mimic-iv-note-2.2}. Values are shown as counts and percentages.
\end{tablenotes}
\end{ThreePartTable}
\end{table}


\subsection{Predictor recognition by the predictor extractor}
\label{sec:sub_sec_predictor_recog}
Reliable predictor extraction is a prerequisite for downstream LEF modeling. Transformer-PT achieved a macro AUROC of 94.5\%, AUPRC of 41.4\%, and F1 score of 38.3\%, outperforming the original Transformer (AUROC 89.8\%, AUPRC 30.7\%) and demonstrating performance comparable to previously reported state-of-the-art results on PTB-XL \citep{zhou2025bridging}. Classification performance for 10 representative predictors is shown in Table~\ref{tab:results_single_predictor}, with results for all 71 predictors provided in Appendix Table~\ref{tab:results_single_predictor-2}. Across predictors, Transformer-PT achieved consistently strong discriminative capacity as reflected by AUROC values, whereas AUPRC and F1 varied due to class imbalance (see Appendix Section \ref{Apedix:predictor_recog}). Importantly, the downstream LEF framework leverages continuous predictor scores rather than threshold-dependent binary decisions; thus, the strong ranking performance of Transformer-PT is sufficient to ensure reliable information transfer to subsequent modeling stages, even for ultra-rare categories.

\subsection{Single-predictor approach performance}
\label{sec:sub_sec_single_predictor}
We evaluated the ability of individual predictors to detect LEF using their continuous output scores, without task-specific fine-tuning (zero-shot-like inference). Results for 10 representative predictors in the internal test set are summarized in Table~\ref{tab:results_single_predictor}, with results for the remaining predictors and the external test set provided in Appendix Tables~\ref{tab:results_single_predictor-2} and \ref{tab:results_single_predictor-2-ecgnote}. Predictors are ordered by decreasing F1 score in the internal test set. Notably, several predictors demonstrated substantial standalone discriminative performance. Eight predictors (NORM, ILBBB, INJAL, ISCLA, ANEUR, ISCAL, ASMI, and SVARR) achieved AUROC values ranging from 71.0\% to 81.0\% internally, with five maintaining AUROC values between 70.7\% and 78.8\% externally. The NORM predictor was the strongest individual predictor in both cohorts, yielding an AUROC of 81.0\%, AUPRC of 47.4\%, and F1 score of 51.4\% internally, and an AUROC of 78.8\%, AUPRC of 36.2\%, and F1 score of 42.3\% externally. Among abnormal ECG diagnoses, ILBBB and INJAL achieved the highest discriminative performance, with internal AUROCs of 80.0\% and 75.3\%, and external AUROCs of 73.4\% and 71.6\%, respectively.

Across all 71 predictors, discriminative performance varied substantially. Nevertheless, the majority of predictors, 58 in the internal test set and 54 in the external test set, achieved AUROC values above chance level (50\%), suggesting that LEF-related information is distributed across diverse ECG-derived predictors rather than confined to a small subset of diagnoses. Predictors with weaker standalone performance still contributed incremental improvements when integrated into the multi-predictor model described in the next subsection.

An additional observation was that optimal LEF detection thresholds were consistently substantially lower than those used for conventional ECG classification. For example, while a threshold of 0.370 for the NORM predictor identified abnormal ECGs, a substantially lower threshold of 0.003641 was optimal for LEF detection, with similar patterns observed across most predictors (Appendix Figure~\ref{fig:LEF_thresholds}).

\begin{table}[!htbp]
	\centering
	\begin{ThreePartTable}
\caption{
	Performance of traditional AI-ECG predictors and LEF detection using a single-predictor approach.
}
	\label{tab:results_single_predictor}
	\footnotesize
	\setlength{\tabcolsep}{3pt}
	\begin{tabular}{l|cccc|cccc}
		\toprule
		\multirow{2.5}{*}{\textbf{Predictor} } & 
		\multicolumn{4}{c}{\textbf{Traditional ECG Model}} & 
		\multicolumn{4}{c}{\textbf{LEF Detection (Single-predictor)}} \\
		\cmidrule(lr){2-5} \cmidrule(lr){6-9}
		& \textbf{AUROC} & \textbf{AUPRC} & \textbf{F1 Score} & \textbf{Thresh} & \textbf{AUROC} & \textbf{AUPRC} & \textbf{F1 Score} & \textbf{Thresh} \\
		\midrule
		NORM  & \makecell{94.9\\(94.1--95.7)} & \makecell{92.8\\(91.4--94.2)} & \makecell{85.1\\(83.5--86.7)} & 0.370     & \makecell{81.0\\(79.6--82.4)} & \makecell{47.4\\(44.2--50.9)} & \makecell{51.4\\(49.3--53.8)} & 0.003641 \\
		ILBBB & \makecell{90.9\\(65.3--99.7)} & \makecell{31.6\\( 8.1--70.3)} & \makecell{30.0\\( 0.0--55.6)} & 0.163     & \makecell{80.0\\(78.4--81.5)} & \makecell{48.5\\(45.3--52.0)} & \makecell{50.5\\(48.0--53.1)} & 0.000349 \\
		INJAL & \makecell{98.6\\(97.3--99.6)} & \makecell{52.2\\(27.0--79.8)} & \makecell{47.6\\(19.0--72.0)} & 0.500     & \makecell{75.3\\(73.7--76.8)} & \makecell{34.8\\(32.4--37.6)} & \makecell{45.1\\(42.8--47.4)} & 0.000386 \\
		ISCLA & \makecell{92.3\\(85.5--97.6)} & \makecell{19.0\\( 5.6--45.9)} & \makecell{12.5\\( 0.0--36.4)} & 0.248     & \makecell{75.9\\(74.5--77.4)} & \makecell{38.9\\(36.2--42.3)} & \makecell{43.9\\(41.6--46.3)} & 0.001918 \\
		ANEUR & \makecell{96.7\\(91.7--99.2)} & \makecell{15.0\\( 6.0--37.3)} & \makecell{11.8\\( 0.0--33.3)} & 0.294     & \makecell{74.8\\(73.1--76.6)} & \makecell{38.9\\(35.9--42.3)} & \makecell{42.9\\(40.4--45.4)} & 0.001924 \\
		ISCAL & \makecell{95.3\\(94.0--96.6)} & \makecell{33.9\\(25.2--46.8)} & \makecell{34.8\\(23.3--45.4)} & 0.315     & \makecell{73.7\\(72.1--75.2)} & \makecell{32.1\\(29.8--34.6)} & \makecell{42.3\\(40.1--44.4)} & 0.001148 \\
		ASMI  & \makecell{98.1\\(97.4--98.7)} & \makecell{88.4\\(84.9--91.4)} & \makecell{80.3\\(76.4--83.8)} & 0.300     & \makecell{72.8\\(71.0--74.7)} & \makecell{37.4\\(34.3--40.5)} & \makecell{41.8\\(39.4--43.9)} & 0.023987 \\
		SVARR & \makecell{92.0\\(85.2--97.2)} & \makecell{21.3\\( 6.1--45.9)} & \makecell{22.2\\( 5.9--40.0)} & 0.061     & \makecell{71.0\\(69.3--72.6)} & \makecell{31.8\\(29.5--34.5)} & \makecell{40.4\\(38.1--42.9)} & 0.000218 \\
		INJIL & \makecell{92.9\\(85.8--99.2)} & \makecell{ 2.3\\( 0.3--12.0)} & \makecell{ 0.0\\( 0.0-- 0.0)} & 0.012     & \makecell{69.8\\(68.0--71.5)} & \makecell{29.5\\(27.3--32.0)} & \makecell{40.1\\(37.9--42.3)} & 0.000035 \\
		CRBBB & \makecell{99.8\\(99.6--99.9)} & \makecell{89.1\\(80.8--95.3)} & \makecell{83.2\\(74.8--89.8)} & 0.144     & \makecell{69.7\\(68.1--71.5)} & \makecell{28.4\\(26.3--30.8)} & \makecell{39.5\\(37.6--41.6)} & 0.000005 \\
		\bottomrule
	\end{tabular}
\begin{tablenotes}
	\footnotesize
	\item Predictor performance was obtained from the traditional automatic AI-ECG model. LEF detection performance was derived using the single-predictor approach proposed in this study. AUROC, AUPRC, and F1 are reported with 95\% confidence intervals. The two Thresh columns indicate the thresholds used to maximize the F1 score on the validation set for predictor performance and LEF detection, respectively. The 10 predictors with the highest F1 scores for LEF detection are reported in this table. NORM, normal ECG; ILBBB, incomplete left bundle branch block; INJAL, subendocardial injury in anterolateral leads; ISCLA, ischemic in lateral leads; ANEUR, ST-T changes compatible with ventricular aneurysm; ISCAL, ischemic in anterolateral leads; ASMI, anteroseptal myocardial infarction; SVARR, supraventricular arrhythmia; INJIL, subendocardial injury in inferolateral leads; CRBBB, complete right bundle branch block.
\end{tablenotes}
	\end{ThreePartTable}
\end{table}

\newpage

\subsection{Multi-predictor approach performance}
\label{subsec:multi-predictor}

We evaluated the ECGPD-LEF multi-predictor framework across combinations of predictor extractors and tabular models, benchmarking against the official end-to-end Columbia mini model \citep{poterucha2025detecting} (Table \ref{tab:method_config_comparison}). Among all configurations, XGBoost combined with the post-trained Transformer predictor extractor (Transformer-PT-XGBoost) achieved the highest performance in both internal and external test sets. In the internal test set, Transformer-PT-XGBoost yielded an AUROC of 88.4\% (95\% CI, 87.1--89.5), an AUPRC of 69.2\% (66.2--72.0), and an F1 score of 64.5\% (62.2--66.7), significantly outperforming the Columbia mini model, which achieved an AUROC of 85.2\% (83.9--86.5), an AUPRC of 59.9\% (56.6--63.3), and an F1 score of 57.9\% (55.4--60.3), with non-overlapping confidence intervals. Comparable performance improvements were observed in the external test set (Table \ref{tab:method_config_comparison}).

\begin{figure}[!htbp]
	\centering
		\begin{minipage}{0.32\textwidth}
		\centering
		\includegraphics[width=\linewidth]{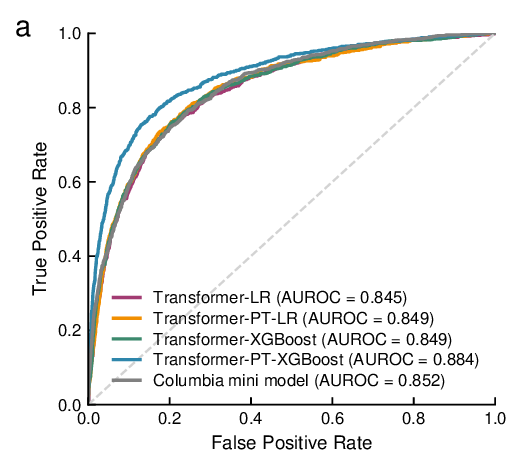}
	\end{minipage}
		\begin{minipage}{0.32\textwidth}
	\centering
	\includegraphics[width=\linewidth]{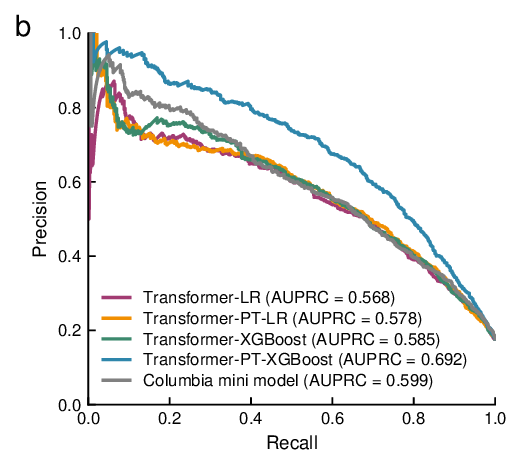}
\end{minipage} \\
	\begin{minipage}{0.32\textwidth}
		\centering
		\includegraphics[width=\linewidth]{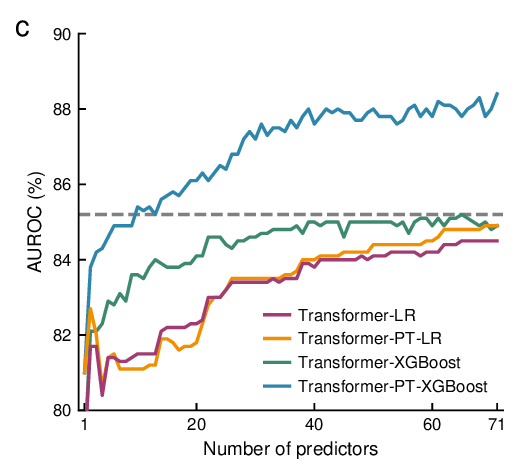}
	\end{minipage}
	\hfill
	\begin{minipage}{0.32\textwidth}
		\centering
		\includegraphics[width=\linewidth]{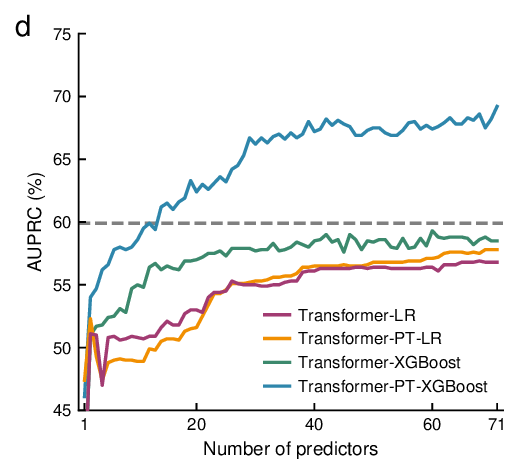}
	\end{minipage}
	\hfill
	\begin{minipage}{0.32\textwidth}
		\centering
		\includegraphics[width=\linewidth]{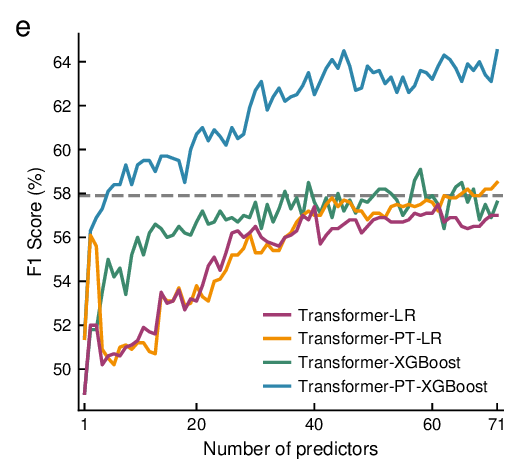}
	\end{minipage}
\caption{
	\textbf{Model performance across different evaluation settings.}
	The first row shows receiver operating characteristic (ROC) curves (a) and precision--recall (PR) curves (b) for five methods using the full feature set (71 predictors).
	The second row presents the performance of four models evaluated with an increasing number of predictors (1--71), in terms of AUROC (c), AUPRC (d) and F1 score (e).
	The gray dashed line indicates the Columbia mini model and serves as a reference baseline.
}
	\label{fig:performance}
\end{figure}

To disentangle the contributions of individual components, we conducted controlled comparisons on the internal test set (Table \ref{tab:method_config_comparison}; Figure \ref{fig:performance}). Holding the predictor extractor constant, XGBoost consistently outperformed logistic regression for both Transformer and Transformer-PT, with the largest gains observed for Transformer-PT (AUROC +3.5 percentage points; AUPRC +11.4 points). Holding the tabular model constant, post-training of the predictor extractor (Transformer-PT vs. Transformer) yielded consistent improvements across all metrics.

We further examined performance as predictors were progressively added to the tabular model in descending order of single-predictor F1 score. Across configurations, performance generally improved with additional predictors, with Transformer-PT-XGBoost consistently achieving the highest performance across all predictor counts and metrics (Figure \ref{fig:performance}b-d). Other configurations followed similar trends but did not exceed the Columbia mini model baseline for AUROC or AUPRC, even at their respective peak performance.

\begin{table}[!htbp]
	\begin{ThreePartTable}
	\centering
	\caption{Performance comparison of the proposed ECGPD-LEF framework and baseline Columbia mini model for LEF detection.
	}
	\label{tab:method_config_comparison}
	\footnotesize
	\setlength{\tabcolsep}{5pt}
	\begin{tabular}{l|c|c|ccc}
		\toprule
		\textbf{Method} & \textbf{Tabular Model} & \textbf{Predictor Extractor} & \textbf{AUROC} & \textbf{AUPRC} & \textbf{F1 Score} \\
		\midrule
			\multicolumn{6}{l}{\textbf{Internal test set }} \\	
		\midrule
		\makecell{Columbia\\mini model} & -- & -- & \makecell{85.2\\(83.9--86.5)} & \makecell{59.9\\(56.6--63.3)} & \makecell{57.9\\(55.4--60.3)} \\
		\midrule
		\multirow{7}{*}{ECGPD-LEF} 
		& \multirow{3}{*}{Logistic Regression}  & Transformer & \makecell{84.5\\(83.1--85.9)} & \makecell{56.8\\(53.4--60.3)} & \makecell{57.0\\(54.6--59.7)} \\
		& & Transformer-PT & \makecell{84.9\\(83.6--86.3)} & \makecell{57.8\\(54.6--61.2)} & \makecell{58.5\\(56.2--60.8)} \\
		
		\cline{2-6}
		& \multirow{3}{*}{XGBoost}  & Transformer & \makecell{84.9\\(83.5--86.2)} & \makecell{58.5\\(55.1--61.7)} & \makecell{57.6\\(55.1--60.1)} \\
		& & Transformer-PT & \makecell{ \textbf{88.4}\\(87.1--89.5)} & \makecell{\textbf{69.2}\\(66.2--72.0)} & \makecell{\textbf{64.5}\\(62.2--66.7)} \\
			\midrule
			\multicolumn{6}{l}{\textbf{External test set }} \\	
				\midrule
					\makecell{Columbia\\mini model} & -- & -- & \makecell{80.8\\(79.9-81.7)} & \makecell{45.8\\(43.7-47.7)} & \makecell{47.7\\(46.3-49.1)} \\
			\midrule
			\makecell{ECGPD-LEF} & XGBoost & Transformer-PT & \makecell{\textbf{86.9}\\(86.2-87.7)} & \makecell{\textbf{57.6}\\(55.6-59.7)} & \makecell{\textbf{53.8}\\(52.5-55.0)} \\
		\bottomrule
	\end{tabular}
\begin{tablenotes}
	\footnotesize
	\item The Columbia mini model is the official benchmark on the internal test set, EchoNext \citep{poterucha2025detecting}, with trained official weights. This model relies on seven tabular features; since the internal test set provides all seven features, it can be applied directly. For the external cohort, five of these features are not directly available, so we computed them from the MIMIC-IV ECG module to enable inference with the Columbia mini model (details of this feature computation are provided in Section \ref{Apedix:external_comparison}). The proposed ECGPD-LEF framework is evaluated with different configurations of automatic AI-ECG models and tabular models. For example, "XGBoost" indicates that the tabular model is XGBoost applied to the extracted predictors, and "Transformer-PT" indicates that the predictor extractor of the proposed framework is based on Transformer-PT. Since the baseline Columbia mini model is an end-to-end method, it does not depend on the choice of predictor extractor or tabular model. This table thus compares both the performance of the baseline model and the performance of the proposed framework under different configurations. AUROC, AUPRC, and F1 are reported with 95\% confidence intervals.
\end{tablenotes}
\end{ThreePartTable}
\end{table}

%
%

\subsection{Model explanation}
\label{sec:model_explanation}

\begin{figure}[!htbp]
	\centering
	\begin{minipage}{0.32\textwidth}
		\centering
		\includegraphics[width=\linewidth]{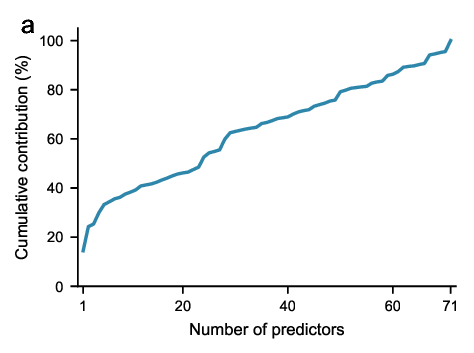}
		\label{fig:global_shap_a}
	\end{minipage}
	\begin{minipage}{0.32\textwidth}
	\centering
	\includegraphics[width=\linewidth]{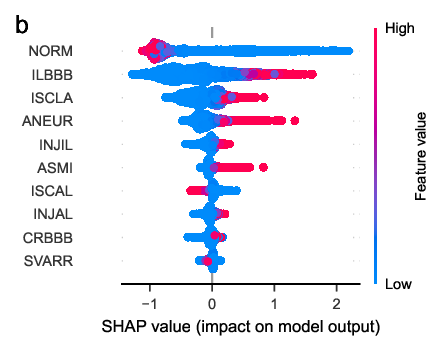}
	\label{fig:global_shap_b}
\end{minipage}
\\
\hfill
\begin{minipage}{0.32\textwidth}
\centering
\includegraphics[width=\linewidth]{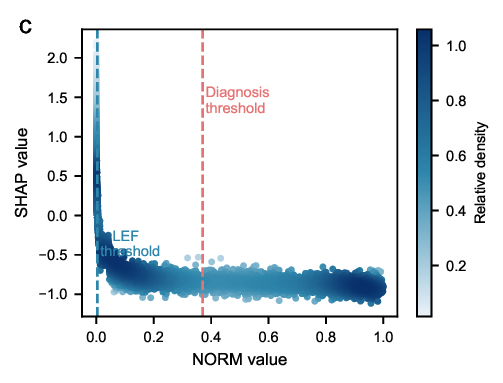}
	\label{fig:global_shap_c}
\end{minipage}
\hfill
\begin{minipage}{0.32\textwidth}
\centering
\includegraphics[width=\linewidth]{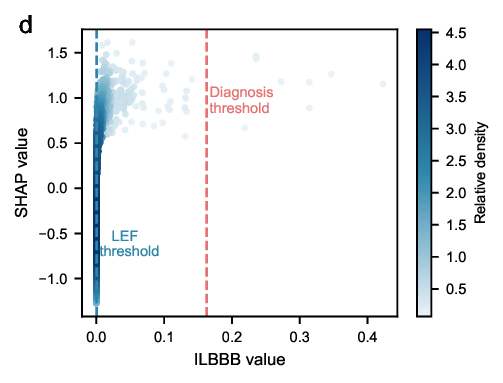}
	\label{fig:global_shap_d}
\end{minipage}
\hfill
\begin{minipage}{0.32\textwidth}
\centering
\includegraphics[width=\linewidth]{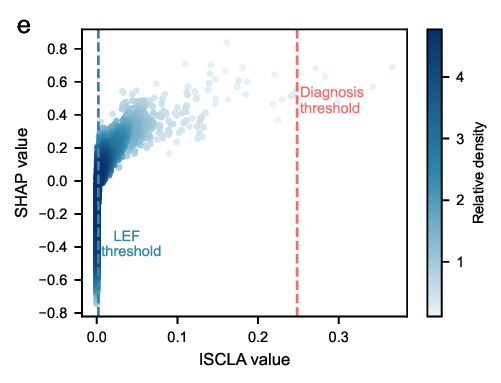}
	\label{fig:global_shap_e}
\end{minipage}
\\

\begin{minipage}{0.32\textwidth}
\centering
\includegraphics[width=\linewidth]{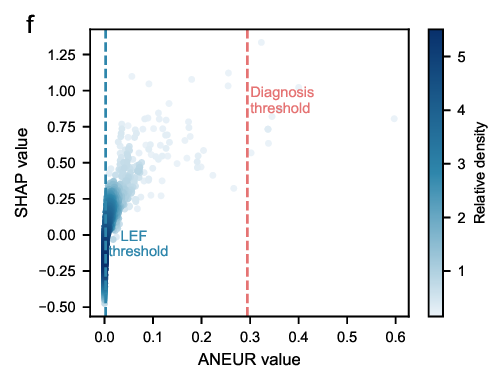}
\end{minipage}
\hfill
\begin{minipage}{0.32\textwidth}
\centering
\includegraphics[width=\linewidth]{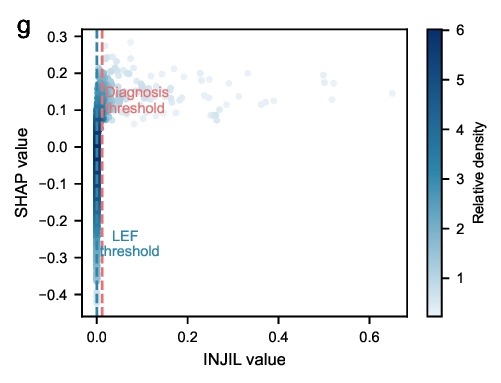}
\end{minipage}
\hfill
\begin{minipage}{0.32\textwidth}
\centering
\includegraphics[width=\linewidth]{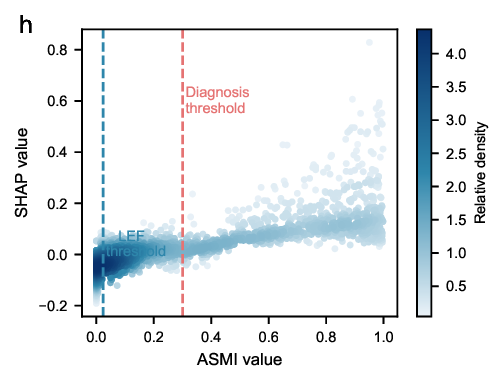}
\end{minipage}
\hfill

\caption{\textbf{Global-level explanation of ECG model predictions across all predictors.}
	The first row shows (a) the cumulative absolute SHAP contributions of all 71 predictors, expressed as percentages of the total contribution, and (b) a SHAP beeswarm plot for the top 10 predictors ranked by F1 score obtained from the single-predictor method. 
	The second and third rows show (c-h) the relationships between individual predictors (NORM, ILBBB, ISCLA, ANEUR, INJIL and ASMI) and their SHAP values. Point density was estimated using a Gaussian kernel density on the log-transformed values. Each plot includes two vertical reference lines: the first (LEF threshold) indicates the threshold that achieves the optimal F1 score using a single-predictor method, and the second (Diagnosis threshold) indicates the positivity threshold defined by an independent ECG diagnosis model.
}

	\label{fig:global_shap}
\end{figure}

\begin{figure}[!htbp]
	\centering
	\begin{minipage}{0.4\textwidth}
		\centering
		\includegraphics[width=\linewidth]{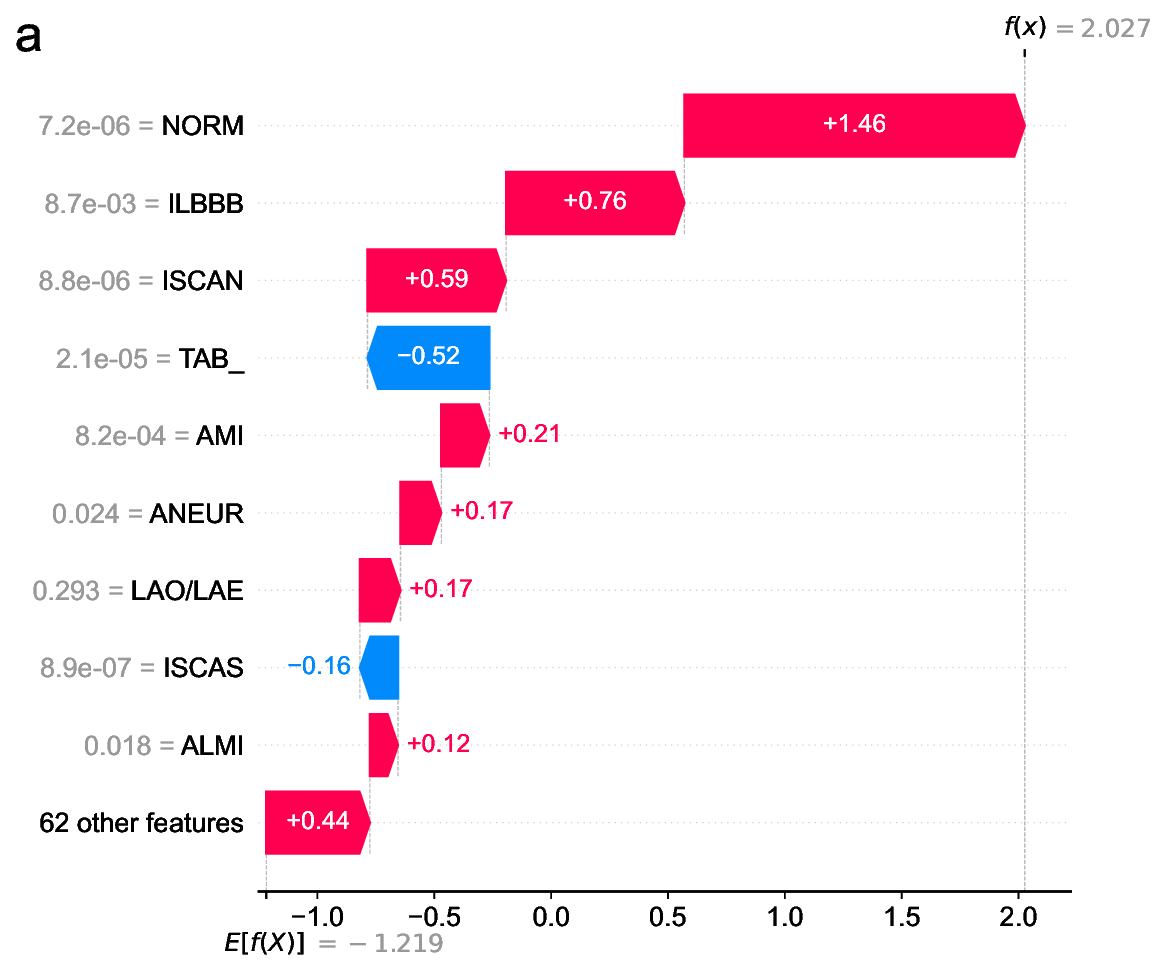}
	\end{minipage}
	\begin{minipage}{0.4\textwidth}
		\centering
		\includegraphics[width=\linewidth]{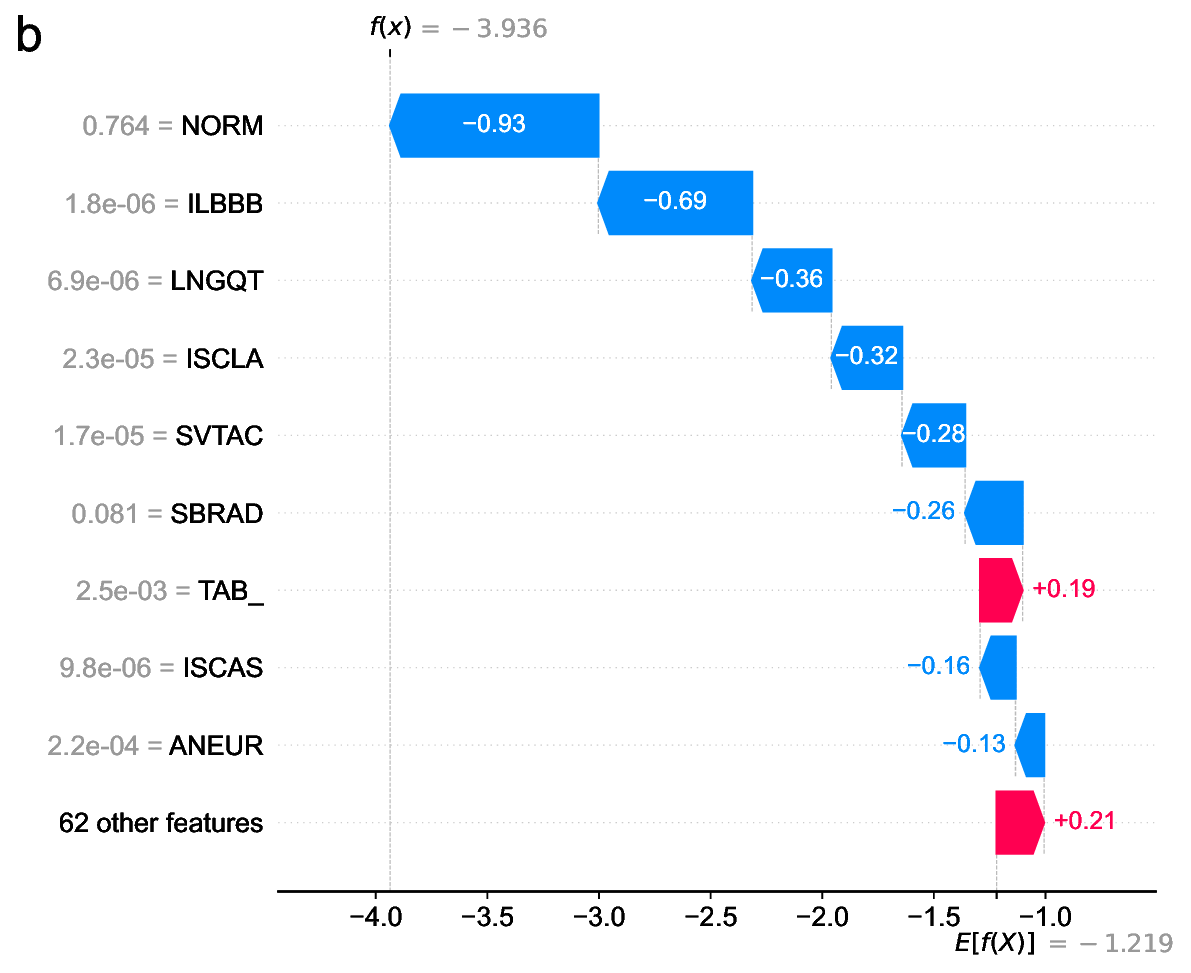}
	\end{minipage}

	\begin{minipage}{0.4\textwidth}
		\centering
		\includegraphics[width=\linewidth]{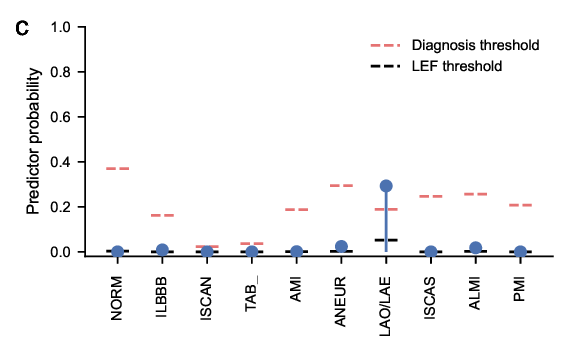}
	\end{minipage}
	\begin{minipage}{0.4\textwidth}
		\centering
		\includegraphics[width=\linewidth]{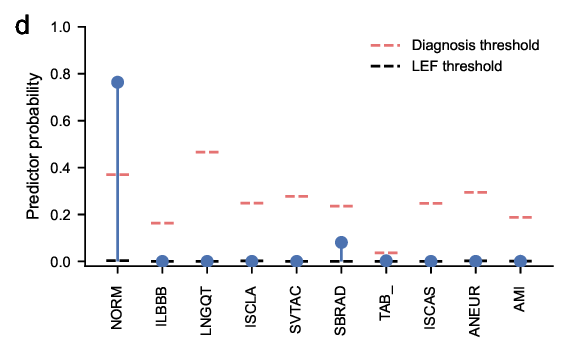}
	\end{minipage}

\caption{
	\textbf{Local-level explanation of model predictions for a positive and a negative case.} The first row shows SHAP waterfall plots for a positive case (a) and a negative case (b), illustrating the top ten ECG predictors contributing to the prediction, with the remaining predictors aggregated as ``62 other features''. Feature contributions are shown in the log-odds space and sum to the final model output. 
	The second row presents the corresponding predictor--probability relationships for the same cases (c,d). Vertical reference lines indicate positivity thresholds derived from an independent ECG diagnosis model.
}
	\label{fig:local_shap}
\end{figure}

The tabular component of ECGPD-LEF enables both global and local interpretability via SHAP analysis. We analyzed Transformer-PT-XGBoost on the internal test set (Figure \ref{fig:global_shap}, Figure \ref{fig:local_shap}). At the global level, cumulative SHAP contributions increased with the number of predictors included in the model (Figure \ref{fig:global_shap}a), consistent with the performance trends observed in Figure \ref{fig:performance} of Section \ref{subsec:multi-predictor}. The ranking of predictors by mean absolute SHAP value showed slight differences compared with the ranking based on single-predictor F1 scores, although the top contributors remained largely consistent (Figure~\ref{fig:global_shap}b). For example, NORM, ILBBB, INJAL, ISCLA, and ANEUR ranked 1-5 by single-predictor F1 score, whereas their SHAP-based ranking was 1, 2, 5, 3, and 4, respectively. In addition, SHAP values varied markedly at low predictor values (Figure \ref{fig:global_shap}c-g), often at magnitudes substantially below the diagnostic thresholds, which is consistent with the observation in Section \ref{sec:sub_sec_single_predictor}.

At the local level, explanation plots for a positive and a negative case illustrate individualized predictor contributions (Figure~\ref{fig:local_shap}). In these two examples, global importance patterns are reflected in case-specific prediction profiles, with NORM and ILBBB contributing the most. When combined with the single-predictor approach and corresponding diagnostic thresholds, the predictions for both cases are interpretable. In particular, the values of NORM and ILBBB exceeded the LEF-positive thresholds in the positive case and the LEF-negative thresholds in the negative case. Furthermore, the LAO/LAE predictor in the positive case and the NORM predictor in the negative case exceeded their standard diagnostic thresholds (Figures \ref{fig:local_shap}b-d), indicating that these patterns are potentially human-recognizable.

\subsection{Subgroup analysis}
\label{sec:subgroup}

Model performance was evaluated across clinically relevant subgroups for ECGPD-LEF. Across subgroups defined by age, sex, race/ethnicity, and clinical context, the performance of the single-predictor approach (NORM, ILBBB, INJIL, and ISCLA) in the internal and external test sets is presented in Section \ref{apedix:subgroup_single_predictor} of the Appendix. These predictors remained effective across both test sets, with NORM, ILBBB, and INJIL generally achieving higher performance.

For the multi-predictor approach, subgroup results for Transformer-PT-XGBoost in the internal test set are shown in Table \ref{table:subgroup}. Transformer-PT-XGBoost consistently outperformed the Columbia mini model. Relative improvements ranged from 2.0\%-12.2\% for AUROC, 3.5\%-32.3\% for AUPRC, and 5.9\%-30.4\% for F1 score across 16 subgroups. Improvements in AUPRC and F1 score exceeded 10\% in 15 and 12 subgroups, respectively. Similar trends were observed in the external test set (Table \ref{table:subgroup_lvefllm}).

Stratified analyses by the presence of structural heart disease (SHD) and valvular heart disease (VHD) are presented in Figures \ref{fig:shd_subgroup} and \ref{fig:vhd_subgroup} in the Appendix. In both settings, Transformer-PT-XGBoost demonstrated superior performance compared with the Columbia mini model. Notably, the performance of Transformer-PT-XGBoost remained stable, whereas the Columbia mini model showed reduced performance in subgroups with SHD and VHD.


%

\begin{table*}[!htbp]
\begin{ThreePartTable}
	\centering
	\footnotesize
	\setlength{\tabcolsep}{1pt}{
	\caption{
		Subgroup performance of ECGPD-LEF and the Columbia mini model in the internal test set.
		\label{table:subgroup}
	}
	\begin{tabular}{lcc|ccc | ccc}
		\toprule
		&  &  & \multicolumn{3}{c}{\textbf{Columbia mini model}} & \multicolumn{3}{c}{\textbf{ECGPD-LEF (Multi-predictor)}} \\
		\cmidrule(lr){4-6} \cmidrule(lr){7-9}
		\textbf{Subgroup} & \textbf{n} & \textbf{Prevalence (\%)} &
		\textbf{AUROC} & \textbf{AUPRC} & \textbf{F1 Score} &
		\textbf{AUROC} & \textbf{AUPRC} & \textbf{F1 Score} \\
		\midrule
		
		\multicolumn{9}{l}{\textbf{Age groups}} \\
		\midrule 
		\quad 18--59 & 2124 & 13.6 &
		\makecell{85.2 \\ (82.8--87.5)} &
		\makecell{53.5 \\ (48.1--60.0)} &
		\makecell{53.8 \\ (49.0--58.3)} &
		\makecell{\textbf{87.4} \\ (85.0--89.8)} &
		\makecell{\textbf{64.6} \\ (59.4--69.9)} &
		\makecell{\textbf{59.0} \\ (54.7--63.6)} \\
		
			\quad 60--69 & 1318 & 16.6 &
		\makecell{83.8 \\ (80.9--86.6)} &
		\makecell{57.6 \\ (51.1--64.6)} &
		\makecell{54.7 \\ (49.0--59.7)} &
		\makecell{\textbf{87.2} \\ (84.2--90.2)} &
		\makecell{\textbf{70.6} \\ (64.9--76.3)} &
		\makecell{\textbf{63.3} \\ (58.1--68.5)} \\
		
			\quad 70--79 & 1154 & 21.7 &
		\makecell{86.2 \\ (83.7--88.7)} &
		\makecell{65.8 \\ (59.5--72.2)} &
		\makecell{63.7 \\ (58.9--67.9)} &
		\makecell{\textbf{89.7} \\ (87.2--91.8)} &
		\makecell{\textbf{74.7} \\ (69.5--80.0)} &
		\makecell{\textbf{70.5} \\ (65.8--74.8)} \\
		
			\quad 80+ & 846 & 24.1 &
		\makecell{83.5 \\ (80.4--86.5)} &
		\makecell{65.4 \\ (59.5--71.8)} &
		\makecell{59.5 \\ (54.7--64.4)} &
		\makecell{\textbf{87.3} \\ (84.8--89.7)} &
		\makecell{\textbf{67.7} \\ (60.9--74.1)} &
		\makecell{\textbf{66.1} \\ (61.4--70.8)} \\
		
		\midrule
		\multicolumn{9}{l}{\textbf{Sex}} \\
		\midrule 
			\quad  Female & 2731 & 12.4 &
		\makecell{86.9 \\ (85.0--88.8)} &
		\makecell{52.3 \\ (46.9--58.0)} &
		\makecell{53.3 \\ (49.0--57.4)} &
		\makecell{\textbf{89.9} \\ (88.0--91.7)} &
		\makecell{\textbf{63.4} \\ (58.2--68.3)} &
		\makecell{\textbf{60.4} \\ (56.0--64.5)} \\
		
			\quad  Male & 2711 & 23.0 &
		\makecell{83.1 \\ (81.2--84.7)} &
		\makecell{64.0 \\ (59.9--68.0)} &
		\makecell{60.6 \\ (57.5--63.5)} &
		\makecell{\textbf{86.9} \\ (85.1--88.5)} &
		\makecell{\textbf{72.9} \\ (69.1--76.1)} &
		\makecell{\textbf{67.0} \\ (64.0--69.7)} \\
		
		\midrule
		\multicolumn{9}{l}{\textbf{Race / ethnicity}} \\
			\midrule 
			\quad Hispanic & 1649 & 16.7 &
		\makecell{86.9 \\ (84.5--89.1)} &
		\makecell{63.3 \\ (57.2--68.5)} &
		\makecell{60.0 \\ (55.3--64.5)} &
		\makecell{\textbf{89.5} \\ (87.3--91.7)} &
		\makecell{\textbf{71.7} \\ (66.0--76.6)} &
		\makecell{\textbf{65.2} \\ (60.1--69.3)} \\
		
			\quad White & 1569 & 16.3 &
		\makecell{84.0 \\ (81.1--86.6)} &
		\makecell{55.7 \\ (49.4--62.1)} &
		\makecell{55.1 \\ (50.2--60.0)} &
		\makecell{\textbf{88.6} \\ (86.4--90.8)} &
		\makecell{\textbf{64.8} \\ (58.7--71.1)} &
		\makecell{\textbf{63.6} \\ (59.0--68.0)} \\
		
			\quad  Black & 846 & 19.3 &
		\makecell{85.4 \\ (82.3--88.7)} &
		\makecell{65.5 \\ (58.5--72.4)} &
		\makecell{59.8 \\ (53.6--65.6)} &
		\makecell{\textbf{88.1} \\ (84.9--91.1)} &
		\makecell{\textbf{73.6} \\ (67.2--79.3)} &
		\makecell{\textbf{66.9} \\ (61.2--72.2)} \\
		
			\quad  Asian & 153 & 16.3 &
		\makecell{87.8 \\ (80.2--94.3)} &
		\makecell{61.4 \\ (44.5--81.4)} &
		\makecell{57.1 \\ (41.5--69.4)} &
		\makecell{\textbf{92.5} \\ (87.3--96.7)} &
		\makecell{\textbf{72.6} \\ (54.7--86.2)} &
		\makecell{\textbf{66.7} \\ (50.0--79.1)} \\
		
		\quad 	Other & 457 & 15.8 &
		\makecell{83.1 \\ (78.0--87.9)} &
		\makecell{52.1 \\ (40.7--64.0)} &
		\makecell{49.4 \\ (40.0--58.0)} &
		\makecell{\textbf{86.9} \\ (81.5--91.6)} &
		\makecell{\textbf{65.3} \\ (53.4--76.8)} &
		\makecell{\textbf{59.4} \\ (50.0--68.4)} \\
		
			\quad Unknown & 768 & 22.3 &
		\makecell{84.1 \\ (80.8--87.2)} &
		\makecell{61.5 \\ (53.9--69.1)} &
		\makecell{61.1 \\ (55.0--66.7)} &
		\makecell{\textbf{85.8} \\ (82.4--89.1)} &
		\makecell{\textbf{68.8} \\ (61.6--75.9)} &
		\makecell{\textbf{64.7} \\ (58.5--70.0)} \\
		
		\midrule
		\multicolumn{9}{l}{\textbf{Clinical context}} \\
		\midrule 
		\quad Emergency & 1971 & 15.7 &
		\makecell{86.0 \\ (83.6--88.2)} &
		\makecell{59.9 \\ (54.1--65.3)} &
		\makecell{58.5 \\ (53.7--62.8)} &
		\makecell{\textbf{87.8} \\ (85.5--89.9)} &
		\makecell{\textbf{67.1} \\ (61.7--72.2)} &
		\makecell{\textbf{62.0} \\ (57.5--65.8)} \\
		
		\quad 	Inpatient & 2203 & 24.4 &
		\makecell{81.9 \\ (79.7--84.0)} &
		\makecell{62.4 \\ (58.1--67.0)} &
		\makecell{59.1 \\ (55.9--62.5)} &
		\makecell{\textbf{86.1} \\ (84.1--88.1)} &
		\makecell{\textbf{71.8} \\ (67.7--75.7)} &
		\makecell{\textbf{66.8} \\ (63.3--69.8)} \\
		
		\quad 	Outpatient & 1059 & 6.6 &
		\makecell{87.3 \\ (82.0--91.5)} &
		\makecell{44.3 \\ (33.4--56.5)} &
		\makecell{48.4 \\ (38.4--57.0)} &
		\makecell{\textbf{90.6} \\ (85.9--94.5)} &
		\makecell{\textbf{54.6} \\ (42.8--65.4)} &
		\makecell{\textbf{56.4} \\ (46.4--64.9)} \\
		
	 	\quad 	Procedural & 209 & 21.5 &
		\makecell{79.5 \\ (71.6--85.9)} &
		\makecell{57.3 \\ (41.7--72.6)} &
		\makecell{52.3 \\ (37.8--63.8)} &
		\makecell{\textbf{89.2} \\ (83.0--94.0)} &
		\makecell{\textbf{75.8} \\ (62.8--86.3)} &
		\makecell{\textbf{68.2} \\ (54.5--78.2)} \\
		
		\bottomrule
	\end{tabular}
\begin{tablenotes}
	\footnotesize
	\item ECGPD-LEF is configured with the predictor extractor based on Transformer-PT and the tabular model XGBoost. This configuration was selected for illustration in the internal test set. Columbia mini model is the official benchmark on the same set \citep{poterucha2025detecting}. AUROC, AUPRC, and F1 are reported with 95\% confidence intervals.
\end{tablenotes}
}
\end{ThreePartTable}
\end{table*}


\section{Discussion}
\subsection{A structured paradigm for LEF detection}
In this study, we propose a structured predictor-integration framework (ECGPD-LEF) for ECG-based detection of reduced left ventricular function that departs from conventional end-to-end waveform modeling. The proper multi-predictor configuration (Transformer-PT-XGBoost) demonstrated robust performance in both the internal hold-out test set (AUROC 88.4\%, F1 64.5\%) and the external test set (AUROC 86.8\%, F1 53.6\%), significantly outperforming a recent strong end-to-end baseline, the Columbia mini model \citep{poterucha2025detecting} (internal AUROC 85.2\%, F1 57.9\%; external AUROC 80.7\%, F1 46.4\%). Performance gains were maintained across subgroups defined by age, sex, race/ethnicity, clinical context, and the presence or absence of structural or valvular heart disease, supporting the generalizability of the approach. Within this predictor-based architecture, several clinically recognizable ECG diagnoses, including NORM, ILBBB, and INJAL, were strongly associated with LEF detection (internal AUROC 75.3\%--81.0\%; external AUROC 71.6\%--78.6\%), enabling transparent interpretation of model outputs. As a lightweight extension built upon an existing ECG diagnosis model, this structured framework provides improved discrimination while preserving modularity and clinical interpretability.

\subsection{The need for publicly benchmarked AI-ECG evaluation}
Artificial intelligence applied to ECG has demonstrated strong performance in emerging tasks, including detection of aortic stenosis, tricuspid regurgitation and left ventricular dysfunction. However, most prior models have been developed and evaluated on institution-specific private datasets with heterogeneous population characteristics and outcome definitions, limiting reproducibility and preventing rigorous cross-model comparison. Early work by \cite{attia2019screening} trained a convolutional neural network on 44,959 patients to identify ventricular dysfunction defined as ejection fraction $\leq$ 35\%, with prospective validation performed at the same institution \citep{attia2019prospective}. Subsequent refinements extended detection to ejection fraction $\leq$ 40\% and reported multisite validation using digital ECG input alone~\citep{carter2026multisite}. Although these studies achieved strong discrimination, the absence of publicly accessible benchmarking datasets constrains transparent evaluation. Recently, \cite{poterucha2025detecting} addressed this gap by releasing a de-identified ECG dataset comprising 36,286 unique patients with predefined training, validation, and test splits, and by benchmarking a baseline model (the Columbia mini model) that achieved discrimination comparable to models trained on substantially larger proprietary cohorts. By providing standardized data partitions, model weights, and open-source code, this work established a transparent and reproducible evaluation framework for AI-ECG research, upon which our study enables rigorous and fair comparison. Furthermore, to extend validation beyond a single benchmarked dataset, we developed ECG-Note using publicly available datasets and large language models, establishing an additional external validation dataset for LEF detection that captures diverse patient populations and clinical contexts.

\subsection{Moving beyond end-to-end black-box modeling}
Most AI-ECG approaches for left ventricular dysfunction have relied on end-to-end deep learning architectures \citep{attia2019screening, attia2019prospective, carter2026multisite, poterucha2025detecting}. While such models achieve strong predictive performance, their limited interpretability poses challenges for clinical integration, particularly in applications that extend beyond conventional ECG criteria grounded in established ECG principles. In the absence of transparent mechanistic reasoning, black-box predictions may be difficult to reconcile with established diagnostic frameworks, potentially limiting clinician trust and adoption. Efforts to enhance transparency have included tabular models constructed from 555 discrete ECG measurements \citep{hughes2024simple}. Although more interpretable, these approaches depend on proprietary commercial measurement algorithms that may vary across ECG platforms \citep{strodthoff2023ptb}, thereby constraining generalizability and standardized external evaluation. In contrast, we introduce a structured representation paradigm that integrates the predictive capacity of deep learning with the interpretability of tabular modeling. Each predictor corresponds to clinically meaningful features and it does not rely on vendor-specific measurement pipelines. Within the publicly benchmarked setting, this approach demonstrates improved performance over the end-to-end Columbia mini model, supporting the feasibility of clinically interpretable yet high-performing AI-ECG systems.

\subsection{Interpretability reveals clinically meaningful indicators for LEF detection}
Interpretability analyses identified diagnostically informative predictors that provide mechanistic insight into LEF detection. In the single-predictor setting, continuous probability outputs from the trained traditional ECG diagnosis model were sufficient to achieve meaningful discrimination in a zero-shot-like manner. For example, the predicted probability of NORM alone yielded an internal AUROC of 81.0\% and an external AUROC of 78.6\%. Importantly, this does not imply that the binary diagnosis of NORM directly indicates LEF. Rather, the continuous probability output, well below the clinical decision threshold, captures graded deviations from normal ECG patterns that are strongly associated with LEF. This finding suggests that deep neural networks encode subclinical ECG variations within their probabilistic representations, even when such variations do not cross conventional diagnostic boundaries, offering a potential explanation for prior observations that large-scale deep learning models can detect novel cardiovascular phenotypes from subtle signal alterations. Within the multi-predictor framework, SHAP analyses demonstrated consistent importance patterns across predictors. At the population level (Figure \ref{fig:global_shap}), SHAP values varied substantially across probability ranges, indicating that graded shifts in ECG-derived diagnostic probabilities contribute meaningfully to LEF risk estimation. At the individual level (Figure \ref{fig:local_shap}), local explanations highlighted the predictors driving each decision and provided human-interpretable insights into the model’s reasoning (Figures \ref{fig:local_shap}c-d). Together, these findings suggest that LEF-associated ECG signatures may be decomposed into combinations of clinically recognizable diagnostic dimensions, enhancing the structural transparency of the proposed framework.

\subsection{Subgroup analysis demonstrates robustness across populations}
Subgroup analyses demonstrated that ECGPD-LEF maintained consistent discriminatory performance across diverse demographic and clinical strata, including age, sex, race/ethnicity, and care settings. Across all evaluated subgroups, ECGPD-LEF outperformed the Columbia Mini model in terms of AUROC, AUPRC, and F1 score. Notably, performance remained stable in patients with and without concomitant structural heart disease (SHD) or valvular heart disease (VHD), suggesting that the ECG signatures captured by ECGPD-LEF are not merely proxies for coexisting structural abnormalities but instead reflect signal components specifically associated with LEF. Despite limited representation of certain racial subgroups (e.g., Asian participants comprising 3.6\% of the training cohort), comparable performance was observed in both internal and external validation cohorts (AUROC 92.5\% and 91.0\%, respectively), supporting the generalizability and potential clinical applicability of the framework.

\subsection{Scalability and extensibility of the framework}
The proposed ECGPD-LEF framework exhibits scalability and extensibility across multiple dimensions. First, its modular design enables integration with existing ECG diagnostic models that are increasingly adopted in clinical practice, allowing lightweight extension of current AI-ECG systems without requiring full architectural replacement. Second, performance improvements observed with the Transformer-PT backbone (Table \ref{tab:method_config_comparison}) suggest that advances in upstream ECG diagnostic models may directly translate into enhanced LEF detection. As ECG diagnostic models continue to evolve with larger and higher-quality datasets, further gains may be anticipated. Third, the predictor-based structure is inherently expandable. In this study, we incorporated 71 diagnostic predictors derived from PTB-XL, and observed that model performance scaled with the number of predictors included in the tabular component (Figures \ref{fig:performance} and \ref{fig:global_shap}a). Additional clinically established ECG diagnostic features \citep{zhuanjiagongshi} may be incorporated within the same framework, enabling progressive refinement without fundamental architectural redesign. Together, these properties underscore the scalability and extensibility of the framework for future AI-ECG development.

\subsection{Limitations and future directions}
Despite the favorable performance and robustness of ECGPD-LEF compared with the latest deep learning baseline, several limitations warrant consideration. First, the multi-predictor framework was developed using labels derived from ECHO reports, which may be subject to inter-observer variability in ultrasound interpretation. Although external validation on the MIMIC-IV-Note cohort—comprising data from heterogeneous cardiac imaging sources—partially mitigates this concern, potential labeling inconsistencies cannot be fully excluded. Second, while both the EchoNext and MIMIC- IV-Note datasets include diverse populations, further validation in larger, international cohorts is warranted to ensure broad generalizability. Third, the predictor extractor was pretrained on a moderately sized ECG diagnosis dataset due to the limited availability of large-scale public ECG corpora. Leveraging larger, high-quality ECG diagnosis datasets may further enhance feature representation and downstream LEF detection performance. Future work should explore scaling strategies and prospective clinical validation to confirm real-world clinical applicability.

%

\section{Conclusion}
In summary, ECGPD-LEF provides a clinically interpretable, modular, and high-performing framework for ECG-based detection of LEF. By building upon existing ECG diagnostic models, it achieves superior performance compared with strong black-box baselines, while remaining lightweight and scalable. The structured predictor-based design enables transparent interpretation, revealing clinically meaningful indicators of LEF, such as probabilistic outputs from NORM and other predictors. By combining accuracy, stability, and interpretability, this framework provides a practical and scalable screening tool for real-world clinical applications.


\newpage

\appendix
\setcounter{table}{0} 
\setcounter{figure}{0}
\renewcommand{\thetable}{A.\arabic{table}}
\renewcommand{\thefigure}{A.\arabic{figure}}

\section{Supplementary  for the datasets}
\label{sec:appendix}
\subsection{Population characteristics (estimated predictors)}
\label{appendix:population_characteristics}

Table \ref{tab:num_echonext_2} summarizes the remaining population characteristics of the ECG-ECHO and ECG-Note datasets, as estimated by a traditional automatic ECG diagnosis model (Transformer-PT). It should be emphasized that these characteristics are derived from model-based predictions rather than direct annotations from the original data collection process. 
Based on these estimates, these datasets cover a broad range of diagnostic ECG predictors.
\begin{ThreePartTable}
\begin{TableNotes}
	\footnotesize
	\item Values are shown as counts and percentages. Counts are derived from automatic AI-ECG diagnoses generated by Transformer-PT. Predictor abbreviations are defined in \cite{wagner2020ptb}.
\end{TableNotes}
\begin{longtable}{l|c|c|c|c}
	\caption{Estimated patient characteristics across the training, validation, and test splits of the EchoNext dataset and the external cohort.}
	
	\label{tab:num_echonext_2} \\
	
	\toprule
	& \multicolumn{3}{c|}{\textbf{ECG-ECHO}} & \multicolumn{1}{c}{\textbf{ECG--Note}} \\
	\cmidrule(lr){2-4}\cmidrule(lr){5-5}
	& \textbf{Training set} & \textbf{Validation set} & \textbf{Test set} & \textbf{External Set} \\
	\midrule
	\multicolumn{5}{l}{\textbf{Estimated predictor}} \\
	\midrule
	\endfirsthead
	
	\toprule
	\multicolumn{5}{c}{\small \textbf{Table \thetable\ (continued)} }\\
	\toprule
	& \multicolumn{3}{c|}{\textbf{ECG-ECHO}} & \multicolumn{1}{c}{\textbf{ECG--Note}} \\
	\cmidrule(lr){2-4}\cmidrule(lr){5-5}
	& \textbf{Training set} & \textbf{Validation set} & \textbf{Internal test set} & \textbf{External test set} \\
	\midrule
	\multicolumn{5}{l}{\textbf{{Estimated predictor}}} \\
	\midrule
	\endhead
	
	\multicolumn{5}{r}{\small Continued on next page} \\
	\endfoot
	
	\bottomrule
	\insertTableNotes
	\endlastfoot
	
	\hspace{0.5em} NORM  & 19,259 (26.6\%) & 1,485 (32.1\%) & 1,806 (33.2\%) & 3,343 (20.9\%) \\
	\hspace{0.5em} ILBBB & 161 (0.2\%)    & 8 (0.2\%)     & 8 (0.1\%) & 74 (0.5\%) \\
	\hspace{0.5em} INJAL & 666 (0.9\%)    & 48 (1.0\%)    & 48 (0.9\%) & 108 (0.7\%) \\
	\hspace{0.5em} ISCLA & 109 (0.2\%)    & 3 (0.1\%)     & 6 (0.1\%) & 24 (0.1\%) \\
	\hspace{0.5em} ANEUR & 185 (0.3\%)    & 13 (0.3\%)    & 9 (0.2\%) & 27 (0.2\%) \\
	\hspace{0.5em} ISCAL & 941 (1.3\%)    & 50 (1.1\%)    & 59 (1.1\%) & 181 (1.1\%) \\
	\hspace{0.5em} ASMI  & 12,498 (17.2\%)& 641 (13.9\%)  & 759 (13.9\%) & 2,469 (15.4\%) \\
	\hspace{0.5em} SVARR & 690 (1.0\%)    & 47 (1.0\%)    & 57 (1.0\%) & 127 (0.8\%) \\
	\hspace{0.5em} INJIL & 2,173 (3.0\%)  & 125 (2.7\%)   & 171 (3.1\%) & 393 (2.5\%) \\
	\hspace{0.5em} CRBBB & 6,223 (8.6\%)  & 361 (7.8\%)   & 382 (7.0\%) & 1,147 (7.2\%) \\
	
	\hspace{0.5em} LAFB     & 9,713 (13.4\%)  & 590 (12.8\%)   & 701 (12.9\%) & 1,965 (12.3\%) \\
	\hspace{0.5em} ALMI     & 1,487 (2.1\%)   & 67 (1.4\%)     & 62 (1.1\%) & 256 (1.6\%) \\
	\hspace{0.5em} ABQRS    & 19,488 (26.9\%) & 1,074 (23.2\%) & 1,179 (21.7\%) & 3,267 (20.4\%) \\
	\hspace{0.5em} CLBBB    & 2,076 (2.9\%)   & 137 (3.0\%)    & 166 (3.1\%) & 506 (3.2\%) \\
	\hspace{0.5em} ILMI     & 1,536 (2.1\%)   & 95 (2.1\%)     & 105 (1.9\%) & 334 (2.1\%) \\
	\hspace{0.5em} INJAS    & 3,236 (4.5\%)   & 167 (3.6\%)    & 190 (3.5\%) & 637 (4.0\%) \\
	\hspace{0.5em} INVT     & 6,849 (9.5\%)   & 396 (8.6\%)    & 400 (7.4\%) & 1,268 (7.9\%) \\
	\hspace{0.5em} PVC      & 3,850 (5.3\%)   & 253 (5.5\%)    & 260 (4.8\%) & 864 (5.4\%) \\
	\hspace{0.5em} ISCIL    & 1,140 (1.6\%)   & 63 (1.4\%)     & 79 (1.5\%) & 180 (1.1\%) \\
	\hspace{0.5em} 1AVB     & 3,462 (4.8\%)   & 202 (4.4\%)    & 231 (4.2\%) & 640 (4.0\%) \\
	\hspace{0.5em} ISC\_    & 5,632 (7.8\%)   & 346 (7.5\%)    & 398 (7.3\%) & 1,675 (10.5\%) \\
	\hspace{0.5em} IVCD     & 3,062 (4.2\%)   & 173 (3.7\%)    & 204 (3.7\%) & 696 (4.3\%) \\
	\hspace{0.5em} LAO/LAE  & 6,886 (9.5\%)   & 375 (8.1\%)    & 460 (8.5\%) & 1,330 (8.3\%) \\
	\hspace{0.5em} ISCAN    & 3,160 (4.4\%)   & 154 (3.3\%)    & 153 (2.8\%) & 507 (3.2\%) \\
	\hspace{0.5em} ISCAS    & 1,109 (1.5\%)   & 62 (1.3\%)     & 51 (0.9\%) & 202 (1.3\%) \\
	\hspace{0.5em} AFIB     & 5,778 (8.0\%)   & 362 (7.8\%)    & 414 (7.6\%) & 1,158 (7.2\%) \\
	\hspace{0.5em} BIGU     & 430 (0.6\%)     & 26 (0.6\%)     & 34 (0.6\%) & 89 (0.6\%) \\
	\hspace{0.5em} SVTAC    & 173 (0.2\%)     & 14 (0.3\%)     & 12 (0.2\%) & 44 (0.3\%) \\
	\hspace{0.5em} AMI      & 758 (1.0\%)     & 43 (0.9\%)     & 54 (1.0\%) & 99 (0.6\%) \\
	\hspace{0.5em} NST\_    & 5,952 (8.2\%)   & 371 (8.0\%)    & 463 (8.5\%) & 1,241 (7.7\%) \\
	\hspace{0.5em} 3AVB     & 74 (0.1\%)      & 7 (0.2\%)      & 5 (0.1\%) & 21 (0.1\%) \\
	\hspace{0.5em} IMI      & 9,848 (13.6\%)  & 574 (12.4\%)   & 614 (11.3\%) & 1,679 (10.5\%) \\
	\hspace{0.5em} LPR      & 2,619 (3.6\%)   & 168 (3.6\%)    & 243 (4.5\%) & 607 (3.8\%) \\
	\hspace{0.5em} 2AVB     & 326 (0.4\%)     & 29 (0.6\%)     & 29 (0.5\%) & 83 (0.5\%) \\
	\hspace{0.5em} DIG      & 71 (0.1\%)      & 1 (0.0\%)      & 8 (0.1\%) & 9 (0.1\%) \\
	\hspace{0.5em} LMI      & 2,002 (2.8\%)   & 117 (2.5\%)    & 118 (2.2\%) & 343 (2.1\%) \\
	\hspace{0.5em} LOWT     & 2,415 (3.3\%)   & 167 (3.6\%)    & 211 (3.9\%) & 443 (2.8\%) \\
	\hspace{0.5em} SR       & 54,557 (75.3\%) & 3,546 (76.7\%) & 4,211 (77.4\%) & 11,166 (69.7\%) \\
	\hspace{0.5em} STACH    & 9,938 (13.7\%)  & 546 (11.8\%)   & 620 (11.4\%) & 1,642 (10.3\%) \\
	\hspace{0.5em} LPFB     & 3,138 (4.3\%)   & 160 (3.5\%)    & 159 (2.9\%) & 497 (3.1\%) \\
	\hspace{0.5em} PACE     & 154 (0.2\%)     & 3 (0.1\%)      & 7 (0.1\%) & 18 (0.1\%) \\
	\hspace{0.5em} WPW      & 57 (0.1\%)      & 0 (0.0\%)      & 3 (0.1\%) & 3 (0.0\%) \\
	\hspace{0.5em} ISCIN    & 2,244 (3.1\%)   & 141 (3.0\%)    & 155 (2.8\%) & 427 (2.7\%) \\
	\hspace{0.5em} PRC(S)   & 1,016 (1.4\%)   & 72 (1.6\%)     & 70 (1.3\%) & 219 (1.4\%) \\
	\hspace{0.5em} AFLT     & 45 (0.1\%)      & 1 (0.0\%)      & 5 (0.1\%) & 12 (0.1\%) \\
	\hspace{0.5em} INJIN    & 1,746 (2.4\%)   & 104 (2.2\%)    & 124 (2.3\%) & 395 (2.5\%) \\
	\hspace{0.5em} PAC      & 3,187 (4.4\%)   & 202 (4.4\%)    & 251 (4.6\%) & 733 (4.6\%) \\
	\hspace{0.5em} IPMI     & 10 (0.0\%)      & 1 (0.0\%)      & 0 (0.0\%) & 4 (0.0\%) \\
	\hspace{0.5em} STD\_    & 11,813 (16.3\%) & 756 (16.3\%)   & 840 (15.4\%) & 2,745 (17.1\%) \\
	
	\hspace{0.5em} LNGQT    & 993 (1.4\%)     & 43 (0.9\%)     & 62 (1.1\%) & 154 (1.0\%) \\
	\hspace{0.5em} TRIGU    & 1,564 (2.2\%)   & 100 (2.2\%)    & 117 (2.1\%) & 370 (2.3\%) \\
	\hspace{0.5em} NDT      & 5,067 (7.0\%)   & 333 (7.2\%)    & 383 (7.0\%) & 1,270 (7.9\%) \\
	\hspace{0.5em} LVH      & 19,728 (27.2\%) & 1,286 (27.8\%) & 1,483 (27.3\%) & 4,312 (26.9\%) \\
	\hspace{0.5em} PSVT     & 236 (0.3\%)     & 18 (0.4\%)     & 18 (0.3\%) & 49 (0.3\%) \\
	\hspace{0.5em} INJLA    & 4,521 (6.2\%)   & 268 (5.8\%)    & 271 (5.0\%) & 949 (5.9\%) \\
	\hspace{0.5em} PMI      & 4 (0.0\%)       & 0 (0.0\%)      & 1 (0.0\%) & 1 (0.0\%) \\
	\hspace{0.5em} STE\_    & 38,602 (53.3\%) & 2,595 (56.1\%) & 3,021 (55.5\%) & 7,752 (48.4\%) \\
	\hspace{0.5em} SEHYP    & 5,294 (7.3\%)   & 313 (6.8\%)    & 348 (6.4\%) & 1,017 (6.3\%) \\
	\hspace{0.5em} SBRAD    & 876 (1.2\%)     & 75 (1.6\%)     & 94 (1.7\%) & 255 (1.6\%) \\
	\hspace{0.5em} RAO/RAE  & 7,546 (10.4\%)  & 370 (8.0\%)    & 422 (7.8\%) & 1,601 (10.0\%) \\
	\hspace{0.5em} VCLVH    & 20,708 (28.6\%) & 1,384 (29.9\%) & 1,684 (30.9\%) & 4,715 (29.4\%) \\
	\hspace{0.5em} IRBBB    & 6,541 (9.0\%)   & 328 (7.1\%)    & 332 (6.1\%) & 1,037 (6.5\%) \\
	\hspace{0.5em} QWAVE    & 8,068 (11.1\%)  & 453 (9.8\%)    & 507 (9.3\%) & 1,590 (9.9\%) \\
	\hspace{0.5em} NT\_     & 2,052 (2.8\%)   & 150 (3.2\%)    & 177 (3.3\%) & 358 (2.2\%) \\
	\hspace{0.5em} EL       & 3,703 (5.1\%)   & 241 (5.2\%)    & 289 (5.3\%) & 825 (5.1\%) \\
	\hspace{0.5em} HVOLT    & 13,380 (18.5\%) & 1,009 (21.8\%) & 1,181 (21.7\%) & 3,329 (20.8\%) \\
	\hspace{0.5em} RVH      & 1,718 (2.4\%)   & 101 (2.2\%)    & 103 (1.9\%) & 295 (1.8\%) \\
	\hspace{0.5em} LVOLT    & 243 (0.3\%)     & 12 (0.3\%)     & 20 (0.4\%) & 79 (0.5\%) \\
	\hspace{0.5em} SARRH    & 1,298 (1.8\%)   & 112 (2.4\%)    & 131 (2.4\%) & 386 (2.4\%) \\
	\hspace{0.5em} IPLMI    & 159 (0.2\%)     & 8 (0.2\%)      & 11 (0.2\%) & 34 (0.2\%) \\
	\hspace{0.5em} TAB\_    & 86 (0.1\%)      & 11 (0.2\%)     & 5 (0.1\%) & 8 (0.0\%) \\
\end{longtable}
\end{ThreePartTable}

\subsection{Construction of the external test set}
\label{sec:mimic_lef_evaluation}
\begin{figure}[htbp]
	\centering
	\includegraphics[width=\linewidth]{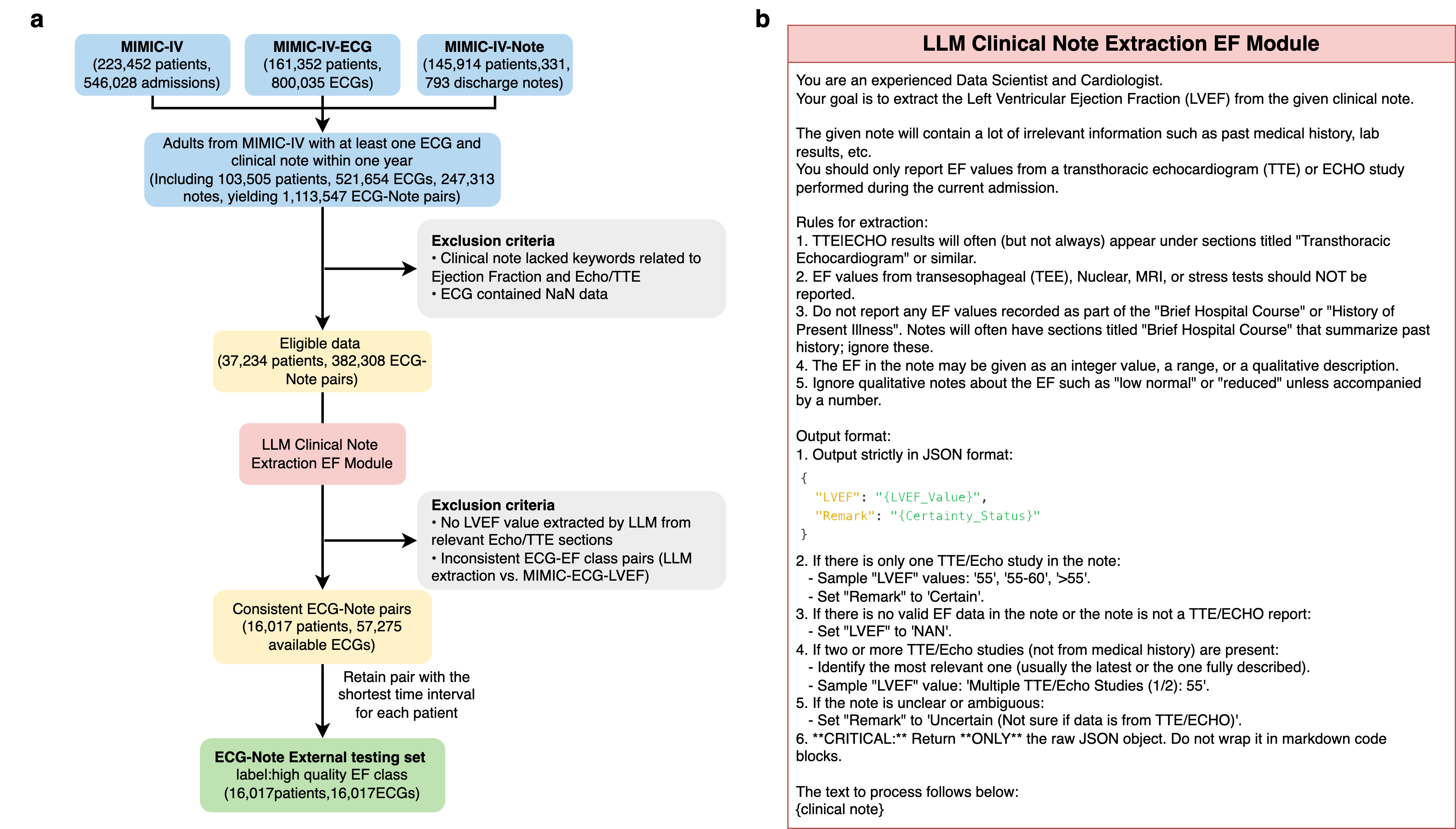}
	\caption{\textbf{Construction of the external test set.} \textbf{(a)} Flowchart illustrating the data selection process, including exclusion criteria applied to the MIMIC-IV database to derive the final high-quality testing set. \textbf{(b)} The prompt template used for the Large Language Model (LLM) Clinical Note Extraction Module, designed to extract quantitative ejection fraction(EF) values from unstructured clinical notes into a structured JSON format.}
	\label{fig:lvef_llm_workflow} 
\end{figure}

We constructed the external test set ECG-Note as Figure \ref{fig:lvef_llm_workflow}. Adult patients, who had at least one standard 10 s 12-lead ECG and at least one clinical note within one year following the ECG from MIMIC-IV \citep{PhysioNet-mimiciv-3.1}, MIMIC-IV-ECG \citep{PhysioNet-mimic-iv-ecg-1.0}, and MIMIC-IV-Note \citep{PhysioNet-mimic-iv-note-2.2}, were initially considered, leading to a starting pool of 1,113,547 ECG- note pairs from 103,505 patients with 521,654 ECGs and 247,313 notes. Exclusion criteria included: 1) ECG-Note pairs where the clinical note did not contain keywords related to ejection fraction(EF) and Echo/TTE; 2) ECG-Note pairs where the ECG contained NaN data. After exclusions, 382,308 ECG-Note paired data of 37,234 patients, comprising 246,543 ECGs and 62,023 notes, were included for further analysis. 
For the note set, we employed a Large Language Model (LLM), specifically Qwen2.5-72B-Instruct\citep{qwen2025qwen25technicalreport}, to extract precise ejection fraction(EF) values derived solely from the concurrent Echo/TTE (Fig.\ref{fig:lvef_llm_workflow} b). Based on the values obtained by the LLM, we determined whether the ejection fraction was $\le$ 45\%, suggesting moderately reduced systolic function. These labels were cross-referenced with MIMIC-ECG-LVEF \citep{li2025electrocardiogram} to identify consistent ECG-EF class pairs, resulting in the collection of 55,318 available ECGs. To ensure the accuracy of subsequent evaluations, only the ECG-Note pair with the shortest time interval for each patient was retained among the valid samples, resulting in a final testing set of 16,017 ECGs.
\subsection{Comparison with the Columbia mini model in the external test set}
\label{Apedix:external_comparison}
In the ECG-Note dataset, we compared the proposed ECGPD-EF with the Columbia mini model \citep{poterucha2025detecting}. ECGPD-EF requires only raw ECG signals for inference. In contrast, the Columbia mini model requires seven tabular features in addition to the ECG waveform: sex, age, PR interval, QRS duration, corrected QT interval (QTc), atrial rate, and ventricular rate. While age and sex were directly obtained from demographic records, the remaining five electrocardiographic metrics were not explicitly available and were derived using the \texttt{machine\_measurements} table from the MIMIC-IV ECG module. Specifically, let $T_{P_{onset}}$, $T_{QRS_{onset}}$, $T_{QRS_{end}}$, and $T_{T_{end}}$ denote the timings of P-wave onset, QRS onset, QRS offset, and T-wave offset, respectively, and let $RR$ denote the RR interval in milliseconds. The PR interval was calculated as $T_{QRS_{onset}} - T_{P_{onset}}$, and the QRS duration was derived as $T_{QRS_{end}} - T_{QRS_{onset}}$. The QTc interval was computed using Bazett's formula: $(T_{T_{end}} - T_{QRS_{onset}}) / \sqrt{RR/1000}$. The ventricular rate was calculated as $60,000 / RR$\citep{doi:10.1161/CIRCULATIONAHA.106.180200}. Due to the absence of specific atrial rate data, the derived ventricular rate was used as a proxy for the atrial rate.

\newpage
\setcounter{table}{0} 
\setcounter{figure}{0}
\renewcommand{\thetable}{B.\arabic{table}}
\renewcommand{\thefigure}{B.\arabic{figure}}
\section{Supplementary analyses for the multi-predictor approach}

\subsection{Subgroup analysis of the multi-predictor approach stratified by SHD and VHD}
\label{apedix:subgroup_multi_predictor}

\begin{figure}[!htbp]
	\centering
	\includegraphics[width=0.32\linewidth]{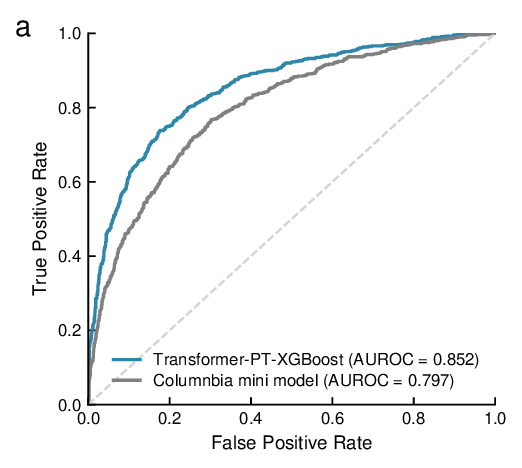}
	\includegraphics[width=0.32\linewidth]{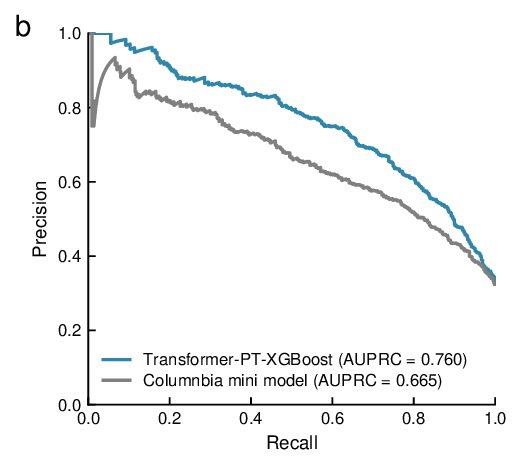}
	
	
	\includegraphics[width=0.32\linewidth]{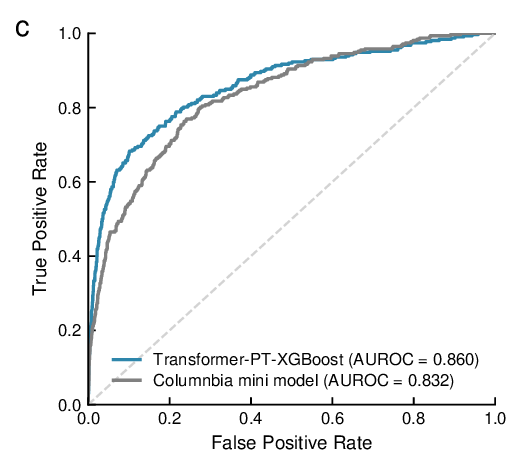}
	\includegraphics[width=0.32\linewidth]{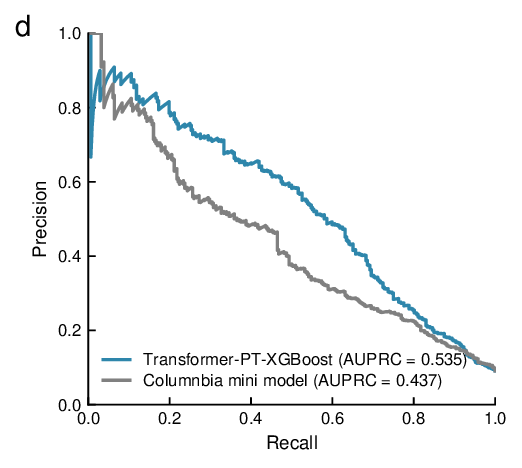}
	
	\caption{
		\textbf{Model performance stratified by other structural heart disease (SHD).}
		ROC (a,c) and PR (b,d) curves for individuals with other SHD (a,b) and without other SHD (c,d). 
		Other SHD was defined as the presence of at least one predefined echocardiographic abnormality independent of left ventricular ejection fraction (see Appendix \ref{apedix:subgroup_multi_predictor}). 
		Transformer-PT-XGBoost is compared with the Columbia mini model. 
		AUROC and AUPRC are indicated in each panel.
	}

	\label{fig:shd_subgroup}
\end{figure}

\begin{figure}[!htbp]
	\centering
	\includegraphics[width=0.32\linewidth]{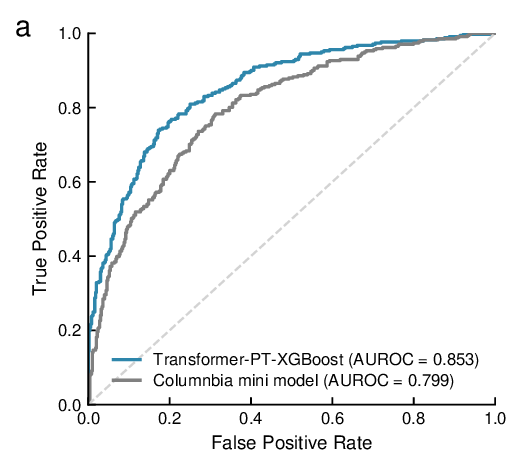}
	\includegraphics[width=0.32\linewidth]{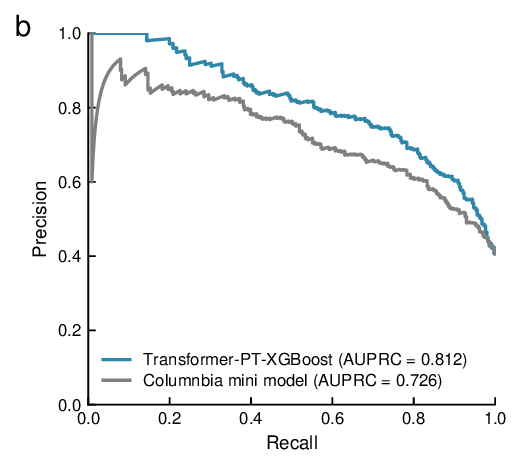}
	
	
	\includegraphics[width=0.32\linewidth]{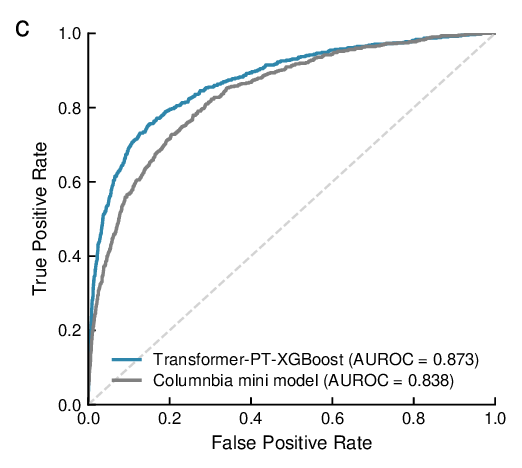}
	\includegraphics[width=0.32\linewidth]{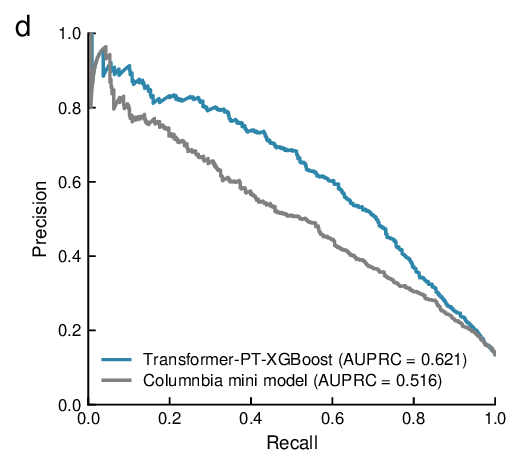}
	
	\caption{
		\textbf{Model performance stratified by valvular heart disease (VHD).}
		ROC (a,c) and PR (b,d) curves for individuals with moderate or greater VHD (a,b) and without moderate or greater VHD (c,d). 
		VHD was defined as moderate or greater valvular heart disease on echocardiography (see Appendix \ref{apedix:subgroup_multi_predictor}). 
		Transformer-PT-XGBoost is compared with the Columbia mini model. 
		AUROC and AUPRC are indicated in each panel.
	}
	
	\label{fig:vhd_subgroup}
\end{figure}

Baseline echocardiographic characteristics are presented in Table~\ref{tab:num_echonext_otherSHD}. Subgroup analysis results stratified by the presence of SHD and VHD are shown in Figures~\ref{fig:shd_subgroup} and \ref{fig:vhd_subgroup}.

\begin{table}[!htbp]
	\centering
\begin{threeparttable}
\caption{
	Baseline echocardiographic characteristics across the training, validation, and test splits of the ECG-ECHO dataset.
	\label{tab:num_echonext_otherSHD}
}
	\begin{tabular}{l|c|c|c}
		\toprule
		& \textbf{{Training set}} & \textbf{{Validation set}} & \textbf{{Test set}} \\
		\midrule
		\multicolumn{4}{l}{\textbf{Echocardiographic findings}} \\
		\midrule
		\hspace{0.5em} LVWT \(\geq\) 1.3cm & 17,667 (24.4\%) & 877 (19.0\%) & 1,061 (19.5\%) \\
		\hspace{0.5em} Aortic stenosis & 2,919 (4.0\%) & 252 (5.4\%) & 286 (5.3\%) \\
		\hspace{0.5em} Aortic regurgitation & 878 (1.2\%) & 62 (1.3\%) & 66 (1.2\%)  \\
		\hspace{0.5em} Mitral regurgitation & 6,137 (8.5\%) & 282 (6.1\%) & 337 (6.2\%)  \\
		\hspace{0.5em} Tricuspid regurgitation & 7,707 (10.6\%) & 305 (6.6\%) & 353 (6.5\%)  \\
		\hspace{0.5em} Pulmonary regurgitation & 603 (0.8\%) & 21 (0.5\%) & 20 (0.4\%)  \\
		\hspace{0.5em} RV systolic dysfunction & 9,597 (13.2\%) & 368 (8.0\%) & 419 (7.7\%)  \\
		\hspace{0.5em} Pericardial effusion & 2,079 (2.9\%) & 52 (1.1\%) & 69 (1.3\%)  \\
		\hspace{0.5em} PASP \(\geq\) 45mmHg & 13,727 (18.9\%) & 581 (12.6\%) & 699 (12.8\%)  \\
		\hspace{0.5em} TR Vmax \(\geq\) 3.2cm/s & 7,492 (10.3\%) & 267 (5.8\%) & 375 (6.9\%) \\
		\bottomrule
	\end{tabular}
\begin{tablenotes}
	\footnotesize
	\item The training, validation, and test splits correspond to the official splits of the ECG-ECHO dataset (EchoNext) \citep{elias2025echonext}. Values are shown as counts and percentages. LVWT, left ventricular wall thickness; RV, right ventricular; PASP, pulmonary artery systolic pressure.
\end{tablenotes}
\end{threeparttable}
\end{table}

\textbf{Other structural heart disease (SHD).} 
For subgroup analysis, individuals were classified as having other structural heart disease (SHD) if at least one of the following echocardiographic abnormalities was present: left ventricular hypertrophy (left ventricular wall thickness $\geq$ 1.3~cm); moderate or greater valvular heart disease (aortic, mitral, tricuspid, or pulmonary); moderate or greater right ventricular systolic dysfunction; moderate or large pericardial effusion; or pulmonary hypertension, defined as pulmonary artery systolic pressure (PASP) $\geq$ 45~mmHg or tricuspid regurgitation peak velocity (TR Vmax) $\geq$ 3.2~m/s. 

Individuals without any of the above findings were classified as having no other SHD. The performance comparison of ECGPD-LEF and the Columbia mini model in patients with and without other SHD is presented in Figure \ref{fig:shd_subgroup}.

\textbf{Valvular heart disease (VHD).} 
For the VHD subgroup analysis, individuals were categorized according to the presence of moderate or greater valvular heart disease on echocardiography, including aortic, mitral, tricuspid, or pulmonary valve disease. Those without moderate or greater valvular abnormalities were classified as non-VHD. The corresponding performance comparison between ECGPD-LEF and the Columbia mini model in VHD and non-VHD populations is shown in Figure \ref{fig:vhd_subgroup}.

\subsection{Subgroup analysis for multi-predictor method in the external test set}
To evaluate model robustness and fairness across patient characteristics and clinical settings, we conducted a subgroup analysis comparing our proposed Transformer-PT-XGBoost with Columbia mini model. Subgroups were defined by age, sex, race and ethnicity, and clinical context. Data linkage between the MIMIC-IV and MIMIC-IV-ECG datasets was performed to extract these subgroup attributes. Patient age was recalculated specifically at the time of ECG acquisition by leveraging the time-invariant interval method to restore temporal alignment \citep{johnson2023mimic}. For race and ethnicity, we aggregated granular categories into broader clusters (Table \ref{tab:race_grouped}). Regarding clinical context, we determined the admission type by mapping the ECG timestamp to the corresponding hospital admission and discharge window. All performance metrics are reported using the AUROC and the AUPRC (Table \ref{table:subgroup_lvefllm}).

Overall, our model demonstrates comparable or superior performance across almost all subgroups, suggesting that the proposed model generalizes well across heterogeneous patient populations and care environments, supporting its potential utility in real-world deployment scenarios.

\begin{table*}[!htbp]
	\centering
\begin{threeparttable}
	\footnotesize
	\setlength{\tabcolsep}{1pt}{
\caption{
	Subgroup performance of ECGPD-LEF and the Columbia mini model in the external test set. 
	\label{table:subgroup_lvefllm}
}
		\begin{tabular}{lcc|ccc|ccc}
			\toprule
			&  &  & \multicolumn{3}{c}{\textbf{Columbia mini model}} & \multicolumn{3}{c}{\textbf{ECGPD-LEF (Multi-predictor)}} \\
			\cmidrule(lr){4-6} \cmidrule(lr){7-9}
			\textbf{Subgroup} & \textbf{n} & \textbf{Prevalence (\%)} &
			\textbf{AUROC} & \textbf{AUPRC} & \textbf{F1 Score} &
			\textbf{AUROC} & \textbf{AUPRC} & \textbf{F1 Score} \\
			\midrule
			\multicolumn{9}{l}{\textbf{Age groups}} \\
			\midrule
			\quad 18--59 & 4270 & 12.2 & \makecell{83.0\\(80.9--84.9)} & \makecell{46.0\\(41.5--50.8)} & \makecell{49.1\\(45.4--52.4)} & \makecell{\textbf{89.5}\\(88.1--90.9)} & \makecell{\textbf{62.1}\\(57.4--66.4)} & \makecell{\textbf{57.3}\\(54.2--60.4)} \\
			\quad 60--69 & 3637 & 15.7 & \makecell{82.0\\(80.2--83.9)} & \makecell{47.7\\(43.7--52.6)} & \makecell{50.1\\(46.9--53.1)} & \makecell{\textbf{88.8}\\(87.3--90.3)} & \makecell{\textbf{65.1}\\(61.4--69.2)} & \makecell{\textbf{56.3}\\(53.4--59.3)} \\
			\quad 70--79 & 3761 & 18.0 & \makecell{81.1\\(79.2--82.8)} & \makecell{51.6\\(47.7--55.9)} & \makecell{50.6\\(48.0--53.3)} & \makecell{\textbf{85.5}\\(84.0--86.9)} & \makecell{\textbf{56.6}\\(53.0--60.9)} & \makecell{\textbf{54.4}\\(51.6--57.0)} \\
			\quad 80+ & 4349 & 17.2 & \makecell{77.9\\(76.1--79.7)} & \makecell{42.9\\(39.5--46.8)} & \makecell{43.5\\(40.9--45.9)} & \makecell{\textbf{83.5}\\(82.1--85.1)} & \makecell{\textbf{49.2}\\(45.6--53.2)} & \makecell{\textbf{49.6}\\(47.2--52.1)} \\
			\midrule
			\multicolumn{9}{l}{\textbf{Sex}} \\
			\midrule
			\quad Female & 7222 & 11.2 & \makecell{79.4\\(77.8--80.9)} & \makecell{32.5\\(29.5--36.0)} & \makecell{38.3\\(35.9--40.5)} & \makecell{\textbf{85.9}\\(84.6--87.2)} & \makecell{\textbf{45.6}\\(41.9--48.9)} & \makecell{\textbf{44.9}\\(42.6--47.1)} \\
			\quad Male & 8795 & 19.4 & \makecell{81.3\\(80.2--82.4)} & \makecell{52.7\\(50.2--55.5)} & \makecell{53.4\\(51.7--55.1)} & \makecell{\textbf{87.2}\\(86.3--88.1)} & \makecell{\textbf{63.9}\\(61.4--66.4)} & \makecell{\textbf{59.0}\\(57.3--60.6)} \\
			\midrule
			\multicolumn{9}{l}{\textbf{Race/ethnicity}} \\
			\midrule
			\quad Hispanic & 638 & 14.7 & \makecell{84.9\\(80.4--88.9)} & \makecell{53.6\\(43.3--64.1)} & \makecell{\textbf{57.1}\\(49.1--64.1)} & \makecell{\textbf{89.9}\\(86.9--92.6)} & \makecell{\textbf{64.7}\\(55.0--74.1)} & \makecell{53.5\\(46.4--60.4)} \\
			\quad White & 11688 & 15.5 & \makecell{80.3\\(79.2--81.3)} & \makecell{43.9\\(41.6--46.5)} & \makecell{46.5\\(44.9--48.2)} & \makecell{\textbf{86.9}\\(86.0--87.8)} & \makecell{\textbf{61.9}\\(59.4--64.4)} & \makecell{\textbf{56.9}\\(55.4--58.4)} \\
			\quad Black & 1845 & 17.3 & \makecell{84.7\\(82.3--86.9)} & \makecell{57.7\\(51.9--63.1)} & \makecell{54.1\\(50.1--58.1)} & \makecell{\textbf{89.8}\\(88.0--91.4)} & \makecell{\textbf{72.9}\\(67.5--78.0)} & \makecell{\textbf{62.1}\\(57.9--66.1)} \\
			\quad Asian & 414 & 12.6 & \makecell{84.6\\(79.2--89.1)} & \makecell{43.4\\(31.5--58.1)} & \makecell{47.7\\(37.7--56.1)} & \makecell{\textbf{90.7}\\(86.1--94.1)} & \makecell{\textbf{56.0}\\(40.5--72.6)} & \makecell{\textbf{55.3}\\(45.2--64.7)} \\
			\quad Other & 524 & 13.0 & \makecell{79.8\\(73.5--85.4)} & \makecell{39.5\\(29.3--51.9)} & \makecell{44.4\\(35.8--52.9)} & \makecell{\textbf{87.5}\\(82.7--91.6)} & \makecell{\textbf{53.5}\\(41.3--66.4)} & \makecell{\textbf{52.7}\\(44.4--60.9)} \\
			\quad Unknown & 908 & 18.8 & \makecell{75.9\\(71.6--79.6)} & \makecell{47.4\\(40.3--56.4)} & \makecell{45.6\\(40.0--51.5)} & \makecell{\textbf{81.3}\\(77.7--84.7)} & \makecell{\textbf{54.6}\\(45.6--65.2)} & \makecell{\textbf{51.9}\\(46.3--57.1)} \\
			\midrule
			\multicolumn{9}{l}{\textbf{Clinical context}} \\
			\midrule
			\quad Emergency & 9170 & 14.6 & \makecell{81.4\\(80.1--82.6)} & \makecell{44.3\\(41.8--47.3)} & \makecell{46.3\\(44.4--48.2)} & \makecell{\textbf{88.4}\\(87.4--89.4)} & \makecell{\textbf{61.0}\\(58.4--63.7)} & \makecell{\textbf{55.6}\\(53.7--57.2)} \\
			\quad Urgent & 3028 & 20.9 & \makecell{79.1\\(77.1--81.0)} & \makecell{50.7\\(46.7--54.9)} & \makecell{53.1\\(50.1--55.8)} & \makecell{\textbf{84.1}\\(82.6--85.7)} & \makecell{\textbf{60.2}\\(56.0--64.2)} & \makecell{\textbf{60.4}\\(57.7--62.7)} \\
			\quad Observation & 2173 & 19.6 & \makecell{83.9\\(81.8--85.9)} & \makecell{58.8\\(53.7--64.0)} & \makecell{56.3\\(52.8--59.9)} & \makecell{\textbf{89.4}\\(87.7--91.1)} & \makecell{\textbf{72.6}\\(67.7--77.3)} & \makecell{\textbf{63.6}\\(60.2--66.8)} \\
			\quad \makecell{Surgical \\ Same Day} & 1022 & 6.5 & \makecell{76.8\\(70.9--82.2)} & \makecell{22.1\\(15.1--33.2)} & \makecell{25.7\\(18.9--32.5)} & \makecell{\textbf{85.9}\\(81.3--89.8)} & \makecell{\textbf{26.8}\\(18.7--38.1)} & \makecell{\textbf{36.6}\\(27.5--44.6)} \\
			\quad Elective & 624 & 9.3 & \makecell{72.2\\(65.1--79.0)} & \makecell{21.0\\(14.7--32.0)} & \makecell{27.0\\(20.6--33.6)} & \makecell{\textbf{82.2}\\(76.8--87.1)} & \makecell{\textbf{34.3}\\(25.0--47.5)} & \makecell{\textbf{35.1}\\(27.9--42.8)} \\
			\bottomrule
		\end{tabular}
	}
	\begin{tablenotes}
		\footnotesize
		\item ECGPD-LEF is configured with the predictor extractor based on Transformer-PT and the tabular model XGBoost. This configuration was selected for illustration in the internal test set. Columbia mini model is the official benchmark on the same set \citep{poterucha2025detecting}. AUROC, AUPRC, and F1 are reported with 95\% confidence intervals.
	\end{tablenotes}
\end{threeparttable}
\end{table*}

\begin{table}[ht]
	\centering
	\small 
	\caption{Mapping of original race/ethnicity categories to aggregated subgroups.}
	\label{tab:race_grouped}
	\begin{tabular}{lp{0.55\linewidth}} 
		\toprule
		\textbf{Grouped} & \textbf{Original Categories} \\
		\midrule
		Asian & ASIAN, ASIAN - ASIAN INDIAN, ASIAN - CHINESE, ASIAN - KOREAN, ASIAN - SOUTH EAST ASIAN \\
		\midrule
		Black & BLACK/AFRICAN, BLACK/AFRICAN AMERICAN, BLACK/CAPE VERDEAN, BLACK/CARIBBEAN ISLAND \\
		\midrule
		Hispanic & HISPANIC OR LATINO, CENTRAL AMERICAN, COLUMBIAN, CUBAN, DOMINICAN, GUATEMALAN, HONDURAN, MEXICAN, PUERTO RICAN, SALVADORAN, SOUTH AMERICAN \\
		\midrule
		White & WHITE, WHITE - BRAZILIAN, EASTERN EUROPEAN, OTHER EUROPEAN, RUSSIAN, PORTUGUESE \\
		\midrule
		Other & OTHER, AMERICAN INDIAN/ALASKA NATIVE, MULTIPLE RACE/ETHNICITY, PACIFIC ISLANDER \\
		\midrule
		Unknown & UNKNOWN, UNABLE TO OBTAIN, DECLINED \\
		\bottomrule
	\end{tabular}
\end{table}

\clearpage
\setcounter{table}{0} 
\setcounter{figure}{0}
\renewcommand{\thetable}{C.\arabic{table}}
\renewcommand{\thefigure}{C.\arabic{figure}}

\section{Supplementary analyses for the single-predictor approach}

\subsection{Predictor recognition by the predictor extractor (remaining predictors)}
\label{Apedix:predictor_recog}

The classification performance of the predictor extractor for the remaining 61 predictors is reported in Table \ref{tab:results_single_predictor-2}. Across the 71 predictors, AUROC values ranged from 73.8\% to 100\%, with 58 exceeding 90\%, indicating consistently strong discriminative capacity. AUPRC and F1 scores varied substantially (0.3\%-97.6\% and 0.0\%-94.1\%), largely due to class imbalance. For instance, the INJIL category included only two positive cases in the PTB-XL test set (<0.1\% prevalence), resulting in unstable precision-recall estimates (test F1 = 0.0\%; validation F1 = 8.7\%) despite a high AUROC of 92.9\%.

Decision thresholds were set to maximize F1 on the validation set, ensuring all predictors achieved non-zero validation F1 scores. Since the downstream LEF framework leverages continuous predictor scores rather than threshold-dependent binary decisions, robust ranking is sufficient to reliably transfer information even for ultra-rare categories.

\subsection{Single-predictor performance (remaining predictors and external results)}

The performance of the remaining single predictors for LEF detection is reported in Table \ref{tab:results_single_predictor-2}. External test set performance is provided in Table \ref{tab:results_single_predictor-2-ecgnote}.

\begin{ThreePartTable}
	\begin{TableNotes}
		\footnotesize
		\item Predictor performance was obtained from the traditional automatic AI-ECG model. LEF detection performance was derived using the single-predictor approach proposed in this study. AUROC, AUPRC, and F1 are reported with 95\% confidence intervals. The two Thresh columns indicate the thresholds used to maximize the F1 score on the validation set for predictor performance and LEF detection, respectively. Predictor abbreviations are defined in \cite{wagner2020ptb}.
	\end{TableNotes}
\footnotesize 
\setlength{\tabcolsep}{1pt}
\begin{longtable}{l|c c c c|c c c c}
	\caption{Performance of traditional AI-ECG predictors and LEF detection using a single-predictor approach (remaining 61 predictors)} 
	\label{tab:results_single_predictor-2} \\
	
	\toprule
   \multirow{2.5}{*}{ \textbf{Predictor}} & \multicolumn{4}{c}{\textbf{Traditional ECG Model}} & \multicolumn{4}{c}{\textbf{LEF Detection (One Predictor)}} \\
	\cmidrule(lr){2-5} \cmidrule(lr){6-9}
	& \textbf{AUROC} & \textbf{AUPRC} & \textbf{F1 Score} & \textbf{Thresh} & \textbf{AUROC} & \textbf{AUPRC} & \textbf{F1 Score} & \textbf{Thresh} \\
	\midrule
	\endfirsthead
	
	\toprule
	\multicolumn{9}{c}{\small  \textbf{Table \thetable\ (continued)}} \\
	\cmidrule(lr){2-5} \cmidrule(lr){6-9}
	  \multirow{2.5}{*}{\textbf{Predictor}} & \multicolumn{4}{c}{\textbf{Traditional ECG Model}} & \multicolumn{4}{c}{\textbf{LEF Detection (One Predictor)}} \\
	\cmidrule(lr){2-5} \cmidrule(lr){6-9}
	&  \textbf{AUROC} &  \textbf{AUPRC} &  \textbf{F1 Score} &  \textbf{Thresh} &  \textbf{AUROC} &  \textbf{AUPRC} &  \textbf{F1 Score} &  \textbf{Thresh} \\
	\midrule
	\endhead
	
	\multicolumn{9}{r}{\small Continued on next page} \\
	\endfoot
	
	\bottomrule
	\insertTableNotes
	\endlastfoot
	
				LAFB  & \makecell{98.8\\(98.3--99.2)} & \makecell{87.0\\(82.1--91.2)} & \makecell{78.9\\(74.2--83.5)} & 0.274     & \makecell{70.0\\(68.2--71.7)} & \makecell{29.1\\(26.9--31.6)} & \makecell{39.0\\(36.8--41.1)} & 0.001226 \\
				ALMI  & \makecell{97.3\\(95.0--99.1)} & \makecell{61.3\\(43.6--77.1)} & \makecell{58.3\\(40.0--73.2)} & 0.257     & \makecell{68.6\\(66.8--70.6)} & \makecell{35.0\\(32.3--38.3)} & \makecell{38.6\\(36.1--41.1)} & 0.002106 \\
				ABQRS & \makecell{87.5\\(85.5--89.5)} & \makecell{57.4\\(51.7--63.5)} & \makecell{54.4\\(50.5--58.2)} & 0.053     & \makecell{68.3\\(66.5--70.1)} & \makecell{29.8\\(27.5--32.5)} & \makecell{38.0\\(35.9--40.1)} & 0.012123 \\
				CLBBB & \makecell{99.8\\(99.6--100.0)} & \makecell{93.0\\(85.9--98.4)} & \makecell{89.3\\(82.8--94.7)} & 0.053     & \makecell{67.9\\(66.0--69.7)} & \makecell{35.4\\(32.5--38.8)} & \makecell{38.0\\(35.5--40.2)} & 0.000006 \\
				ILMI  & \makecell{95.7\\(91.9--98.5)} & \makecell{63.4\\(50.6--75.7)} & \makecell{61.1\\(48.9--71.2)} & 0.157     & \makecell{67.6\\(65.7--69.4)} & \makecell{31.7\\(29.0--34.8)} & \makecell{37.9\\(35.9--40.1)} & 0.000172 \\
				INJAS & \makecell{99.1\\(98.6--99.5)} & \makecell{51.2\\(32.1--70.9)} & \makecell{44.4\\(25.8--60.7)} & 0.257     & \makecell{66.8\\(64.9--68.7)} & \makecell{27.0\\(25.1--29.2)} & \makecell{37.9\\(36.0--40.0)} & 0.000571 \\
				INVT  & \makecell{95.3\\(92.5--97.4)} & \makecell{27.1\\(14.4--45.4)} & \makecell{30.8\\(15.4--44.4)} & 0.269     & \makecell{66.6\\(64.7--68.6)} & \makecell{28.4\\(26.3--30.8)} & \makecell{37.8\\(35.8--39.9)} & 0.005779 \\
				PVC   & \makecell{99.3\\(98.9--99.6)} & \makecell{79.1\\(69.8--88.2)} & \makecell{85.3\\(80.4--89.8)} & 0.309     & \makecell{67.4\\(65.6--69.1)} & \makecell{29.7\\(27.3--32.4)} & \makecell{37.3\\(35.2--39.5)} & 0.001254 \\
				ISCIL & \makecell{95.8\\(93.4--97.8)} & \makecell{17.4\\( 7.6--36.7)} & \makecell{25.0\\(12.5--37.7)} & 0.053     & \makecell{66.0\\(64.3--67.6)} & \makecell{24.5\\(22.8--26.3)} & \makecell{37.3\\(35.5--39.4)} & 0.000142 \\
				1AVB  & \makecell{98.6\\(98.1--99.1)} & \makecell{68.9\\(57.4--80.4)} & \makecell{68.2\\(59.6--76.1)} & 0.226     & \makecell{67.8\\(65.8--69.5)} & \makecell{28.8\\(26.5--31.3)} & \makecell{37.2\\(35.1--39.1)} & 0.000059 \\
				ISC\_  & \makecell{96.1\\(94.4--97.4)} & \makecell{67.1\\(58.7--74.7)} & \makecell{59.3\\(52.3--66.7)} & 0.516     & \makecell{66.8\\(64.8--68.6)} & \makecell{29.9\\(27.4--32.5)} & \makecell{37.0\\(34.9--39.1)} & 0.007637 \\
				IVCD  & \makecell{80.3\\(74.4--84.8)} & \makecell{20.6\\(13.7--30.4)} & \makecell{29.3\\(19.8--38.2)} & 0.209     & \makecell{66.2\\(64.1--68.0)} & \makecell{32.4\\(29.5--35.3)} & \makecell{36.7\\(34.2--39.1)} & 0.027054 \\
				LAO/LAE & \makecell{88.3\\(83.6--92.6)} & \makecell{16.0\\(10.0--28.2)} & \makecell{26.5\\(14.1--37.6)} & 0.189     & \makecell{64.1\\(61.9--66.3)} & \makecell{32.9\\(30.0--36.0)} & \makecell{36.5\\(34.0--39.0)} & 0.051849 \\
				ISCAN & \makecell{93.8\\(91.3--96.7)} & \makecell{ 1.7\\( 0.6-- 4.6)} & \makecell{ 0.0\\( 0.0-- 0.0)} & 0.024     & \makecell{64.7\\(62.8--66.6)} & \makecell{26.2\\(24.2--28.7)} & \makecell{35.8\\(33.7--37.9)} & 0.000312 \\
				ISCAS & \makecell{96.7\\(94.9--98.1)} & \makecell{21.3\\( 8.4--41.1)} & \makecell{20.5\\( 5.1--37.5)} & 0.247     & \makecell{62.4\\(60.7--64.2)} & \makecell{21.6\\(20.2--23.2)} & \makecell{35.5\\(33.6--37.2)} & 0.000066 \\
				AFIB  & \makecell{98.6\\(97.4--99.6)} & \makecell{93.4\\(89.6--96.4)} & \makecell{91.4\\(87.8--94.3)} & 0.134     & \makecell{64.4\\(62.5--66.3)} & \makecell{27.5\\(25.2--30.1)} & \makecell{35.4\\(33.4--37.5)} & 0.000004 \\
				BIGU  & \makecell{97.1\\(92.3--99.8)} & \makecell{37.5\\( 9.7--73.1)} & \makecell{42.1\\(11.8--66.7)} & 0.143     & \makecell{63.2\\(61.4--65.0)} & \makecell{23.9\\(22.1--26.0)} & \makecell{34.9\\(32.9--37.0)} & 0.000013 \\
				SVTAC & \makecell{98.6\\(96.5--100.0)} & \makecell{16.1\\( 1.4--75.0)} & \makecell{25.0\\( 0.0--66.7)} & 0.277     & \makecell{63.6\\(61.7--65.4)} & \makecell{26.5\\(24.4--29.0)} & \makecell{34.4\\(32.2--36.6)} & 0.000113 \\
				AMI   & \makecell{92.3\\(88.7--95.4)} & \makecell{31.4\\(17.0--48.7)} & \makecell{38.8\\(23.5--52.6)} & 0.188     & \makecell{61.3\\(59.6--62.9)} & \makecell{22.2\\(20.5--24.1)} & \makecell{34.4\\(32.7--36.0)} & 0.000868 \\

					NST\_ & \makecell{86.2\\(82.6--89.7)} & \makecell{19.4\\(13.6--28.6)} & \makecell{25.9\\(18.6--33.0)} & 0.171
				& \makecell{60.2\\(58.4--61.9)} & \makecell{20.7\\(19.2--22.5)} & \makecell{34.4\\(32.7--36.1)} & 0.005043 \\
				
				3AVB & \makecell{99.6\\(99.0--100.0)} & \makecell{55.6\\(4.8--100.0)} & \makecell{50.0\\(0.0--100.0)} & 0.254
				& \makecell{61.0\\(59.0--63.1)} & \makecell{24.8\\(22.9--27.1)} & \makecell{34.2\\(32.4--36.1)} & <0.000001 \\
				
				IMI & \makecell{94.8\\(93.8--95.8)} & \makecell{73.2\\(67.8--78.2)} & \makecell{66.1\\(61.7--70.5)} & 0.218
				& \makecell{63.1\\(61.2--65.0)} & \makecell{25.6\\(23.5--28.1)} & \makecell{34.2\\(32.3--36.2)} & 0.003210 \\
				
				LPR & \makecell{98.4\\(97.5--99.1)} & \makecell{51.4\\(35.8--69.4)} & \makecell{47.6\\(33.3--60.5)} & 0.260
				& \makecell{61.6\\(59.7--63.6)} & \makecell{23.2\\(21.5--25.4)} & \makecell{33.7\\(31.7--35.6)} & 0.000049 \\
				
				2AVB & \makecell{99.6\\(99.4--99.9)} & \makecell{11.1\\(7.1--40.0)} & \makecell{0.0\\(0.0--0.0)} & 0.028
				& \makecell{61.2\\(59.3--63.0)} & \makecell{24.0\\(22.2--26.1)} & \makecell{33.3\\(31.4--35.2)} & 0.000002 \\
				
				DIG & \makecell{93.6\\(90.1--96.5)} & \makecell{8.6\\(4.4--19.3)} & \makecell{7.4\\(0.0--22.2)} & 0.433
				& \makecell{60.6\\(58.7--62.4)} & \makecell{22.2\\(20.6--24.1)} & \makecell{33.3\\(31.5--35.1)} & 0.000024 \\
				
				LMI & \makecell{93.6\\(90.6--96.3)} & \makecell{10.6\\(5.6--21.3)} & \makecell{20.8\\(5.3--35.7)} & 0.176
				& \makecell{60.8\\(58.9--62.8)} & \makecell{25.1\\(23.1--27.5)} & \makecell{33.0\\(31.2--34.9)} & 0.000937 \\
				
				LOWT & \makecell{91.8\\(89.8--93.7)} & \makecell{12.5\\(8.7--19.5)} & \makecell{15.4\\(7.1--25.0)} & 0.225
				& \makecell{58.1\\(56.2--59.8)} & \makecell{20.3\\(18.8--21.9)} & \makecell{32.9\\(31.3--34.8)} & 0.000556 \\
				
				SR & \makecell{92.7\\(91.2--94.1)} & \makecell{96.7\\(95.7--97.6)} & \makecell{94.1\\(93.3--94.9)} & 0.277
				& \makecell{62.3\\(60.3--64.3)} & \makecell{25.6\\(23.4--28.2)} & \makecell{32.7\\(30.8--34.4)} & 0.943848 \\
				
				STACH & \makecell{99.4\\(98.8--99.8)} & \makecell{86.5\\(77.1--94.4)} & \makecell{85.9\\(80.2--91.0)} & 0.231
				& \makecell{60.2\\(58.2--62.0)} & \makecell{22.0\\(20.3--23.9)} & \makecell{32.5\\(30.9--34.2)} & 0.000001 \\
				
				LPFB & \makecell{98.6\\(97.7--99.4)} & \makecell{48.3\\(26.4--69.2)} & \makecell{43.9\\(22.8--61.9)} & 0.242
				& \makecell{59.8\\(58.0--61.8)} & \makecell{23.1\\(21.3--25.2)} & \makecell{32.4\\(30.6--34.4)} & 0.000076 \\
				
				PACE & \makecell{98.1\\(95.4--100.0)} & \makecell{87.4\\(73.5--96.9)} & \makecell{82.4\\(69.2--92.6)} & 0.425
				& \makecell{58.7\\(56.8--60.5)} & \makecell{22.4\\(20.7--24.5)} & \makecell{31.9\\(30.1--33.6)} & 0.000033 \\
				
				WPW & \makecell{95.0\\(84.6--100.0)} & \makecell{62.2\\(24.3--97.6)} & \makecell{66.7\\(17.7--93.4)} & 0.695
				& \makecell{60.1\\(58.2--62.1)} & \makecell{24.0\\(22.1--26.2)} & \makecell{31.9\\(29.8--33.9)} & 0.000028 \\
				
				ISCIN & \makecell{94.3\\(90.0--97.2)} & \makecell{23.8\\(8.6--41.1)} & \makecell{27.9\\(15.0--40.4)} & 0.060
				& \makecell{58.4\\(56.6--60.2)} & \makecell{20.8\\(19.4--22.4)} & \makecell{31.8\\(29.9--33.8)} & 0.000432 \\
				
				PRC(S) & \makecell{99.8\\(99.5--100.0)} & \makecell{16.7\\(9.1--57.1)} & \makecell{11.8\\(8.3--38.1)} & 0.009
				& \makecell{58.6\\(56.8--60.5)} & \makecell{21.7\\(20.1--23.7)} & \makecell{31.7\\(29.6--33.6)} & 0.000008 \\
				
				AFLT & \makecell{86.5\\(53.9--100.0)} & \makecell{61.5\\(19.1--97.3)} & \makecell{44.4\\(0.0--83.4)} & 0.950
				& \makecell{58.0\\(56.1--60.0)} & \makecell{22.9\\(21.0--25.1)} & \makecell{31.5\\(29.9--33.2)} & 0.000010 \\
				
				INJIN & \makecell{99.9\\(99.6--100.0)} & \makecell{64.3\\(12.5--100.0)} & \makecell{11.8\\(5.0--27.0)} & 0.006
				& \makecell{58.9\\(57.0--60.8)} & \makecell{22.9\\(21.0--25.1)} & \makecell{31.2\\(29.0--33.3)} & 0.000063 \\
				
				PAC & \makecell{98.1\\(97.2--98.8)} & \makecell{48.6\\(34.3--67.8)} & \makecell{51.9\\(40.4--62.6)} & 0.298
				& \makecell{56.9\\(54.8--58.7)} & \makecell{21.0\\(19.4--22.8)} & \makecell{31.1\\(29.1--33.0)} & 0.000290 \\
				
				IPMI & \makecell{98.6\\(97.3--99.9)} & \makecell{10.2\\(1.7--47.4)} & \makecell{0.0\\(0.0--0.0)} & 0.287
				& \makecell{57.9\\(55.9--59.9)} & \makecell{22.8\\(20.9--25.1)} & \makecell{31.1\\(29.4--32.8)} & 0.000016 \\
				
				STD\_ & \makecell{89.6\\(86.6--92.1)} & \makecell{26.1\\(19.9--34.5)} & \makecell{39.3\\(32.5--46.2)} & 0.178
				& \makecell{58.0\\(55.9--60.1)} & \makecell{22.4\\(20.7--24.4)} & \makecell{30.8\\(29.2--32.4)} & 0.004936 \\

		LNGQT & \makecell{97.5\\(95.0--99.2)} & \makecell{20.4\\(7.6--45.8)} & \makecell{30.0\\(0.0--54.6)} & 0.466
		& \makecell{54.0\\(52.0--55.9)} & \makecell{19.5\\(18.0--21.3)} & \makecell{30.4\\(28.8--31.9)} & 0.000042 \\
		
		TRIGU & \makecell{99.3\\(98.3--100.0)} & \makecell{28.2\\(2.9--100.0)} & \makecell{9.1\\(3.9--22.9)} & 0.004
		& \makecell{55.5\\(53.5--57.6)} & \makecell{21.0\\(19.4--22.8)} & \makecell{30.2\\(28.3--32.1)} & 0.000002 \\
		
		NDT & \makecell{93.8\\(92.5--95.0)} & \makecell{58.0\\(50.7--65.3)} & \makecell{57.9\\(52.0--63.4)} & 0.328
		& \makecell{42.2\\(40.1--44.1)} & \makecell{14.9\\(13.8--16.2)} & \makecell{30.1\\(28.7--31.5)} & <0.000001 \\
		
		LVH & \makecell{93.6\\(92.0--95.1)} & \makecell{69.2\\(63.1--74.9)} & \makecell{63.6\\(58.7--68.5)} & 0.267
		& \makecell{55.2\\(52.9--57.3)} & \makecell{22.9\\(21.0--25.2)} & \makecell{30.0\\(28.7--31.5)} & 0.000005 \\
		
		PSVT & \makecell{99.9\\(99.6--100.0)} & \makecell{64.3\\(12.5--100.0)} & \makecell{33.3\\(0.0--80.0)} & 0.160
		& \makecell{52.3\\(50.3--54.4)} & \makecell{19.4\\(18.0--21.2)} & \makecell{30.0\\(28.7--31.5)} & <0.000001 \\
		
		INJLA & \makecell{73.8\\(63.8--83.3)} & \makecell{0.3\\(0.1--0.9)} & \makecell{0.0\\(0.0--0.0)} & 0.005
		& \makecell{54.9\\(52.8--57.0)} & \makecell{20.0\\(18.6--21.9)} & \makecell{30.0\\(28.2--32.0)} & 0.000020 \\
		
		PMI & \makecell{89.8\\(86.3--93.0)} & \makecell{0.6\\(0.3--2.1)} & \makecell{0.0\\(0.0--0.0)} & 0.208
		& \makecell{46.5\\(44.5--48.4)} & \makecell{16.2\\(15.1--17.5)} & \makecell{30.0\\(28.7--31.5)} & <0.000001 \\
		
		STE\_ & \makecell{94.0\\(84.4--99.4)} & \makecell{3.7\\(0.3--15.4)} & \makecell{0.4\\(0.1--1.0)} & <0.001
		& \makecell{45.8\\(43.9--47.8)} & \makecell{16.2\\(15.0--17.5)} & \makecell{30.0\\(28.7--31.5)} & <0.000001 \\
		
		SEHYP & \makecell{100.0\\(99.8--100.0)} & \makecell{83.3\\(25.0--100.0)} & \makecell{28.6\\(9.5--54.5)} & 0.019
		& \makecell{49.2\\(47.2--51.3)} & \makecell{18.5\\(17.1--20.2)} & \makecell{30.0\\(28.7--31.5)} & <0.000001 \\
		
		SBRAD & \makecell{96.3\\(93.8--98.1)} & \makecell{58.7\\(45.3--70.3)} & \makecell{60.3\\(49.5--70.1)} & 0.236
		& \makecell{41.7\\(39.9--43.7)} & \makecell{14.4\\(13.4--15.6)} & \makecell{30.0\\(28.6--31.4)} & <0.000001 \\
		
		RAO/RAE & \makecell{97.3\\(95.1--99.1)} & \makecell{41.4\\(7.8--70.4)} & \makecell{27.8\\(7.1--46.7)} & 0.094
		& \makecell{49.6\\(47.5--51.6)} & \makecell{19.3\\(17.7--21.3)} & \makecell{30.0\\(28.7--31.5)} & <0.000001 \\
		
		VCLVH & \makecell{86.4\\(82.6--90.1)} & \makecell{29.1\\(21.3--40.2)} & \makecell{34.3\\(27.5--41.9)} & 0.122
		& \makecell{42.5\\(40.4--44.6)} & \makecell{15.3\\(14.3--16.7)} & \makecell{30.0\\(28.7--31.5)} & 0.000012 \\
		
		IRBBB & \makecell{98.3\\(97.6--99.0)} & \makecell{78.8\\(71.3--85.2)} & \makecell{61.4\\(52.4--69.1)} & 0.775
		& \makecell{47.7\\(45.8--49.6)} & \makecell{16.1\\(15.1--17.3)} & \makecell{30.0\\(28.7--31.5)} & <0.000001 \\
		
		QWAVE & \makecell{87.5\\(82.8--91.5)} & \makecell{21.2\\(13.0--32.2)} & \makecell{24.5\\(15.4--33.8)} & 0.165
		& \makecell{54.2\\(52.1--56.4)} & \makecell{22.0\\(20.1--24.3)} & \makecell{30.0\\(28.7--31.5)} & 0.000019 \\
		
		NT\_ & \makecell{95.0\\(91.9--97.3)} & \makecell{33.1\\(22.7--49.9)} & \makecell{35.4\\(20.0--48.7)} & 0.276
		& \makecell{41.6\\(39.5--43.7)} & \makecell{14.7\\(13.6--15.9)} & \makecell{30.0\\(28.7--31.5)} & <0.000001 \\
		
		EL & \makecell{93.9\\(88.5--97.6)} & \makecell{5.4\\(2.1--13.6)} & \makecell{9.8\\(0.0--21.7)} & 0.095
		& \makecell{46.4\\(44.5--48.4)} & \makecell{16.4\\(15.1--17.7)} & \makecell{30.0\\(28.6--31.5)} & 0.000001 \\
		
		HVOLT & \makecell{89.2\\(73.8--98.3)} & \makecell{4.6\\(0.7--17.1)} & \makecell{6.3\\(1.6--12.6)} & 0.010
		& \makecell{32.6\\(30.7--34.5)} & \makecell{12.5\\(11.7--13.5)} & \makecell{30.0\\(28.7--31.5)} & <0.000001 \\
		
		RVH & \makecell{95.1\\(89.2--99.1)} & \makecell{26.2\\(8.7--52.5)} & \makecell{28.6\\(7.4--50.0)} & 0.415
		& \makecell{50.9\\(48.9--52.9)} & \makecell{17.8\\(16.6--19.4)} & \makecell{30.0\\(28.7--31.5)} & <0.000001 \\
		
		LVOLT & \makecell{90.9\\(83.7--96.2)} & \makecell{9.6\\(4.0--21.1)} & \makecell{19.0\\(6.2--32.1)} & 0.127
		& \makecell{48.6\\(46.5--50.6)} & \makecell{17.2\\(15.8--18.8)} & \makecell{30.0\\(28.7--31.5)} & <0.000001 \\
		
		SARRH & \makecell{97.5\\(96.5--98.3)} & \makecell{63.4\\(52.1--73.8)} & \makecell{56.8\\(48.0--65.0)} & 0.377
		& \makecell{48.4\\(46.4--50.4)} & \makecell{17.6\\(16.2--18.9)} & \makecell{29.8\\(28.3--31.2)} & 0.000012 \\
		
		IPLMI & \makecell{82.0\\(58.8--98.9)} & \makecell{4.4\\(0.2--23.3)} & \makecell{0.0\\(0.0--0.0)} & 0.128
		& \makecell{52.2\\(50.2--54.2)} & \makecell{19.6\\(18.0--21.5)} & \makecell{29.7\\(28.3--31.3)} & 0.000006 \\
		
		TAB\_ & \makecell{89.9\\(79.0--96.7)} & \makecell{1.1\\(0.2--3.9)} & \makecell{0.0\\(0.0--0.0)} & 0.036
		& \makecell{50.7\\(48.8--52.6)} & \makecell{17.9\\(16.5--19.4)} & \makecell{29.5\\(28.1--31.0)} & 0.000014 \\

\end{longtable}
\end{ThreePartTable}

\begin{ThreePartTable}
	\begin{TableNotes}
		\footnotesize
		\item  LEF detection performance was derived using the single-predictor approach proposed in this study. AUROC, AUPRC, and F1 are reported with 95\% confidence intervals. The Thresh column indicates the thresholds used to maximize the F1 score on the validation set for predictor performance and LEF detection, respectively. Predictor abbreviations are defined in \cite{wagner2020ptb}.
	\end{TableNotes}
\begin{longtable}{l|c c c c}
	\caption{
Performance of LEF detection using the single-predictor approach on the external test cohort (ECG-Note) for all 71 ECG predictors.
} 
	\label{tab:results_single_predictor-2-ecgnote} \\
	\toprule
	\multirow{2.5}{*}{ \textbf{Predictor}} & \multicolumn{4}{c}{\textbf{LEF Detection (One Predictor)}} \\
	
	\cmidrule(lr){2-5}
	
	& \textbf{AUROC} & \textbf{AUPRC} & \textbf{F1 Score} & \textbf{Thresh} \\
	\midrule
	\endfirsthead
	\toprule
	\multicolumn{5}{c}{\small  \textbf{Table \thetable\ (continued)}} \\
	\cmidrule(lr){2-5}
	\multirow{2.5}{*}{\textbf{Predictor}} & \multicolumn{4}{c}{\textbf{LEF Detection (One Predictor)}} \\
	\cmidrule(lr){2-5}
	&  \textbf{AUROC} &  \textbf{AUPRC} &  \textbf{F1 Score} &  \textbf{Thresh} \\
	\midrule
	\endhead
	\multicolumn{5}{r}{\small Continued on next page} \\
	\endfoot
	\bottomrule
	\insertTableNotes
	\endlastfoot
	NORM    & \makecell{78.8 \\ (77.8-79.6)} & \makecell{36.2 \\ (34.5-38.1)} & \makecell{42.3 \\ (41.0-43.5)} & 0.003641 \\
	ILBBB   & \makecell{71.6 \\ (70.5-72.6)} & \makecell{33.2 \\ (31.5-35.0)} & \makecell{38.2 \\ (36.8-39.6)} & 0.000349 \\
	INJAL   & \makecell{73.4 \\ (72.4-74.3)} & \makecell{27.9 \\ (26.6-29.2)} & \makecell{40.9 \\ (39.6-42.1)} & 0.000386 \\
	ISCLA   & \makecell{67.7 \\ (66.6-68.8)} & \makecell{26.7 \\ (25.4-28.4)} & \makecell{34.0 \\ (32.6-35.2)} & 0.001918 \\
	ANEUR   & \makecell{71.0 \\ (70.0-72.0)} & \makecell{32.1 \\ (30.4-33.8)} & \makecell{38.0 \\ (36.7-39.4)} & 0.001924 \\
	ISCAL   & \makecell{65.7 \\ (64.7-66.8)} & \makecell{21.8 \\ (20.8-22.9)} & \makecell{33.2 \\ (32.0-34.3)} & 0.001148 \\
	ASMI    & \makecell{70.7 \\ (69.6-71.7)} & \makecell{29.3 \\ (27.8-31.0)} & \makecell{37.1 \\ (35.8-38.3)} & 0.023987 \\
	SVARR   & \makecell{64.0 \\ (62.8-65.1)} & \makecell{22.8 \\ (21.7-24.0)} & \makecell{32.1 \\ (30.8-33.4)} & 0.000218 \\
	INJIL   & \makecell{66.7 \\ (65.6-67.7)} & \makecell{22.9 \\ (21.8-24.2)} & \makecell{34.7 \\ (33.5-35.9)} & 0.000035 \\
	CRBBB   & \makecell{66.9 \\ (65.8-67.9)} & \makecell{23.4 \\ (22.2-24.7)} & \makecell{33.2 \\ (32.1-34.3)} & 0.000005 \\
	LAFB    & \makecell{67.5 \\ (66.5-68.5)} & \makecell{22.9 \\ (21.9-24.0)} & \makecell{34.9 \\ (33.7-36.1)} & 0.001226 \\
	ALMI    & \makecell{66.7 \\ (65.5-67.8)} & \makecell{29.7 \\ (28.0-31.4)} & \makecell{34.2 \\ (32.8-35.6)} & 0.002106 \\
	ABQRS   & \makecell{64.4 \\ (63.3-65.5)} & \makecell{24.2 \\ (22.9-25.6)} & \makecell{32.8 \\ (31.6-33.9)} & 0.012123 \\
	CLBBB   & \makecell{69.3 \\ (68.1-70.4)} & \makecell{29.7 \\ (28.1-31.3)} & \makecell{36.0 \\ (34.8-37.2)} & 0.000006 \\
	ILMI    & \makecell{65.8 \\ (64.7-66.9)} & \makecell{25.9 \\ (24.5-27.3)} & \makecell{33.0 \\ (31.8-34.2)} & 0.000172 \\
	INJAS   & \makecell{65.1 \\ (64.0-66.2)} & \makecell{21.9 \\ (20.9-23.1)} & \makecell{33.9 \\ (32.7-35.1)} & 0.000571 \\
	INVT    & \makecell{62.5 \\ (61.1-63.7)} & \makecell{22.9 \\ (21.7-24.2)} & \makecell{31.5 \\ (30.3-32.6)} & 0.005779 \\
	PVC     & \makecell{59.5 \\ (58.1-60.7)} & \makecell{22.6 \\ (21.2-24.0)} & \makecell{29.5 \\ (28.2-30.6)} & 0.001254 \\
	ISCIL   & \makecell{59.2 \\ (58.1-60.3)} & \makecell{18.7 \\ (17.9-19.8)} & \makecell{29.9 \\ (28.9-30.9)} & 0.000142 \\
	1AVB    & \makecell{61.5 \\ (60.3-62.6)} & \makecell{21.5 \\ (20.5-22.7)} & \makecell{30.5 \\ (29.5-31.6)} & 0.000059 \\
	ISC\_   & \makecell{59.6 \\ (58.2-60.8)} & \makecell{22.0 \\ (20.8-23.3)} & \makecell{29.5 \\ (28.3-30.6)} & 0.007637 \\
	IVCD    & \makecell{64.3 \\ (63.0-65.5)} & \makecell{28.7 \\ (27.0-30.5)} & \makecell{33.2 \\ (32.0-34.6)} & 0.027054 \\
	LAO/LAE & \makecell{62.5 \\ (61.0-63.7)} & \makecell{28.6 \\ (26.7-30.3)} & \makecell{32.9 \\ (31.4-34.3)} & 0.051849 \\
	ISCAN   & \makecell{56.8 \\ (55.5-57.9)} & \makecell{18.3 \\ (17.4-19.2)} & \makecell{27.4 \\ (26.2-28.5)} & 0.000312 \\
	ISCAS   & \makecell{54.8 \\ (53.5-55.9)} & \makecell{16.5 \\ (15.7-17.2)} & \makecell{28.3 \\ (27.3-29.3)} & 0.000066 \\
	AFIB    & \makecell{57.8 \\ (56.6-58.9)} & \makecell{18.3 \\ (17.5-19.2)} & \makecell{29.0 \\ (28.0-30.0)} & 0.000004 \\
	BIGU    & \makecell{58.9 \\ (57.7-60.0)} & \makecell{19.1 \\ (18.2-20.2)} & \makecell{29.5 \\ (28.5-30.6)} & 0.000013 \\
	SVTAC   & \makecell{52.7 \\ (51.5-54.0)} & \makecell{16.7 \\ (15.9-17.6)} & \makecell{24.2 \\ (23.0-25.4)} & 0.000113 \\
	AMI     & \makecell{54.0 \\ (52.9-55.2)} & \makecell{16.7 \\ (15.9-17.5)} & \makecell{28.0 \\ (27.0-29.0)} & 0.000868 \\
	NST\_   & \makecell{46.4 \\ (45.2-47.4)} & \makecell{13.7 \\ (13.1-14.3)} & \makecell{26.1 \\ (25.2-27.1)} & 0.005043 \\
	3AVB    & \makecell{57.7 \\ (56.6-58.9)} & \makecell{19.0 \\ (18.1-20.0)} & \makecell{28.3 \\ (27.4-29.3)} & <0.000001 \\
	IMI     & \makecell{58.9 \\ (57.8-60.0)} & \makecell{19.7 \\ (18.8-20.8)} & \makecell{29.8 \\ (28.7-30.7)} & 0.003210 \\
	LPR     & \makecell{56.4 \\ (55.1-57.5)} & \makecell{17.6 \\ (16.8-18.5)} & \makecell{28.7 \\ (27.7-29.7)} & 0.000049 \\
	2AVB    & \makecell{56.6 \\ (55.4-57.9)} & \makecell{17.6 \\ (16.7-18.6)} & \makecell{28.8 \\ (27.7-29.9)} & 0.000002 \\
	DIG     & \makecell{53.9 \\ (52.7-55.1)} & \makecell{16.2 \\ (15.5-17.0)} & \makecell{28.2 \\ (27.2-29.2)} & 0.000024 \\
	LMI     & \makecell{59.0 \\ (57.8-60.2)} & \makecell{20.2 \\ (19.2-21.4)} & \makecell{29.4 \\ (28.3-30.4)} & 0.000937 \\
	LOWT    & \makecell{49.7 \\ (48.5-50.9)} & \makecell{15.2 \\ (14.5-16.0)} & \makecell{26.3 \\ (25.3-27.2)} & 0.000556 \\
	SR      & \makecell{57.1 \\ (55.9-58.3)} & \makecell{19.1 \\ (18.1-20.2)} & \makecell{29.1 \\ (28.0-30.2)} & 0.943848 \\
	STACH   & \makecell{54.5 \\ (53.3-55.6)} & \makecell{16.4 \\ (15.6-17.2)} & \makecell{28.6 \\ (27.6-29.5)} & 0.000001 \\
	LPFB    & \makecell{58.9 \\ (57.7-60.2)} & \makecell{19.4 \\ (18.5-20.5)} & \makecell{29.4 \\ (28.4-30.5)} & 0.000075 \\
	PACE    & \makecell{57.9 \\ (56.7-59.1)} & \makecell{22.3 \\ (21.0-23.8)} & \makecell{28.3 \\ (27.3-29.2)} & 0.000033 \\
	WPW     & \makecell{55.8 \\ (54.5-56.9)} & \makecell{18.6 \\ (17.6-19.7)} & \makecell{26.5 \\ (25.3-27.7)} & 0.000028 \\
	ISCIN   & \makecell{51.3 \\ (50.1-52.5)} & \makecell{15.6 \\ (14.9-16.4)} & \makecell{25.5 \\ (24.4-26.5)} & 0.000432 \\
	PRC(S)  & \makecell{53.2 \\ (51.9-54.4)} & \makecell{16.6 \\ (15.8-17.5)} & \makecell{26.0 \\ (25.0-27.0)} & 0.000008 \\
	AFLT    & \makecell{48.4 \\ (47.1-49.6)} & \makecell{14.9 \\ (14.1-15.6)} & \makecell{25.1 \\ (24.0-26.0)} & 0.000010 \\
	INJIN   & \makecell{56.6 \\ (55.3-57.8)} & \makecell{18.4 \\ (17.5-19.5)} & \makecell{27.8 \\ (26.7-29.0)} & 0.000062 \\
	PAC     & \makecell{53.9 \\ (52.6-55.0)} & \makecell{17.0 \\ (16.2-18.0)} & \makecell{27.0 \\ (26.0-28.0)} & 0.000290 \\
	IPMI    & \makecell{51.1 \\ (49.8-52.4)} & \makecell{17.1 \\ (16.2-18.3)} & \makecell{26.3 \\ (25.4-27.3)} & 0.000016 \\
	STD\_   & \makecell{53.6 \\ (52.4-54.9)} & \makecell{17.5 \\ (16.7-18.5)} & \makecell{26.6 \\ (25.7-27.5)} & 0.004936 \\
	LNGQT   & \makecell{44.8 \\ (43.5-46.1)} & \makecell{13.9 \\ (13.2-14.6)} & \makecell{26.0 \\ (25.1-26.8)} & 0.000042 \\
	TRIGU   & \makecell{51.2 \\ (49.9-52.6)} & \makecell{16.2 \\ (15.5-17.1)} & \makecell{25.8 \\ (24.8-26.7)} & 0.000002 \\
	NDT     & \makecell{32.4 \\ (31.2-33.6)} & \makecell{11.0 \\ (10.6-11.5)} & \makecell{27.1 \\ (26.2-27.9)} & <0.000001 \\
	LVH     & \makecell{50.8 \\ (49.4-52.2)} & \makecell{18.2 \\ (17.1-19.4)} & \makecell{27.1 \\ (26.3-27.9)} & 0.000005 \\
	PSVT    & \makecell{36.6 \\ (35.4-37.9)} & \makecell{12.0 \\ (11.5-12.6)} & \makecell{27.2 \\ (26.3-28.0)} & <0.000001 \\
	INJLA   & \makecell{52.1 \\ (50.7-53.4)} & \makecell{17.0 \\ (16.1-17.9)} & \makecell{25.2 \\ (24.0-26.2)} & 0.000020 \\
	PMI     & \makecell{37.2 \\ (36.0-38.3)} & \makecell{12.0 \\ (11.4-12.5)} & \makecell{27.2 \\ (26.3-28.0)} & <0.000001 \\
	STE\_   & \makecell{44.9 \\ (43.7-46.2)} & \makecell{14.0 \\ (13.4-14.7)} & \makecell{27.2 \\ (26.3-28.0)} & <0.000001 \\
	SEHYP   & \makecell{51.4 \\ (50.3-52.6)} & \makecell{16.3 \\ (15.5-17.1)} & \makecell{27.1 \\ (26.3-27.9)} & <0.000001 \\
	SBRAD   & \makecell{49.0 \\ (47.8-50.1)} & \makecell{14.7 \\ (14.0-15.3)} & \makecell{27.2 \\ (26.4-28.1)} & <0.000001 \\
	RAO/RAE & \makecell{48.1 \\ (46.9-49.4)} & \makecell{16.3 \\ (15.4-17.3)} & \makecell{27.2 \\ (26.3-28.0)} & <0.000001 \\
	VCLVH   & \makecell{38.8 \\ (37.5-40.1)} & \makecell{12.4 \\ (11.9-13.0)} & \makecell{27.0 \\ (26.1-27.8)} & 0.000012 \\
	IRBBB   & \makecell{41.8 \\ (40.6-43.1)} & \makecell{12.8 \\ (12.2-13.4)} & \makecell{27.1 \\ (26.3-27.9)} & <0.000001 \\
	QWAVE   & \makecell{52.8 \\ (51.5-54.0)} & \makecell{18.7 \\ (17.7-19.8)} & \makecell{27.1 \\ (26.3-28.0)} & 0.000018 \\
	NT\_    & \makecell{35.4 \\ (34.1-36.5)} & \makecell{11.6 \\ (11.2-12.2)} & \makecell{27.2 \\ (26.3-28.0)} & <0.000001 \\
	EL      & \makecell{39.3 \\ (38.0-40.6)} & \makecell{12.5 \\ (12.0-13.2)} & \makecell{27.1 \\ (26.2-27.9)} & 0.000001 \\
	HVOLT   & \makecell{34.0 \\ (32.8-35.1)} & \makecell{11.2 \\ (10.7-11.7)} & \makecell{27.2 \\ (26.3-28.0)} & <0.000001 \\
	RVH     & \makecell{46.7 \\ (45.5-48.0)} & \makecell{14.1 \\ (13.5-14.8)} & \makecell{27.2 \\ (26.3-28.0)} & <0.000001 \\
	LVOLT   & \makecell{46.6 \\ (45.4-47.8)} & \makecell{14.5 \\ (13.8-15.2)} & \makecell{27.2 \\ (26.3-28.0)} & <0.000001 \\
	SARRH   & \makecell{50.8 \\ (49.5-52.1)} & \makecell{16.3 \\ (15.5-17.2)} & \makecell{27.2 \\ (26.3-28.0)} & 0.000012 \\
	IPLMI   & \makecell{51.1 \\ (49.8-52.3)} & \makecell{16.6 \\ (15.7-17.7)} & \makecell{27.2 \\ (26.3-28.1)} & 0.000006 \\
	TAB\_   & \makecell{49.1 \\ (47.9-50.3)} & \makecell{15.4 \\ (14.6-16.2)} & \makecell{26.6 \\ (25.7-27.5)} & 0.000014 \\
\end{longtable}
\end{ThreePartTable}

\subsection{Threshold selection for the single-predictor approach}
Following \citep{ribeiro2020automatic}, we adopt the F1 score as a primary performance metric due to its robustness to class imbalance. 
For the single-predictor method, the classification threshold is selected to maximize the F1 score on the validation set, consistent with prior studies \citep{ribeiro2020automatic, diao2025speed}. 
As an alternative operating point, we also consider a recall-based threshold that enforces a minimum recall of 90\%.

\begin{figure}[htbp]
	\centering
	\includegraphics[width=0.95\textwidth]{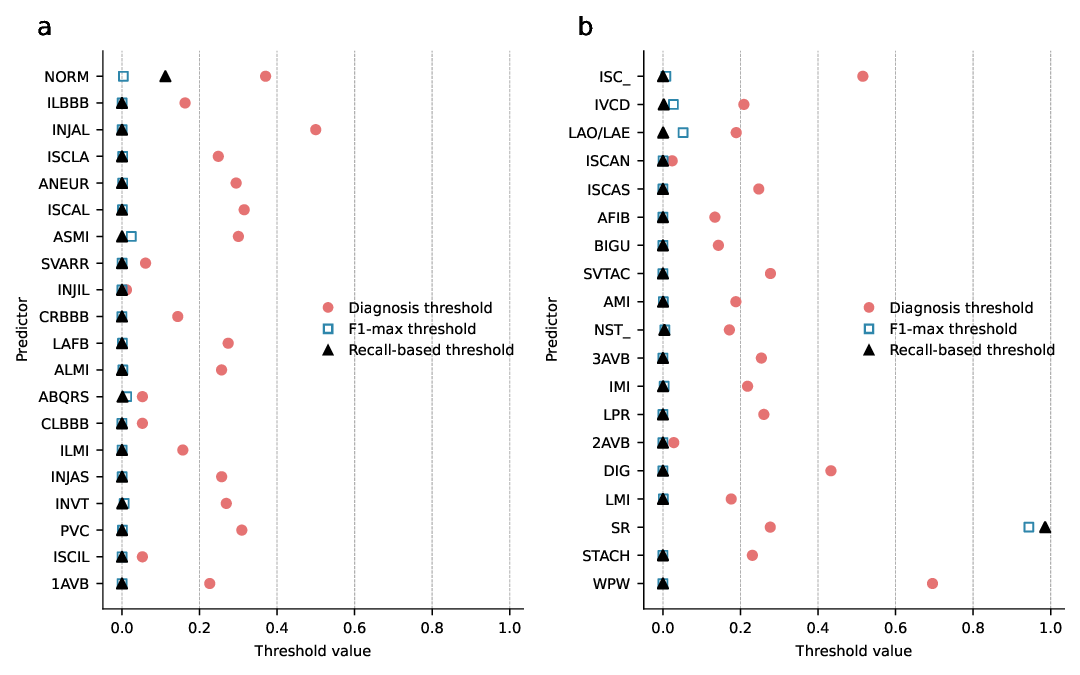}
\caption{
	\textbf{Operating thresholds of individual predictors for LEF detection.} 
	\textbf{a}, Left column: thresholds for the first half of predictors (AUROC $\ge$ 60\%). 
	\textbf{b}, Right column: thresholds for the remaining predictors. 
	For each predictor, three types of thresholds are shown: 
	1) \textit{Diagnosis threshold}: threshold for determining whether the ECG shows a positive signal for that predictor; 
	2) \textit{F1-max threshold}: threshold for predicting LEF positivity using that predictor, chosen to maximize F1 score on the validation set; 
	3) \textit{Recall-based threshold}: threshold for predicting LEF positivity using that predictor, set to achieve recall $\ge$ 90\%. 
	Only predictors with AUROC $\ge$ 60\% are shown for clarity; complete results for all predictors are provided in Tables \ref{tab:results_single_predictor} and \ref{tab:results_single_predictor-2}.
}
	\label{fig:LEF_thresholds}
\end{figure}

Notably, under both thresholding strategies, the thresholds used by the single-predictor method for detecting LEF are substantially lower than the corresponding diagnosis thresholds employed in traditional automatic ECG diagnosis models. This trend is observed for most predictors with AUROC greater than 0.60, as summarized in Figure \ref{fig:LEF_thresholds}.

\subsection{Subgroups analysis for single-predictor approach}
\label{apedix:subgroup_single_predictor}
The subgroup analysis results for NORM, ILBBB, INJIL, and ISCLA in the internal test set are presented in Tables \ref{table:subgroup_NORM_ILBBB} and \ref{table:subgroup_INJIL_ISCLA}. Corresponding results in the external test set are shown in Tables \ref{table:subgroup_NORM_ILBBB_lvefllm} and \ref{table:subgroup_INJIL_ISCLA_lvefllm}.

\begin{table*}[!htbp]
	\centering
	\begin{threeparttable}
	\footnotesize
	\setlength{\tabcolsep}{1pt}
\caption{
	Subgroup performance of the single-predictor approach for the NORM and ILBBB predictors on the internal test set. \label{table:subgroup_NORM_ILBBB}
}
	\begin{tabular}{lcc|ccc|c ccc}
		\toprule
		& & & \multicolumn{3}{c}{\textbf{NORM}} & \multicolumn{3}{c}{\textbf{ILBBB}} \\
		\cmidrule(lr){4-6} \cmidrule(lr){7-9}
		\textbf{Subgroup} & \textbf{n} & \textbf{Prevalence (\%)} &
		\textbf{AUROC} & \textbf{AUPRC} & \textbf{F1 Score} &
		\textbf{AUROC} & \textbf{AUPRC} & \textbf{F1 Score} \\
		\midrule
		
		\multicolumn{9}{l}{\textbf{Age groups}} \\
		\midrule 
		\quad 18--59 & 2124 & 13.6 &
		\makecell{80.9 \\ (78.2--83.8)} &
		\makecell{41.3 \\ (36.0--48.3)} &
		\makecell{47.1 \\ (42.6--51.6)} &
		\makecell{79.3 \\ (76.2--82.1)} &
		\makecell{44.1 \\ (38.7--49.9)} &
		\makecell{44.4 \\ (39.6--49.3)} \\
		
		\quad 60--69 & 1318 & 16.6 &
		\makecell{81.6 \\ (78.3--84.8)} &
		\makecell{49.3 \\ (42.2--56.6)} &
		\makecell{50.4 \\ (45.5--55.0)} &
		\makecell{78.1 \\ (74.7--81.1)} &
		\makecell{44.8 \\ (37.9--51.4)} &
		\makecell{45.8 \\ (40.1--51.0)} \\
		
		\quad 70--79 & 1154 & 21.7 &
		\makecell{80.6 \\ (77.6--83.5)} &
		\makecell{53.6 \\ (47.7--60.5)} &
		\makecell{55.4 \\ (51.0--59.9)} &
		\makecell{81.1 \\ (78.0--84.0)} &
		\makecell{55.7 \\ (49.5--61.8)} &
		\makecell{58.5 \\ (53.4--62.9)} \\
		
		\quad 80+ & 846 & 24.1 &
		\makecell{77.0 \\ (73.4--80.2)} &
		\makecell{47.9 \\ (41.9--54.8)} &
		\makecell{52.8 \\ (48.2--57.1)} &
		\makecell{77.8 \\ (74.5--81.3)} &
		\makecell{50.7 \\ (44.8--58.2)} &
		\makecell{53.8 \\ (48.6--58.9)} \\
		
		\midrule
		\multicolumn{9}{l}{\textbf{Sex}} \\
		\midrule 
		\quad Female & 2731 & 12.4 &
		\makecell{81.2 \\ (78.6--83.6)} &
		\makecell{39.0 \\ (34.2--44.9)} &
		\makecell{43.9 \\ (40.4--47.5)} &
		\makecell{81.9 \\ (79.5--84.1)} &
		\makecell{43.6 \\ (38.0--49.4)} &
		\makecell{45.7 \\ (41.3--49.9)} \\
		
		\quad Male & 2711 & 23.0 &
		\makecell{80.2 \\ (78.4--82.0)} &
		\makecell{53.5 \\ (49.4--57.8)} &
		\makecell{56.4 \\ (53.6--59.4)} &
		\makecell{77.5 \\ (75.4--79.6)} &
		\makecell{52.3 \\ (48.2--56.6)} &
		\makecell{53.5 \\ (50.5--56.9)} \\
		
		\midrule
		\multicolumn{9}{l}{\textbf{Race/ethnicity}} \\
		\midrule 
		\quad Hispanic & 1649 & 16.7 &
		\makecell{83.3 \\ (80.6--85.9)} &
		\makecell{52.6 \\ (46.5--58.8)} &
		\makecell{53.9 \\ (50.0--58.2)} &
		\makecell{81.1 \\ (78.3--83.7)} &
		\makecell{50.3 \\ (44.5--56.4)} &
		\makecell{50.0 \\ (45.0--54.6)} \\
		
		\quad White & 1569 & 16.3 &
		\makecell{81.7 \\ (79.1--84.4)} &
		\makecell{45.0 \\ (39.1--51.7)} &
		\makecell{48.3 \\ (43.9--52.4)} &
		\makecell{79.4 \\ (76.4--82.2)} &
		\makecell{45.3 \\ (39.1--52.0)} &
		\makecell{48.9 \\ (43.5--53.6)} \\
		
		\quad Black & 846 & 19.3 &
		\makecell{78.1 \\ (74.4--81.8)} &
		\makecell{47.2 \\ (39.9--56.2)} &
		\makecell{50.2 \\ (44.5--55.9)} &
		\makecell{79.8 \\ (75.8--83.3)} &
		\makecell{50.0 \\ (42.4--58.8)} &
		\makecell{50.7 \\ (44.2--56.6)} \\
		
		\quad Asian & 153 & 16.3 &
		\makecell{82.7 \\ (74.6--90.2)} &
		\makecell{50.4 \\ (32.3--69.7)} &
		\makecell{45.9 \\ (31.0--58.7)} &
		\makecell{88.1 \\ (81.0--94.2)} &
		\makecell{65.5 \\ (48.4--81.2)} &
		\makecell{57.6 \\ (40.8--71.0)} \\
		
		\quad Other & 457 & 15.8 &
		\makecell{81.4 \\ (75.8--86.6)} &
		\makecell{42.1 \\ (32.3--54.0)} &
		\makecell{50.9 \\ (42.9--58.9)} &
		\makecell{77.2 \\ (70.9--83.4)} &
		\makecell{50.4 \\ (39.3--62.1)} &
		\makecell{50.0 \\ (41.2--58.6)} \\
		
		\quad Unknown & 768 & 22.3 &
		\makecell{77.7 \\ (73.6--81.7)} &
		\makecell{47.7 \\ (40.4--56.3)} &
		\makecell{54.7 \\ (49.0--60.0)} &
		\makecell{78.8 \\ (74.8--82.5)} &
		\makecell{49.5 \\ (42.0--57.6)} &
		\makecell{53.2 \\ (46.5--59.1)} \\
		
		\midrule
		\multicolumn{9}{l}{\textbf{Clinical context}} \\
		\midrule 
		\quad Emergency & 1971 & 15.7 &
		\makecell{79.9 \\ (77.2--82.4)} &
		\makecell{40.8 \\ (35.8--46.3)} &
		\makecell{47.9 \\ (43.9--51.7)} &
		\makecell{82.0 \\ (79.5--84.2)} &
		\makecell{48.3 \\ (42.7--53.9)} &
		\makecell{49.9 \\ (45.2--54.0)} \\
		
		\quad Inpatient & 2203 & 24.4 &
		\makecell{79.0 \\ (76.7--81.1)} &
		\makecell{54.9 \\ (50.2--59.7)} &
		\makecell{56.4 \\ (53.3--59.5)} &
		\makecell{75.8 \\ (73.5--78.1)} &
		\makecell{52.5 \\ (48.2--57.0)} &
		\makecell{52.4 \\ (49.1--55.9)} \\
		
		\quad Outpatient & 1059 & 6.6 &
		\makecell{83.8 \\ (78.1--88.8)} &
		\makecell{34.6 \\ (25.1--45.4)} &
		\makecell{35.7 \\ (28.8--42.7)} &
		\makecell{82.0 \\ (76.3--87.2)} &
		\makecell{31.5 \\ (21.8--43.7)} &
		\makecell{40.9 \\ (32.0--49.3)} \\
		
		\quad Procedural & 209 & 21.5 &
		\makecell{81.1 \\ (74.6--87.0)} &
		\makecell{46.3 \\ (34.1--61.7)} &
		\makecell{55.2 \\ (42.7--65.2)} &
		\makecell{75.3 \\ (65.9--83.1)} &
		\makecell{44.9 \\ (31.8--60.9)} &
		\makecell{50.0 \\ (35.9--61.0)} \\
		
		\bottomrule
	\end{tabular}
\begin{tablenotes}
	\item  AUROC, AUPRC, and F1 are reported with 95\% confidence intervals. NORM, normal ECG; ILBBB, incomplete left bundle branch block.
\end{tablenotes}
	\end{threeparttable}
\end{table*}

\begin{table*}[!htbp]
	\centering
\begin{threeparttable}
	\footnotesize
	\setlength{\tabcolsep}{1pt}{
	\caption{Subgroup performance of the single-predictor method for INJIL and ISCLA in the internal test set.  \label{table:subgroup_INJIL_ISCLA}}
	\begin{tabular}{lcc|ccc|c  ccc}
		\toprule
		& & & \multicolumn{3}{c}{\textbf{INJIL}} & \multicolumn{3}{c}{\textbf{ISCLA}} \\
		\cmidrule(lr){4-6} \cmidrule(lr){7-9}
		\textbf{Subgroup} & \textbf{n} & \textbf{Prevalence (\%)} &
		\textbf{AUROC} & \textbf{AUPRC} & \textbf{F1 Score} &
		\textbf{AUROC} & \textbf{AUPRC} & \textbf{F1 Score} \\
		\midrule
		
		\multicolumn{9}{l}{\textbf{Age groups}} \\
		\midrule 
		\quad 18--59 & 2124 & 13.6 &
		\makecell{77.4 \\ (74.6--80.2)} &
		\makecell{32.4 \\ (28.4--37.9)} &
		\makecell{40.8 \\ (36.6--45.1)} &
		\makecell{79.0 \\ (76.4--81.8)} &
		\makecell{36.6 \\ (31.8--43.0)} &
		\makecell{43.3 \\ (39.3--47.4)} \\
		
		\quad 60--69 & 1318 & 16.6 &
		\makecell{75.4 \\ (72.0--78.8)} &
		\makecell{34.0 \\ (29.0--40.1)} &
		\makecell{45.5 \\ (40.6--50.2)} &
		\makecell{74.4 \\ (70.9--78.0)} &
		\makecell{37.0 \\ (30.8--44.3)} &
		\makecell{43.0 \\ (38.1--48.0)} \\
		
		\quad 70--79 & 1154 & 21.7 &
		\makecell{73.9 \\ (70.6--77.1)} &
		\makecell{39.5 \\ (34.5--45.6)} &
		\makecell{48.0 \\ (43.1--52.7)} &
		\makecell{73.9 \\ (70.6--77.2)} &
		\makecell{42.5 \\ (36.9--49.3)} &
		\makecell{45.8 \\ (40.9--50.1)} \\
		
		\quad 80+ & 846 & 24.1 &
		\makecell{69.2 \\ (65.1--72.9)} &
		\makecell{36.3 \\ (31.2--42.8)} &
		\makecell{47.5 \\ (42.3--52.5)} &
		\makecell{70.8 \\ (67.2--74.5)} &
		\makecell{43.3 \\ (37.2--50.1)} &
		\makecell{43.8 \\ (38.5--48.8)} \\
		
		\midrule
		\multicolumn{9}{l}{\textbf{Sex}} \\
		\midrule 
		\quad Female & 2731 & 12.4 &
		\makecell{75.2 \\ (72.7--77.7)} &
		\makecell{26.1 \\ (22.8--30.3)} &
		\makecell{36.2 \\ (32.8--40.0)} &
		\makecell{76.8 \\ (74.4--79.3)} &
		\makecell{29.5 \\ (25.6--34.7)} &
		\makecell{37.7 \\ (34.1--41.6)} \\
		
		\quad Male & 2711 & 23.0 &
		\makecell{74.3 \\ (72.2--76.3)} &
		\makecell{41.2 \\ (37.8--45.1)} &
		\makecell{51.1 \\ (48.2--54.0)} &
		\makecell{75.3 \\ (73.2--77.3)} &
		\makecell{46.2 \\ (42.6--50.4)} &
		\makecell{48.5 \\ (45.6--51.5)} \\
		
		\midrule
		\multicolumn{9}{l}{\textbf{Race/ethnicity}} \\
		\midrule 
		\quad Hispanic & 1649 & 16.7 &
		\makecell{76.1 \\ (73.0--79.2)} &
		\makecell{34.6 \\ (30.1--40.3)} &
		\makecell{45.1 \\ (40.8--49.6)} &
		\makecell{75.5 \\ (72.6--78.4)} &
		\makecell{39.0 \\ (33.1--45.3)} &
		\makecell{42.0 \\ (37.5--46.3)} \\
		
		\quad White & 1569 & 16.3 &
		\makecell{76.2 \\ (73.2--79.3)} &
		\makecell{35.0 \\ (30.2--41.0)} &
		\makecell{43.8 \\ (39.6--48.3)} &
		\makecell{76.8 \\ (74.0--79.7)} &
		\makecell{35.9 \\ (31.1--42.1)} &
		\makecell{42.1 \\ (37.5--46.7)} \\
		
		\quad Black & 846 & 19.3 &
		\makecell{72.0 \\ (68.0--76.1)} &
		\makecell{32.6 \\ (27.5--39.6)} &
		\makecell{44.9 \\ (39.1--50.8)} &
		\makecell{76.1 \\ (72.3--80.0)} &
		\makecell{41.5 \\ (34.9--49.8)} &
		\makecell{46.6 \\ (41.2--51.8)} \\
		
		\quad Asian & 153 & 16.3 &
		\makecell{78.1 \\ (67.1--87.2)} &
		\makecell{47.8 \\ (30.3--64.3)} &
		\makecell{45.0 \\ (29.6--56.8)} &
		\makecell{78.0 \\ (67.1--87.5)} &
		\makecell{41.7 \\ (26.7--63.5)} &
		\makecell{45.9 \\ (29.8--59.0)} \\
		
		\quad Other & 457 & 15.8 &
		\makecell{76.1 \\ (70.7--81.7)} &
		\makecell{35.6 \\ (27.4--46.6)} &
		\makecell{41.6 \\ (32.9--50.0)} &
		\makecell{75.6 \\ (70.0--80.8)} &
		\makecell{37.1 \\ (27.4--48.4)} &
		\makecell{42.3 \\ (33.7--50.0)} \\
		
		\quad Unknown & 768 & 22.3 &
		\makecell{74.0 \\ (69.8--77.8)} &
		\makecell{39.0 \\ (33.2--46.7)} &
		\makecell{48.7 \\ (43.2--54.2)} &
		\makecell{74.3 \\ (70.3--78.5)} &
		\makecell{45.0 \\ (37.9--53.2)} &
		\makecell{47.9 \\ (42.2--53.5)} \\
		
		\midrule
		\multicolumn{9}{l}{\textbf{Clinical context}} \\
		\midrule 
		\quad Emergency & 1971 & 15.7 &
		\makecell{76.6 \\ (73.9--79.2)} &
		\makecell{31.9 \\ (28.1--36.6)} &
		\makecell{43.7 \\ (39.6--47.5)} &
		\makecell{77.6 \\ (75.0--79.8)} &
		\makecell{38.6 \\ (33.6--44.3)} &
		\makecell{43.2 \\ (39.3--47.1)} \\
		
		\quad Inpatient & 2203 & 24.4 &
		\makecell{70.8 \\ (68.3--73.2)} &
		\makecell{39.3 \\ (35.7--43.4)} &
		\makecell{48.0 \\ (44.7--51.4)} &
		\makecell{70.4 \\ (67.9--72.8)} &
		\makecell{41.7 \\ (37.5--46.2)} &
		\makecell{46.9 \\ (43.4--50.3)} \\
		
		\quad Outpatient & 1059 & 6.6 &
		\makecell{75.4 \\ (69.5--81.5)} &
		\makecell{20.2 \\ (14.4--29.2)} &
		\makecell{29.2 \\ (21.7--36.2)} &
		\makecell{81.4 \\ (75.7--86.3)} &
		\makecell{32.4 \\ (23.2--45.3)} &
		\makecell{31.1 \\ (23.9--38.3)} \\
		
		\quad Procedural & 209 & 21.5 &
		\makecell{77.2 \\ (69.8--84.8)} &
		\makecell{44.0 \\ (32.0--60.8)} &
		\makecell{55.4 \\ (43.7--66.7)} &
		\makecell{76.6 \\ (69.6--83.4)} &
		\makecell{43.8 \\ (30.8--60.5)} &
		\makecell{42.3 \\ (29.1--53.9)} \\
		
		\bottomrule
	\end{tabular}
}
\begin{tablenotes}
	\item AUROC, AUPRC, and F1 are reported with 95\% confidence intervals. INJIL,  subendocardial injury in inferolateral leads; ISCLA, ischemic in lateral leads.
\end{tablenotes}
\end{threeparttable}
\end{table*}

\begin{table*}[!htbp]
	\centering
	\begin{threeparttable}
	\footnotesize
	\setlength{\tabcolsep}{1pt}{
		\caption{Subgroup performance of the single-predictor method for NORM and ILBBB in the external test set. \label{table:subgroup_NORM_ILBBB_lvefllm}}
		\begin{tabular}{lcc|ccc|ccc}
			\toprule
			& & & \multicolumn{3}{c}{\textbf{NORM}} & \multicolumn{3}{c}{\textbf{ILBBB}} \\
			\cmidrule(lr){4-6} \cmidrule(lr){7-9}
			\textbf{Subgroup} & \textbf{n} & \textbf{Prevalence (\%)} &
			\textbf{AUROC} & \textbf{AUPRC} & \textbf{F1 Score} &
			\textbf{AUROC} & \textbf{AUPRC} & \textbf{F1 Score} \\
			\midrule
			\multicolumn{9}{l}{\textbf{Age groups}} \\
			\midrule
			\quad 18--59 & 4270 & 12.2 & \makecell{81.5\\(79.4--83.3)} & \makecell{37.1\\(33.1--41.5)} & \makecell{44.6\\(41.6--47.7)} & \makecell{73.7\\(71.3--76.0)} & \makecell{33.7\\(30.3--38.0)} & \makecell{40.4\\(37.7--43.2)} \\
			\quad 60--69 & 3637 & 15.7 & \makecell{78.5\\(76.5--80.3)} & \makecell{40.8\\(36.9--45.3)} & \makecell{46.2\\(43.4--48.9)} & \makecell{71.6\\(69.3--73.9)} & \makecell{37.6\\(33.9--41.8)} & \makecell{43.4\\(40.5--46.1)} \\
			\quad 70--79 & 3761 & 18.0 & \makecell{77.7\\(76.1--79.5)} & \makecell{39.3\\(36.1--43.3)} & \makecell{43.7\\(41.3--46.0)} & \makecell{70.5\\(68.4--72.7)} & \makecell{36.0\\(33.1--39.7)} & \makecell{39.9\\(37.4--42.5)} \\
			\quad 80+ & 4349 & 17.2 & \makecell{74.9\\(73.1--76.7)} & \makecell{34.2\\(31.3--37.6)} & \makecell{38.4\\(36.4--40.4)} & \makecell{67.0\\(64.9--69.2)} & \makecell{29.3\\(26.8--32.4)} & \makecell{36.1\\(33.9--38.5)} \\
			\midrule
			\multicolumn{9}{l}{\textbf{Sex}} \\
			\midrule
			\quad Female & 7222 & 11.2 & \makecell{74.2\\(72.4--75.9)} & \makecell{22.6\\(19.9--25.8)} & \makecell{32.4\\(30.4--34.5)} & \makecell{65.3\\(63.0--67.6)} & \makecell{19.5\\(17.1--22.2)} & \makecell{29.8\\(27.4--32.0)} \\
			\quad Male & 8795 & 19.4 & \makecell{77.1\\(75.9--78.4)} & \makecell{42.2\\(39.6--44.9)} & \makecell{45.4\\(43.6--47.2)} & \makecell{71.6\\(70.0--73.0)} & \makecell{39.6\\(36.9--42.3)} & \makecell{43.1\\(41.2--45.0)} \\
			\midrule
			\multicolumn{9}{l}{\textbf{Race/ethnicity}} \\
			\midrule
			\quad Hispanic & 638 & 14.7 & \makecell{81.7\\(77.2--85.8)} & \makecell{39.0\\(30.4--48.8)} & \makecell{45.8\\(39.3--52.6)} & \makecell{68.6\\(63.1--74.1)} & \makecell{30.3\\(23.0--39.1)} & \makecell{39.1\\(31.8--46.2)} \\
			\quad White & 11688 & 15.5 & \makecell{77.0\\(75.9--77.9)} & \makecell{35.2\\(33.2--37.5)} & \makecell{41.7\\(40.4--43.1)} & \makecell{69.4\\(68.2--70.6)} & \makecell{31.1\\(29.0--33.4)} & \makecell{39.4\\(38.0--40.9)} \\
			\quad Black & 1845 & 17.3 & \makecell{79.6\\(77.2--81.9)} & \makecell{44.3\\(39.1--50.2)} & \makecell{46.2\\(42.4--49.8)} & \makecell{72.6\\(69.8--75.4)} & \makecell{40.7\\(35.7--46.4)} & \makecell{43.4\\(39.6--47.2)} \\
			\quad Asian & 414 & 12.6 & \makecell{74.7\\(69.4--80.0)} & \makecell{26.1\\(18.4--36.6)} & \makecell{37.2\\(28.5--45.1)} & \makecell{69.0\\(63.0--75.2)} & \makecell{24.2\\(17.4--33.7)} & \makecell{35.0\\(27.1--43.4)} \\
			\quad Other & 524 & 13.0 & \makecell{73.7\\(68.1--79.1)} & \makecell{22.5\\(15.9--32.1)} & \makecell{34.4\\(27.5--41.4)} & \makecell{66.2\\(60.3--72.5)} & \makecell{21.2\\(15.3--29.9)} & \makecell{32.9\\(26.2--39.6)} \\
			\quad Unknown & 908 & 18.8 & \makecell{75.5\\(71.4--79.5)} & \makecell{39.5\\(33.1--46.7)} & \makecell{41.8\\(36.8--46.7)} & \makecell{66.2\\(61.8--70.6)} & \makecell{34.9\\(29.7--41.1)} & \makecell{38.0\\(33.1--42.7)} \\
			\midrule
			\multicolumn{9}{l}{\textbf{Clinical context}} \\
			\midrule
			\quad Emergency & 9170 & 14.6 & \makecell{76.3\\(75.1--77.5)} & \makecell{33.6\\(31.4--36.0)} & \makecell{40.6\\(39.1--42.2)} & \makecell{69.2\\(67.7--70.6)} & \makecell{31.1\\(28.9--33.7)} & \makecell{39.2\\(37.7--40.7)} \\
			\quad Urgent & 3028 & 20.9 & \makecell{78.7\\(76.9--80.5)} & \makecell{45.4\\(41.5--49.2)} & \makecell{49.0\\(46.5--51.6)} & \makecell{71.2\\(69.2--73.2)} & \makecell{40.8\\(36.9--45.0)} & \makecell{44.9\\(42.2--47.5)} \\
			\quad Observation & 2173 & 19.6 & \makecell{79.6\\(77.7--81.7)} & \makecell{49.0\\(44.0--54.2)} & \makecell{51.3\\(48.4--54.1)} & \makecell{73.9\\(71.7--76.0)} & \makecell{44.9\\(40.0--50.0)} & \makecell{48.6\\(45.5--51.5)} \\
			\quad \makecell{Surgical  \\Same Day} & 1022 & 6.5 & \makecell{70.5\\(65.5--75.5)} & \makecell{13.2\\(9.2--19.6)} & \makecell{21.7\\(16.5--27.2)} & \makecell{67.3\\(61.8--72.9)} & \makecell{12.2\\(8.4--18.2)} & \makecell{21.2\\(16.3--26.5)} \\
			\quad Elective & 624 & 9.3 & \makecell{73.9\\(68.7--79.2)} & \makecell{19.1\\(13.6--28.4)} & \makecell{28.9\\(22.7--35.0)} & \makecell{67.0\\(60.9--73.0)} & \makecell{16.0\\(11.6--24.0)} & \makecell{26.0\\(20.7--31.7)} \\
			\bottomrule
		\end{tabular}
	}
\begin{tablenotes}
	\item  AUROC, AUPRC, and F1 are reported with 95\% confidence intervals. NORM, normal ECG; ILBBB, incomplete left bundle branch block. 
\end{tablenotes}
\end{threeparttable}
\end{table*}

\begin{table*}[!htbp]
	\centering
	\begin{threeparttable}
	\footnotesize
	\setlength{\tabcolsep}{1pt}{
		\caption{Subgroup performance of the single-predictor method for INJIL and ISCLA in the internal test set. 
			 \label{table:subgroup_INJIL_ISCLA_lvefllm}}
		\begin{tabular}{lcc|ccc|ccc}
			\toprule
			& & & \multicolumn{3}{c}{\textbf{INJIL}} & \multicolumn{3}{c}{\textbf{ISCLA}} \\
			\cmidrule(lr){4-6} \cmidrule(lr){7-9}
			\textbf{Subgroup} & \textbf{n} & \textbf{Prevalence (\%)} &
			\textbf{AUROC} & \textbf{AUPRC} & \textbf{F1 Score} &
			\textbf{AUROC} & \textbf{AUPRC} & \textbf{F1 Score} \\
			\midrule
			\multicolumn{9}{l}{\textbf{Age groups}} \\
			\midrule
			\quad 18--59 & 4270 & 12.2 & \makecell{73.1\\(71.0--75.1)} & \makecell{25.7\\(22.4--29.6)} & \makecell{42.2\\(39.1--45.5)} & \makecell{61.0\\(58.6--63.4)} & \makecell{23.2\\(20.3--26.6)} & \makecell{32.4\\(30.0--34.8)} \\
			\quad 60--69 & 3637 & 15.7 & \makecell{72.4\\(70.4--74.5)} & \makecell{28.9\\(25.6--32.8)} & \makecell{45.0\\(42.4--47.6)} & \makecell{61.3\\(59.0--63.6)} & \makecell{26.2\\(23.4--29.6)} & \makecell{35.6\\(33.2--38.1)} \\
			\quad 70--79 & 3761 & 18.0 & \makecell{71.6\\(69.8--73.5)} & \makecell{29.4\\(26.9--32.4)} & \makecell{43.5\\(41.1--45.9)} & \makecell{62.7\\(60.6--64.9)} & \makecell{25.9\\(23.6--29.1)} & \makecell{33.2\\(30.7--35.5)} \\
			\quad 80+ & 4349 & 17.2 & \makecell{69.9\\(68.0--71.7)} & \makecell{27.3\\(25.1--30.2)} & \makecell{39.2\\(36.9--41.3)} & \makecell{62.2\\(60.2--64.4)} & \makecell{24.7\\(22.7--27.4)} & \makecell{32.0\\(29.8--34.3)} \\
			\midrule
			\multicolumn{9}{l}{\textbf{Sex}} \\
			\midrule
			\quad Female & 7222 & 11.2 & \makecell{67.1\\(65.4--68.8)} & \makecell{18.7\\(16.4--21.5)} & \makecell{31.6\\(29.7--33.6)} & \makecell{58.1\\(55.7--60.5)} & \makecell{16.3\\(14.2--18.9)} & \makecell{26.3\\(23.9--28.7)} \\
			\quad Male & 8795 & 19.4 & \makecell{73.8\\(72.6--75.1)} & \makecell{34.5\\(32.3--36.9)} & \makecell{44.5\\(42.7--46.3)} & \makecell{65.6\\(64.0--67.1)} & \makecell{33.0\\(30.6--35.5)} & \makecell{39.5\\(37.7--41.4)} \\
			\midrule
			\multicolumn{9}{l}{\textbf{Race/ethnicity}} \\
			\midrule
			\quad Hispanic & 638 & 14.7 & \makecell{75.8\\(71.3--80.1)} & \makecell{30.7\\(23.9--39.2)} & \makecell{45.9\\(39.5--52.2)} & \makecell{62.3\\(56.4--68.2)} & \makecell{27.1\\(20.3--35.8)} & \makecell{34.8\\(27.6--42.2)} \\
			\quad White & 11688 & 15.5 & \makecell{71.7\\(70.7--72.7)} & \makecell{29.8\\(28.0--31.8)} & \makecell{41.7\\(40.4--43.0)} & \makecell{62.0\\(60.8--63.1)} & \makecell{27.3\\(25.5--29.2)} & \makecell{34.9\\(33.5--36.3)} \\
			\quad Black & 1845 & 17.3 & \makecell{74.6\\(72.3--76.8)} & \makecell{37.0\\(32.6--41.7)} & \makecell{45.1\\(41.5--48.8)} & \makecell{64.5\\(61.6--67.5)} & \makecell{34.2\\(29.9--38.8)} & \makecell{38.8\\(35.0--42.7)} \\
			\quad Asian & 414 & 12.6 & \makecell{69.2\\(63.6--74.6)} & \makecell{20.1\\(14.2--28.8)} & \makecell{36.7\\(28.0--45.5)} & \makecell{62.9\\(56.8--69.4)} & \makecell{20.4\\(14.5--29.2)} & \makecell{32.8\\(25.1--40.9)} \\
			\quad Other & 524 & 13.0 & \makecell{68.8\\(63.0--74.7)} & \makecell{17.7\\(12.6--25.4)} & \makecell{34.5\\(27.5--41.7)} & \makecell{61.4\\(55.4--68.0)} & \makecell{16.8\\(12.3--24.3)} & \makecell{30.8\\(23.7--38.1)} \\
			\quad Unknown & 908 & 18.8 & \makecell{69.5\\(65.2--73.5)} & \makecell{31.5\\(26.4--37.2)} & \makecell{40.8\\(35.9--45.7)} & \makecell{60.7\\(56.0--65.2)} & \makecell{28.5\\(24.2--33.5)} & \makecell{35.9\\(30.9--40.7)} \\
			\midrule
			\multicolumn{9}{l}{\textbf{Clinical context}} \\
			\midrule
			\quad Emergency & 9170 & 14.6 & \makecell{71.6\\(70.3--72.8)} & \makecell{28.2\\(26.4--30.2)} & \makecell{40.6\\(39.1--42.2)} & \makecell{62.2\\(60.8--63.6)} & \makecell{26.4\\(24.6--28.3)} & \makecell{35.8\\(34.3--37.3)} \\
			\quad Urgent & 3028 & 20.9 & \makecell{74.6\\(72.9--76.3)} & \makecell{39.2\\(35.8--42.9)} & \makecell{49.0\\(46.5--51.6)} & \makecell{64.2\\(62.2--66.3)} & \makecell{36.3\\(33.0--39.8)} & \makecell{44.9\\(42.2--47.5)} \\
			\quad Observation & 2173 & 19.6 & \makecell{74.7\\(72.8--76.5)} & \makecell{42.6\\(38.2--47.1)} & \makecell{51.3\\(48.4--54.1)} & \makecell{65.4\\(63.3--67.7)} & \makecell{39.7\\(35.3--44.1)} & \makecell{48.6\\(45.5--51.5)} \\
			\quad \makecell{Surgical \\ Same Day} & 1022 & 6.5 & \makecell{66.9\\(61.9--72.0)} & \makecell{12.3\\(8.6--18.1)} & \makecell{21.7\\(16.5--27.2)} & \makecell{61.5\\(56.0--67.1)} & \makecell{12.0\\(8.4--17.8)} & \makecell{21.2\\(16.3--26.5)} \\
			\quad Elective & 624 & 9.3 & \makecell{70.9\\(66.0--75.9)} & \makecell{18.1\\(12.9--26.8)} & \makecell{28.9\\(22.7--35.0)} & \makecell{62.6\\(56.6--68.7)} & \makecell{16.9\\(12.1--25.2)} & \makecell{26.0\\(20.7--31.7)} \\
			\bottomrule
		\end{tabular}
	}
\begin{tablenotes}
	\item  AUROC, AUPRC, and F1 are reported with 95\% confidence intervals. INJIL,  subendocardial injury in inferolateral leads; ISCLA, ischemic in lateral leads.
\end{tablenotes}
\end{threeparttable}
\end{table*}

\newpage

\bibliographystyle{rss} 
\bibliography{reference}

\end{document}